\documentclass[11pt]{article}

\usepackage[preprint]{acl}

\usepackage{times}
\usepackage{latexsym}
\usepackage[T1]{fontenc}
\usepackage[utf8]{inputenc}
\usepackage{microtype}
\usepackage{inconsolata}
\usepackage{natbib}
\usepackage{amssymb}

\usepackage{xspace}
\usepackage{xcolor}
\usepackage{colortbl}
\usepackage{graphicx}
\usepackage[normalem]{ulem} 
\usepackage{amsmath}
\usepackage{enumitem}
\usepackage{tcolorbox}
\usepackage{algorithm}
\usepackage{algpseudocode}
\usepackage{algorithm}
\usepackage{algpseudocode}
\usepackage{algpseudocodex}
\usepackage{xcolor} 
\algrenewcommand{\algorithmiccomment}[1]{\hfill\textcolor{teal}{// #1}}
\algrenewcommand{\LComment}[1]{\Statex \textcolor{teal}{// #1}}

\usepackage{hyperref}
\hypersetup{linkcolor=black,citecolor=black,anchorcolor=black,filecolor=black,menucolor=black,runcolor=black,urlcolor=black,hidelinks}
\usepackage{breakurl}

\usepackage{booktabs}
\usepackage{multirow}
\usepackage{makecell}
\usepackage{ragged2e}

\usepackage{caption}
\usepackage{subcaption}

\usepackage{listings}

\usepackage{tikz}
\usetikzlibrary{calc}
\usetikzlibrary{shapes.geometric}
\usetikzlibrary{decorations.pathreplacing}
\usetikzlibrary{positioning}

\usepackage{datetime} 
\usepackage{tcolorbox}

\makeatletter
\newcommand{\DefMacro}{\@ifstar\@DefMacroAllowRedefine\@DefMacro}
\newcommand{\@DefMacro}[2]{\expandafter\newcommand\csname rmk-#1\endcsname{#2}}
\newcommand{\@DefMacroAllowRedefine}[2]{\expandafter\providecommand\csname rmk-#1\endcsname{} \expandafter\renewcommand\csname rmk-#1\endcsname{#2}}
\makeatother
\newcommand{\UseMacro}[1]{\csname rmk-#1\endcsname}

\newcommand{\XSpace}[1]{}
\newcommand{\XComment}[1]{}

\newcommand{\Code}[1]{{\ifmmode{\mathtt{#1}}\else$\mathtt{#1}$\fi}}
\newcommand{\CodeIn}[1]{{\ifmmode{\mathtt{#1}}\else$\mathtt{#1}$\fi}}

\newcolumntype{R}[1]{>{\RaggedLeft\arraybackslash}p{#1}}
\newcolumntype{L}[1]{>{\RaggedRight\arraybackslash}p{#1}}

\definecolor{gray}{RGB}{211,211,211}
\newcommand{\jbasicstyle}{\small\sffamily} 

\newcommand{\jnumberstyle}{\scriptsize}

\lstdefinelanguage{pseudo}
{
  morekeywords={},
  keywordstyle=\bfseries,
  lineskip=-0.1em,
  numbers=left, 
  numberstyle=\jnumberstyle,
  numbersep=4pt,
  basicstyle=\jbasicstyle,
  breaklines=true,
  breakautoindent=true,
  tabsize=2,
  columns=fullflexible,
  morecomment=*[l][\textsl]{//},
  mathescape=true,
  xleftmargin=10pt,
}

\lstdefinelanguage{todo-comment}
{
  morekeywords={},
  keywordstyle=\bfseries,
  lineskip=-0.1em,
  numbers=none,
  basicstyle=\jbasicstyle,
  breaklines=true,
  breakautoindent=true,
  tabsize=2,
  columns=fullflexible,
  morecomment=*[l][\textsl]{//},
  mathescape=true,
  xleftmargin=-10pt,
}

\lstdefinelanguage{java-pretty}
{
  language=java,
  numbers=left,
  basicstyle=\scriptsize\ttfamily,
  numberstyle=\scriptsize,
  breaklines=true,
  columns=fullflexible,
  xleftmargin=16pt,
  showstringspaces=false,
}

\newtcbox{\TokenOldBox}{on line, arc=2pt, colback=red!10, colframe=gray, boxrule=0.4pt, boxsep=0pt, top=0.25ex, bottom=0.25ex, left=0.5ex, right=0.5ex, tcbox raise base}
\newtcbox{\TokenNewBox}{on line, arc=2pt, colback=green!10, colframe=gray, boxrule=0.4pt, boxsep=0pt, top=0.25ex, bottom=0.25ex, left=0.5ex, right=0.5ex, tcbox raise base}
\newtcbox{\TokenCtxBox}{on line, arc=2pt, colback=gray!10, colframe=gray, boxrule=0.4pt, boxsep=0pt, top=0.25ex, bottom=0.25ex, left=0.5ex, right=0.5ex, tcbox raise base}
\newcommand{\TokenOld}[1]{\TokenOldBox{\texttt{#1\strut}}}
\newcommand{\TokenNew}[1]{\TokenNewBox{\texttt{#1\strut}}}
\newcommand{\TokenCtx}[1]{\TokenCtxBox{\texttt{#1\strut}}}

\newcommand{\LangJava}{%
  \tikz[baseline=(char.base)]{%
    \node[draw=black, circle, fill=orange!50, minimum size=1.8ex, inner sep=0pt, line width=0.4pt] (char) {\sffamily\bfseries\footnotesize J};%
  }%
}
\newcommand{\LangPython}{%
  \tikz[baseline=(char.base)]{%
    \node[draw=black, circle, fill=blue!20, minimum size=1.8ex, inner sep=0pt, line width=0.4pt] (char) {\sffamily\bfseries\footnotesize P};%
  }%
}
\newcommand{\LangBoth}{\LangJava\LangPython}

\newcommand{\Tool}{\textsc{TokDrift}\xspace}

\newcommand{\Title}{\Tool: When LLM Speaks in Subwords \\but Code Speaks in Grammar}

\newcommand{\subword}{subword\xspace}
\newcommand{\subwords}{subwords\xspace}

\newcommand{\acc}{accuracy\xspace}
\newcommand{\Acc}{Accuracy\xspace}
\newcommand{\accdelta}{\ensuremath{\Delta}\acc{}\xspace}

\newcommand{\sensitivity}{sensitivity\xspace}
\newcommand{\Sensitivity}{Sensitivity\xspace}
\newcommand{\rewriterule}{rewrite rule\xspace}
\newcommand{\rewriterules}{rewrite rules\xspace}

\newcommand{\rewrites}{rewrites\xspace}
\newcommand{\pretrained}{pretrained\xspace}

\newcommand{\pretraining}{pretraining\xspace}
\newcommand{\llmtokenizer}{LLM tokenizer\xspace}
\newcommand{\llmtoken}{LLM token\xspace}
\newcommand{\llmtokens}{LLM tokens\xspace}

\newcommand{\llama}{Llama-3\xspace}
\newcommand{\qwen}{Qwen2.5-Coder\xspace}
\newcommand{\dscoder}{DeepSeek-Coder\xspace}

\DefMacro{llama-medium}{Llama-3.1-8B-Instruct}
\DefMacro{llama-small}{Llama-3.2-3B-Instruct}
\DefMacro{llama-large}{Llama-3.3-70B-Instruct}
\DefMacro{qwen-medium}{Qwen2.5-Coder-7B-Instruct}
\DefMacro{qwen-small}{Qwen2.5-Coder-1.5B-Instruct}
\DefMacro{qwen-large}{Qwen2.5-Coder-32B-Instruct}
\DefMacro{dscoder-small}{deepseek-coder-1.3b-instruct}
\DefMacro{dscoder-medium}{deepseek-coder-6.7b-instruct}
\DefMacro{dscoder-large}{deepseek-coder-33b-instruct}
\DefMacro{codeqwen}{CodeQwen1.5-7B-Chat}
\DefMacro{qwen-base}{Qwen2.5-Coder-7B}

\DefMacro{llama-medium-short}{Llama-8B}
\DefMacro{llama-small-short}{Llama-3B}
\DefMacro{llama-large-short}{Llama-70B}
\DefMacro{qwen-medium-short}{Qwen-7B}
\DefMacro{qwen-small-short}{Qwen-1.5B}
\DefMacro{qwen-large-short}{Qwen-32B}
\DefMacro{deepseek-coder-1.3b-short}{DS-1.3B}
\DefMacro{deepseek-coder-6.7b-short}{DS-6.7B}
\DefMacro{deepseek-coder-33b-short}{DS-33B}
\DefMacro{codeqwen1.5-7b-short}{CodeQwen-7B}
\DefMacro{qwen-base-short}{Qwen2.5-7B-Base}

\DefMacro{llama-medium-init}{Llama-M}
\DefMacro{llama-small-init}{Llama-S}
\DefMacro{llama-large-init}{Llama-L}
\DefMacro{qwen-medium-init}{Qwen-M}
\DefMacro{qwen-small-init}{Qwen-S}
\DefMacro{qwen-large-init}{Qwen-L}
\DefMacro{dscoder-small-init}{DS-S}
\DefMacro{dscoder-medium-init}{DS-M}
\DefMacro{dscoder-large-init}{DS-L}
\DefMacro{codeqwen-init}{CodeQwen-M}
\DefMacro{qwen-base-init}{Qwen-Base}

\DefMacro{camel-case_pascal-case}{camelCase$\rightarrow$PascalCase}
\DefMacro{camel-case_screaming-snake-case}{camelCase$\rightarrow$SCREAMING\_SNAKE\_CASE}
\DefMacro{camel-case_snake-case}{camelCase$\rightarrow$snake\_case}
\DefMacro{snake-case_camel-case}{snake\_case$\rightarrow$CamelCase}
\DefMacro{snake-case_pascal-case}{snake\_case$\rightarrow$PascalCase}
\DefMacro{snake-case_screaming-snake-case}{snake\_case$\rightarrow$SCREAMING\_SNAKE\_CASE}

\DefMacro{baseline}{baseline}
\DefMacro{lparentheses-name}{lparentheses identifier}
\DefMacro{lparentheses-rparentheses}{lparentheses rparentheses}
\DefMacro{rparentheses-colon}{rparentheses colon}
\DefMacro{rparentheses-rparentheses}{rparentheses rparentheses}
\DefMacro{lsquarebracket-name}{lsquarebracket identifier}
\DefMacro{period-name}{period identifier}
\DefMacro{rsquarebracket-rparentheses}{rsquarebracket rparentheses}
\DefMacro{op-lsquarebracket}{operator lsquarebracket}
\DefMacro{op-rsquarebracket}{operator rsquarebracket}
\DefMacro{op-dash}{operator dash}
\DefMacro{op-name}{operator identifier}
\DefMacro{op-all}{operator all}
\DefMacro{rparentheses-semicolon}{rparentheses semicolon}
\DefMacro{period-asterisk}{period asterisk}
\DefMacro{double-plus-rparentheses}{double-plus rparentheses}
\DefMacro{rparentheses-period}{rparentheses period}
\DefMacro{op-semicolon}{operator semicolon}
\DefMacro{op-lparentheses}{operator lparentheses}

\DefMacro{baseline-short}{BL}

\DefMacro{camel-case-snake-case-no}{N1}
\DefMacro{camel-case-pascal-case-no}{N2}
\DefMacro{camel-case-screaming-snake-case-no}{N3}

\DefMacro{snake-case-camel-case-no}{N4}
\DefMacro{snake-case-pascal-case-no}{N5}
\DefMacro{snake-case-screaming-snake-case-no}{N6}

\DefMacro{N1-short-f}{\aCamelCase{}\xspace}
\DefMacro{N1-short-l}{\aSnakeCase{}\xspace}
\DefMacro{N1-short}{\UseMacro{N1-short-f}$\rightarrow$\UseMacro{N1-short-l}}

\DefMacro{N2-short-f}{\aCamelCase{}\xspace}
\DefMacro{N2-short-l}{\aPascalCase{}\xspace}
\DefMacro{N2-short}{\UseMacro{N2-short-f}$\rightarrow$\UseMacro{N2-short-l}}

\DefMacro{N3-short-f}{\aCamelCase{}\xspace}
\DefMacro{N3-short-l}{\aScreamingSnakeCase{}\xspace}
\DefMacro{N3-short}{\UseMacro{N3-short-f}$\rightarrow$\UseMacro{N3-short-l}}

\DefMacro{N4-short-f}{\aSnakeCase{}\xspace}
\DefMacro{N4-short-l}{\aCamelCase{}\xspace}
\DefMacro{N4-short}{\UseMacro{N4-short-f}$\rightarrow$\UseMacro{N4-short-l}}

\DefMacro{N5-short-f}{\aSnakeCase{}\xspace}
\DefMacro{N5-short-l}{\aPascalCase{}\xspace}
\DefMacro{N5-short}{\UseMacro{N5-short-f}$\rightarrow$\UseMacro{N5-short-l}}

\DefMacro{N6-short-f}{\aSnakeCase{}\xspace}
\DefMacro{N6-short-l}{\aScreamingSnakeCase{}\xspace}
\DefMacro{N6-short}{\UseMacro{N6-short-f}$\rightarrow$\UseMacro{N6-short-l}}

\DefMacro{rparentheses-semicolon-no}{S11}
\DefMacro{period-name-no}{S15}
\DefMacro{lparentheses-rparentheses-no}{S14}
\DefMacro{lparentheses-name-no}{S16}
\DefMacro{rparentheses-rparentheses-no}{S13}
\DefMacro{period-asterisk-no}{S9}
\DefMacro{double-plus-rparentheses-no}{S8}
\DefMacro{rparentheses-period-no}{S3}
\DefMacro{rparentheses-colon-no}{S10}
\DefMacro{lsquarebracket-name-no}{S7}
\DefMacro{rsquarebracket-rparentheses-no}{S4}
\DefMacro{op-semicolon-no}{S12}
\DefMacro{op-lparentheses-no}{S6}
\DefMacro{op-lsquarebracket-no}{S2}
\DefMacro{op-rsquarebracket-no}{S5}
\DefMacro{op-dash-no}{S1}
\DefMacro{op-name-no}{S17}
\DefMacro{op-all-no}{S18}

\DefMacro{S11-short-f}{\texttt{)\,;}}
\DefMacro{S11-short-l}{\texttt{)\,\textvisiblespace\,;}}
\DefMacro{S11-short}{\UseMacro{S11-short-f}$\rightarrow$\UseMacro{S11-short-l}}

\DefMacro{S15-short-f}{\texttt{.\,\aID}}
\DefMacro{S15-short-l}{\texttt{.\,\textvisiblespace\,\aID}}
\DefMacro{S15-short}{\UseMacro{S15-short-f}$\rightarrow$\UseMacro{S15-short-l}}

\DefMacro{S14-short-f}{\texttt{(\,)}}
\DefMacro{S14-short-l}{\texttt{(\,\textvisiblespace\,)}}
\DefMacro{S14-short}{\UseMacro{S14-short-f}$\rightarrow$\UseMacro{S14-short-l}}

\DefMacro{S16-short-f}{\texttt{(\,\aID}}
\DefMacro{S16-short-l}{\texttt{(\,\textvisiblespace\,\aID}}
\DefMacro{S16-short}{\UseMacro{S16-short-f}$\rightarrow$\UseMacro{S16-short-l}}

\DefMacro{S13-short-f}{\texttt{)\,)}}
\DefMacro{S13-short-l}{\texttt{)\,\textvisiblespace\,)}}
\DefMacro{S13-short}{\UseMacro{S13-short-f}$\rightarrow$\UseMacro{S13-short-l}}

\DefMacro{S9-short-f}{\texttt{.\,*}}
\DefMacro{S9-short-l}{\texttt{.\,\textvisiblespace\,*}}
\DefMacro{S9-short}{\UseMacro{S9-short-f}$\rightarrow$\UseMacro{S9-short-l}}

\DefMacro{S8-short-f}{\texttt{++\,)}}
\DefMacro{S8-short-l}{\texttt{++\,\textvisiblespace\,)}}
\DefMacro{S8-short}{\UseMacro{S8-short-f}$\rightarrow$\UseMacro{S8-short-l}}

\DefMacro{S3-short-f}{\texttt{)\,.}}
\DefMacro{S3-short-l}{\texttt{)\,\textvisiblespace\,.}}
\DefMacro{S3-short}{\UseMacro{S3-short-f}$\rightarrow$\UseMacro{S3-short-l}}

\DefMacro{S10-short-f}{\texttt{)\,:}}
\DefMacro{S10-short-l}{\texttt{)\,\textvisiblespace\,:}}
\DefMacro{S10-short}{\UseMacro{S10-short-f}$\rightarrow$\UseMacro{S10-short-l}}

\DefMacro{S7-short-f}{\texttt{[\,\aID}}
\DefMacro{S7-short-l}{\texttt{[\,\textvisiblespace\,\aID}}
\DefMacro{S7-short}{\UseMacro{S7-short-f}$\rightarrow$\UseMacro{S7-short-l}}

\DefMacro{S4-short-f}{\texttt{]\,)}}
\DefMacro{S4-short-l}{\texttt{]\,\textvisiblespace\,)}}
\DefMacro{S4-short}{\UseMacro{S4-short-f}$\rightarrow$\UseMacro{S4-short-l}}

\DefMacro{S12-short-f}{\texttt{\aOP{}\,;}}
\DefMacro{S12-short-l}{\texttt{\aOP{}\,\textvisiblespace\,;}}
\DefMacro{S12-short}{\UseMacro{S12-short-f}$\rightarrow$\UseMacro{S12-short-l}}

\DefMacro{S6-short-f}{\texttt{\aOP{}\,(}}
\DefMacro{S6-short-l}{\texttt{\aOP\,\textvisiblespace\,(}}
\DefMacro{S6-short}{\UseMacro{S6-short-f}$\rightarrow$\UseMacro{S6-short-l}}

\DefMacro{S2-short-f}{\texttt{\aOP{}\,[}}
\DefMacro{S2-short-l}{\texttt{\aOP{}\,\textvisiblespace\,[}}
\DefMacro{S2-short}{\UseMacro{S2-short-f}$\rightarrow$\UseMacro{S2-short-l}}

\DefMacro{S5-short-f}{\texttt{\aOP{}\,]}}
\DefMacro{S5-short-l}{\texttt{\aOP{}\,\textvisiblespace\,]}}
\DefMacro{S5-short}{\UseMacro{S5-short-f}$\rightarrow$\UseMacro{S5-short-l}}

\DefMacro{S1-short-f}{\texttt{\aOP{}\,-}}
\DefMacro{S1-short-l}{\texttt{\aOP{}\,\textvisiblespace\,-}}
\DefMacro{S1-short}{\UseMacro{S1-short-f}$\rightarrow$\UseMacro{S1-short-l}}

\DefMacro{S17-short-f}{\texttt{\aOP{}\,\aID}}
\DefMacro{S17-short-l}{\texttt{\aOP{}\,\textvisiblespace\,\aID}}
\DefMacro{S17-short}{\UseMacro{S17-short-f}$\rightarrow$\UseMacro{S17-short-l}}

\DefMacro{S18-short-f}{\texttt{\aOP{}\,ALL}}
\DefMacro{S18-short-l}{\texttt{\aOP{}\,\textvisiblespace\,ALL}}
\DefMacro{S18-short}{\UseMacro{S18-short-f}$\rightarrow$\UseMacro{S18-short-l}}

\DefMacro{humaneval-explain-py}{HumanEval-Explain-py}
\DefMacro{humaneval-explain-java}{HumanEval-Explain-java}
\DefMacro{humaneval-fix-py}{HumanEval-Fix-py}
\DefMacro{humaneval-fix-java}{HumanEval-Fix-java}
\DefMacro{avatar-py2java}{Avatar-py2java}
\DefMacro{avatar-java2py}{Avatar-java2py}
\DefMacro{codenet-py2java}{CodeNet-py2java}
\DefMacro{codenet-java2py}{CodeNet-java2py}

\newcommand{\aCamelCase}{\CodeIn{camelCase}\xspace}
\newcommand{\aPascalCase}{\CodeIn{PascalCase}\xspace}
\newcommand{\aSnakeCase}{\CodeIn{snake\_case}\xspace}
\newcommand{\aScreamingSnakeCase}{\CodeIn{SCREAMING\_CASE}\xspace} 
\newcommand{\aOP}{\CodeIn{OP}\xspace}
\newcommand{\aID}{\CodeIn{ID}\xspace}

\DefMacro{TH-accuracy-average}{Average\xspace}
\DefMacro{TH-variant}{Variant\xspace}
\DefMacro{TH-java-results}{Input PL = Java\xspace}
\DefMacro{TH-python-results}{Input PL = Python\xspace}

\DefMacro{dscoder-small_camel-case_pascal-case_accuracy}{52.07}

\DefMacro{Llama-3_lsquarebracket_name_all}{414}
\DefMacro{Llama-3_lsquarebracket_name_unchanged}{401}
\DefMacro{Llama-3_lsquarebracket_name_changed}{13}
\DefMacro{Llama-3_lsquarebracket_name_merged}{12}
\DefMacro{Llama-3_lsquarebracket_name_split}{1}
\DefMacro{Llama-3_lsquarebracket_name_mixed}{0}
\DefMacro{Qwen25-Coder_lsquarebracket_name_all}{414}
\DefMacro{Qwen25-Coder_lsquarebracket_name_unchanged}{401}
\DefMacro{Qwen25-Coder_lsquarebracket_name_changed}{13}
\DefMacro{Qwen25-Coder_lsquarebracket_name_merged}{12}
\DefMacro{Qwen25-Coder_lsquarebracket_name_split}{0}
\DefMacro{Qwen25-Coder_lsquarebracket_name_mixed}{1}
\DefMacro{deepseek-coder_lsquarebracket_name_all}{414}
\DefMacro{deepseek-coder_lsquarebracket_name_unchanged}{391}
\DefMacro{deepseek-coder_lsquarebracket_name_changed}{23}
\DefMacro{deepseek-coder_lsquarebracket_name_merged}{13}
\DefMacro{deepseek-coder_lsquarebracket_name_split}{10}
\DefMacro{deepseek-coder_lsquarebracket_name_mixed}{0}
\DefMacro{CodeQwen15_lsquarebracket_name_all}{414}
\DefMacro{CodeQwen15_lsquarebracket_name_unchanged}{394}
\DefMacro{CodeQwen15_lsquarebracket_name_changed}{20}
\DefMacro{CodeQwen15_lsquarebracket_name_merged}{13}
\DefMacro{CodeQwen15_lsquarebracket_name_split}{7}
\DefMacro{CodeQwen15_lsquarebracket_name_mixed}{0}

\DefMacro{Llama-3_snake_case-camel_case_all}{343}
\DefMacro{Llama-3_snake_case-camel_case_unchanged}{260}
\DefMacro{Llama-3_snake_case-camel_case_changed}{83}
\DefMacro{Llama-3_snake_case-camel_case_merged}{55}
\DefMacro{Llama-3_snake_case-camel_case_split}{21}
\DefMacro{Llama-3_snake_case-camel_case_mixed}{7}
\DefMacro{Qwen25-Coder_snake_case-camel_case_all}{343}
\DefMacro{Qwen25-Coder_snake_case-camel_case_unchanged}{260}
\DefMacro{Qwen25-Coder_snake_case-camel_case_changed}{83}
\DefMacro{Qwen25-Coder_snake_case-camel_case_merged}{55}
\DefMacro{Qwen25-Coder_snake_case-camel_case_split}{21}
\DefMacro{Qwen25-Coder_snake_case-camel_case_mixed}{7}
\DefMacro{deepseek-coder_snake_case-camel_case_all}{343}
\DefMacro{deepseek-coder_snake_case-camel_case_unchanged}{238}
\DefMacro{deepseek-coder_snake_case-camel_case_changed}{105}
\DefMacro{deepseek-coder_snake_case-camel_case_merged}{6}
\DefMacro{deepseek-coder_snake_case-camel_case_split}{76}
\DefMacro{deepseek-coder_snake_case-camel_case_mixed}{23}
\DefMacro{CodeQwen15_snake_case-camel_case_all}{343}
\DefMacro{CodeQwen15_snake_case-camel_case_unchanged}{216}
\DefMacro{CodeQwen15_snake_case-camel_case_changed}{127}
\DefMacro{CodeQwen15_snake_case-camel_case_merged}{4}
\DefMacro{CodeQwen15_snake_case-camel_case_split}{104}
\DefMacro{CodeQwen15_snake_case-camel_case_mixed}{19}

\DefMacro{Llama-3_op_lparentheses_all}{291}
\DefMacro{Llama-3_op_lparentheses_unchanged}{288}
\DefMacro{Llama-3_op_lparentheses_changed}{3}
\DefMacro{Llama-3_op_lparentheses_merged}{3}
\DefMacro{Llama-3_op_lparentheses_split}{0}
\DefMacro{Llama-3_op_lparentheses_mixed}{0}
\DefMacro{Qwen25-Coder_op_lparentheses_all}{291}
\DefMacro{Qwen25-Coder_op_lparentheses_unchanged}{288}
\DefMacro{Qwen25-Coder_op_lparentheses_changed}{3}
\DefMacro{Qwen25-Coder_op_lparentheses_merged}{3}
\DefMacro{Qwen25-Coder_op_lparentheses_split}{0}
\DefMacro{Qwen25-Coder_op_lparentheses_mixed}{0}
\DefMacro{deepseek-coder_op_lparentheses_all}{291}
\DefMacro{deepseek-coder_op_lparentheses_unchanged}{123}
\DefMacro{deepseek-coder_op_lparentheses_changed}{168}
\DefMacro{deepseek-coder_op_lparentheses_merged}{1}
\DefMacro{deepseek-coder_op_lparentheses_split}{167}
\DefMacro{deepseek-coder_op_lparentheses_mixed}{0}
\DefMacro{CodeQwen15_op_lparentheses_all}{291}
\DefMacro{CodeQwen15_op_lparentheses_unchanged}{123}
\DefMacro{CodeQwen15_op_lparentheses_changed}{168}
\DefMacro{CodeQwen15_op_lparentheses_merged}{1}
\DefMacro{CodeQwen15_op_lparentheses_split}{167}
\DefMacro{CodeQwen15_op_lparentheses_mixed}{0}

\DefMacro{Llama-3_period_asterisk_all}{550}
\DefMacro{Llama-3_period_asterisk_unchanged}{10}
\DefMacro{Llama-3_period_asterisk_changed}{540}
\DefMacro{Llama-3_period_asterisk_merged}{0}
\DefMacro{Llama-3_period_asterisk_split}{540}
\DefMacro{Llama-3_period_asterisk_mixed}{0}
\DefMacro{Qwen25-Coder_period_asterisk_all}{550}
\DefMacro{Qwen25-Coder_period_asterisk_unchanged}{10}
\DefMacro{Qwen25-Coder_period_asterisk_changed}{540}
\DefMacro{Qwen25-Coder_period_asterisk_merged}{0}
\DefMacro{Qwen25-Coder_period_asterisk_split}{540}
\DefMacro{Qwen25-Coder_period_asterisk_mixed}{0}
\DefMacro{deepseek-coder_period_asterisk_all}{550}
\DefMacro{deepseek-coder_period_asterisk_unchanged}{13}
\DefMacro{deepseek-coder_period_asterisk_changed}{537}
\DefMacro{deepseek-coder_period_asterisk_merged}{0}
\DefMacro{deepseek-coder_period_asterisk_split}{537}
\DefMacro{deepseek-coder_period_asterisk_mixed}{0}
\DefMacro{CodeQwen15_period_asterisk_all}{550}
\DefMacro{CodeQwen15_period_asterisk_unchanged}{550}
\DefMacro{CodeQwen15_period_asterisk_changed}{0}
\DefMacro{CodeQwen15_period_asterisk_merged}{0}
\DefMacro{CodeQwen15_period_asterisk_split}{0}
\DefMacro{CodeQwen15_period_asterisk_mixed}{0}

\DefMacro{Llama-3_rparentheses_period_all}{187}
\DefMacro{Llama-3_rparentheses_period_unchanged}{184}
\DefMacro{Llama-3_rparentheses_period_changed}{3}
\DefMacro{Llama-3_rparentheses_period_merged}{3}
\DefMacro{Llama-3_rparentheses_period_split}{0}
\DefMacro{Llama-3_rparentheses_period_mixed}{0}
\DefMacro{Qwen25-Coder_rparentheses_period_all}{187}
\DefMacro{Qwen25-Coder_rparentheses_period_unchanged}{184}
\DefMacro{Qwen25-Coder_rparentheses_period_changed}{3}
\DefMacro{Qwen25-Coder_rparentheses_period_merged}{3}
\DefMacro{Qwen25-Coder_rparentheses_period_split}{0}
\DefMacro{Qwen25-Coder_rparentheses_period_mixed}{0}
\DefMacro{deepseek-coder_rparentheses_period_all}{187}
\DefMacro{deepseek-coder_rparentheses_period_unchanged}{179}
\DefMacro{deepseek-coder_rparentheses_period_changed}{8}
\DefMacro{deepseek-coder_rparentheses_period_merged}{7}
\DefMacro{deepseek-coder_rparentheses_period_split}{0}
\DefMacro{deepseek-coder_rparentheses_period_mixed}{1}
\DefMacro{CodeQwen15_rparentheses_period_all}{187}
\DefMacro{CodeQwen15_rparentheses_period_unchanged}{181}
\DefMacro{CodeQwen15_rparentheses_period_changed}{6}
\DefMacro{CodeQwen15_rparentheses_period_merged}{5}
\DefMacro{CodeQwen15_rparentheses_period_split}{0}
\DefMacro{CodeQwen15_rparentheses_period_mixed}{1}

\DefMacro{Llama-3_op_name_all}{1542}
\DefMacro{Llama-3_op_name_unchanged}{845}
\DefMacro{Llama-3_op_name_changed}{697}
\DefMacro{Llama-3_op_name_merged}{132}
\DefMacro{Llama-3_op_name_split}{408}
\DefMacro{Llama-3_op_name_mixed}{157}
\DefMacro{Qwen25-Coder_op_name_all}{1542}
\DefMacro{Qwen25-Coder_op_name_unchanged}{845}
\DefMacro{Qwen25-Coder_op_name_changed}{697}
\DefMacro{Qwen25-Coder_op_name_merged}{132}
\DefMacro{Qwen25-Coder_op_name_split}{407}
\DefMacro{Qwen25-Coder_op_name_mixed}{158}
\DefMacro{deepseek-coder_op_name_all}{1542}
\DefMacro{deepseek-coder_op_name_unchanged}{733}
\DefMacro{deepseek-coder_op_name_changed}{809}
\DefMacro{deepseek-coder_op_name_merged}{124}
\DefMacro{deepseek-coder_op_name_split}{483}
\DefMacro{deepseek-coder_op_name_mixed}{202}
\DefMacro{CodeQwen15_op_name_all}{1542}
\DefMacro{CodeQwen15_op_name_unchanged}{705}
\DefMacro{CodeQwen15_op_name_changed}{837}
\DefMacro{CodeQwen15_op_name_merged}{162}
\DefMacro{CodeQwen15_op_name_split}{405}
\DefMacro{CodeQwen15_op_name_mixed}{270}

\DefMacro{Llama-3_snake_case-pascal_case_all}{343}
\DefMacro{Llama-3_snake_case-pascal_case_unchanged}{212}
\DefMacro{Llama-3_snake_case-pascal_case_changed}{131}
\DefMacro{Llama-3_snake_case-pascal_case_merged}{33}
\DefMacro{Llama-3_snake_case-pascal_case_split}{67}
\DefMacro{Llama-3_snake_case-pascal_case_mixed}{31}
\DefMacro{Qwen25-Coder_snake_case-pascal_case_all}{343}
\DefMacro{Qwen25-Coder_snake_case-pascal_case_unchanged}{212}
\DefMacro{Qwen25-Coder_snake_case-pascal_case_changed}{131}
\DefMacro{Qwen25-Coder_snake_case-pascal_case_merged}{33}
\DefMacro{Qwen25-Coder_snake_case-pascal_case_split}{67}
\DefMacro{Qwen25-Coder_snake_case-pascal_case_mixed}{31}
\DefMacro{deepseek-coder_snake_case-pascal_case_all}{343}
\DefMacro{deepseek-coder_snake_case-pascal_case_unchanged}{144}
\DefMacro{deepseek-coder_snake_case-pascal_case_changed}{199}
\DefMacro{deepseek-coder_snake_case-pascal_case_merged}{7}
\DefMacro{deepseek-coder_snake_case-pascal_case_split}{154}
\DefMacro{deepseek-coder_snake_case-pascal_case_mixed}{38}
\DefMacro{CodeQwen15_snake_case-pascal_case_all}{343}
\DefMacro{CodeQwen15_snake_case-pascal_case_unchanged}{139}
\DefMacro{CodeQwen15_snake_case-pascal_case_changed}{204}
\DefMacro{CodeQwen15_snake_case-pascal_case_merged}{5}
\DefMacro{CodeQwen15_snake_case-pascal_case_split}{167}
\DefMacro{CodeQwen15_snake_case-pascal_case_mixed}{32}

\DefMacro{Llama-3_period_name_all}{1272}
\DefMacro{Llama-3_period_name_unchanged}{700}
\DefMacro{Llama-3_period_name_changed}{572}
\DefMacro{Llama-3_period_name_merged}{23}
\DefMacro{Llama-3_period_name_split}{501}
\DefMacro{Llama-3_period_name_mixed}{48}
\DefMacro{Qwen25-Coder_period_name_all}{1272}
\DefMacro{Qwen25-Coder_period_name_unchanged}{700}
\DefMacro{Qwen25-Coder_period_name_changed}{572}
\DefMacro{Qwen25-Coder_period_name_merged}{23}
\DefMacro{Qwen25-Coder_period_name_split}{501}
\DefMacro{Qwen25-Coder_period_name_mixed}{48}
\DefMacro{deepseek-coder_period_name_all}{1272}
\DefMacro{deepseek-coder_period_name_unchanged}{715}
\DefMacro{deepseek-coder_period_name_changed}{557}
\DefMacro{deepseek-coder_period_name_merged}{41}
\DefMacro{deepseek-coder_period_name_split}{423}
\DefMacro{deepseek-coder_period_name_mixed}{93}
\DefMacro{CodeQwen15_period_name_all}{1272}
\DefMacro{CodeQwen15_period_name_unchanged}{690}
\DefMacro{CodeQwen15_period_name_changed}{582}
\DefMacro{CodeQwen15_period_name_merged}{50}
\DefMacro{CodeQwen15_period_name_split}{375}
\DefMacro{CodeQwen15_period_name_mixed}{157}

\DefMacro{Llama-3_double_plus_rparentheses_all}{421}
\DefMacro{Llama-3_double_plus_rparentheses_unchanged}{420}
\DefMacro{Llama-3_double_plus_rparentheses_changed}{1}
\DefMacro{Llama-3_double_plus_rparentheses_merged}{0}
\DefMacro{Llama-3_double_plus_rparentheses_split}{1}
\DefMacro{Llama-3_double_plus_rparentheses_mixed}{0}
\DefMacro{Qwen25-Coder_double_plus_rparentheses_all}{421}
\DefMacro{Qwen25-Coder_double_plus_rparentheses_unchanged}{420}
\DefMacro{Qwen25-Coder_double_plus_rparentheses_changed}{1}
\DefMacro{Qwen25-Coder_double_plus_rparentheses_merged}{0}
\DefMacro{Qwen25-Coder_double_plus_rparentheses_split}{1}
\DefMacro{Qwen25-Coder_double_plus_rparentheses_mixed}{0}
\DefMacro{deepseek-coder_double_plus_rparentheses_all}{421}
\DefMacro{deepseek-coder_double_plus_rparentheses_unchanged}{420}
\DefMacro{deepseek-coder_double_plus_rparentheses_changed}{1}
\DefMacro{deepseek-coder_double_plus_rparentheses_merged}{0}
\DefMacro{deepseek-coder_double_plus_rparentheses_split}{0}
\DefMacro{deepseek-coder_double_plus_rparentheses_mixed}{1}
\DefMacro{CodeQwen15_double_plus_rparentheses_all}{421}
\DefMacro{CodeQwen15_double_plus_rparentheses_unchanged}{421}
\DefMacro{CodeQwen15_double_plus_rparentheses_changed}{0}
\DefMacro{CodeQwen15_double_plus_rparentheses_merged}{0}
\DefMacro{CodeQwen15_double_plus_rparentheses_split}{0}
\DefMacro{CodeQwen15_double_plus_rparentheses_mixed}{0}

\DefMacro{Llama-3_lparentheses_name_all}{1537}
\DefMacro{Llama-3_lparentheses_name_unchanged}{1287}
\DefMacro{Llama-3_lparentheses_name_changed}{250}
\DefMacro{Llama-3_lparentheses_name_merged}{216}
\DefMacro{Llama-3_lparentheses_name_split}{16}
\DefMacro{Llama-3_lparentheses_name_mixed}{18}
\DefMacro{Qwen25-Coder_lparentheses_name_all}{1537}
\DefMacro{Qwen25-Coder_lparentheses_name_unchanged}{1287}
\DefMacro{Qwen25-Coder_lparentheses_name_changed}{250}
\DefMacro{Qwen25-Coder_lparentheses_name_merged}{216}
\DefMacro{Qwen25-Coder_lparentheses_name_split}{16}
\DefMacro{Qwen25-Coder_lparentheses_name_mixed}{18}
\DefMacro{deepseek-coder_lparentheses_name_all}{1537}
\DefMacro{deepseek-coder_lparentheses_name_unchanged}{1097}
\DefMacro{deepseek-coder_lparentheses_name_changed}{440}
\DefMacro{deepseek-coder_lparentheses_name_merged}{138}
\DefMacro{deepseek-coder_lparentheses_name_split}{247}
\DefMacro{deepseek-coder_lparentheses_name_mixed}{55}
\DefMacro{CodeQwen15_lparentheses_name_all}{1537}
\DefMacro{CodeQwen15_lparentheses_name_unchanged}{1123}
\DefMacro{CodeQwen15_lparentheses_name_changed}{414}
\DefMacro{CodeQwen15_lparentheses_name_merged}{184}
\DefMacro{CodeQwen15_lparentheses_name_split}{174}
\DefMacro{CodeQwen15_lparentheses_name_mixed}{56}

\DefMacro{Llama-3_op_rsquarebracket_all}{287}
\DefMacro{Llama-3_op_rsquarebracket_unchanged}{257}
\DefMacro{Llama-3_op_rsquarebracket_changed}{30}
\DefMacro{Llama-3_op_rsquarebracket_merged}{15}
\DefMacro{Llama-3_op_rsquarebracket_split}{15}
\DefMacro{Llama-3_op_rsquarebracket_mixed}{0}
\DefMacro{Qwen25-Coder_op_rsquarebracket_all}{287}
\DefMacro{Qwen25-Coder_op_rsquarebracket_unchanged}{257}
\DefMacro{Qwen25-Coder_op_rsquarebracket_changed}{30}
\DefMacro{Qwen25-Coder_op_rsquarebracket_merged}{15}
\DefMacro{Qwen25-Coder_op_rsquarebracket_split}{15}
\DefMacro{Qwen25-Coder_op_rsquarebracket_mixed}{0}
\DefMacro{deepseek-coder_op_rsquarebracket_all}{287}
\DefMacro{deepseek-coder_op_rsquarebracket_unchanged}{230}
\DefMacro{deepseek-coder_op_rsquarebracket_changed}{57}
\DefMacro{deepseek-coder_op_rsquarebracket_merged}{6}
\DefMacro{deepseek-coder_op_rsquarebracket_split}{46}
\DefMacro{deepseek-coder_op_rsquarebracket_mixed}{5}
\DefMacro{CodeQwen15_op_rsquarebracket_all}{287}
\DefMacro{CodeQwen15_op_rsquarebracket_unchanged}{224}
\DefMacro{CodeQwen15_op_rsquarebracket_changed}{63}
\DefMacro{CodeQwen15_op_rsquarebracket_merged}{3}
\DefMacro{CodeQwen15_op_rsquarebracket_split}{56}
\DefMacro{CodeQwen15_op_rsquarebracket_mixed}{4}

\DefMacro{Llama-3_camel_case-snake_case_all}{403}
\DefMacro{Llama-3_camel_case-snake_case_unchanged}{287}
\DefMacro{Llama-3_camel_case-snake_case_changed}{116}
\DefMacro{Llama-3_camel_case-snake_case_merged}{16}
\DefMacro{Llama-3_camel_case-snake_case_split}{95}
\DefMacro{Llama-3_camel_case-snake_case_mixed}{5}
\DefMacro{Qwen25-Coder_camel_case-snake_case_all}{403}
\DefMacro{Qwen25-Coder_camel_case-snake_case_unchanged}{287}
\DefMacro{Qwen25-Coder_camel_case-snake_case_changed}{116}
\DefMacro{Qwen25-Coder_camel_case-snake_case_merged}{16}
\DefMacro{Qwen25-Coder_camel_case-snake_case_split}{95}
\DefMacro{Qwen25-Coder_camel_case-snake_case_mixed}{5}
\DefMacro{deepseek-coder_camel_case-snake_case_all}{406}
\DefMacro{deepseek-coder_camel_case-snake_case_unchanged}{291}
\DefMacro{deepseek-coder_camel_case-snake_case_changed}{115}
\DefMacro{deepseek-coder_camel_case-snake_case_merged}{90}
\DefMacro{deepseek-coder_camel_case-snake_case_split}{4}
\DefMacro{deepseek-coder_camel_case-snake_case_mixed}{21}
\DefMacro{CodeQwen15_camel_case-snake_case_all}{406}
\DefMacro{CodeQwen15_camel_case-snake_case_unchanged}{282}
\DefMacro{CodeQwen15_camel_case-snake_case_changed}{124}
\DefMacro{CodeQwen15_camel_case-snake_case_merged}{111}
\DefMacro{CodeQwen15_camel_case-snake_case_split}{3}
\DefMacro{CodeQwen15_camel_case-snake_case_mixed}{10}

\DefMacro{Llama-3_rparentheses_semicolon_all}{707}
\DefMacro{Llama-3_rparentheses_semicolon_unchanged}{550}
\DefMacro{Llama-3_rparentheses_semicolon_changed}{157}
\DefMacro{Llama-3_rparentheses_semicolon_merged}{2}
\DefMacro{Llama-3_rparentheses_semicolon_split}{150}
\DefMacro{Llama-3_rparentheses_semicolon_mixed}{5}
\DefMacro{Qwen25-Coder_rparentheses_semicolon_all}{707}
\DefMacro{Qwen25-Coder_rparentheses_semicolon_unchanged}{550}
\DefMacro{Qwen25-Coder_rparentheses_semicolon_changed}{157}
\DefMacro{Qwen25-Coder_rparentheses_semicolon_merged}{2}
\DefMacro{Qwen25-Coder_rparentheses_semicolon_split}{150}
\DefMacro{Qwen25-Coder_rparentheses_semicolon_mixed}{5}
\DefMacro{deepseek-coder_rparentheses_semicolon_all}{707}
\DefMacro{deepseek-coder_rparentheses_semicolon_unchanged}{685}
\DefMacro{deepseek-coder_rparentheses_semicolon_changed}{22}
\DefMacro{deepseek-coder_rparentheses_semicolon_merged}{12}
\DefMacro{deepseek-coder_rparentheses_semicolon_split}{1}
\DefMacro{deepseek-coder_rparentheses_semicolon_mixed}{9}
\DefMacro{CodeQwen15_rparentheses_semicolon_all}{707}
\DefMacro{CodeQwen15_rparentheses_semicolon_unchanged}{524}
\DefMacro{CodeQwen15_rparentheses_semicolon_changed}{183}
\DefMacro{CodeQwen15_rparentheses_semicolon_merged}{10}
\DefMacro{CodeQwen15_rparentheses_semicolon_split}{160}
\DefMacro{CodeQwen15_rparentheses_semicolon_mixed}{13}

\DefMacro{Llama-3_camel_case-pascal_case_all}{403}
\DefMacro{Llama-3_camel_case-pascal_case_unchanged}{295}
\DefMacro{Llama-3_camel_case-pascal_case_changed}{108}
\DefMacro{Llama-3_camel_case-pascal_case_merged}{1}
\DefMacro{Llama-3_camel_case-pascal_case_split}{79}
\DefMacro{Llama-3_camel_case-pascal_case_mixed}{28}
\DefMacro{Qwen25-Coder_camel_case-pascal_case_all}{403}
\DefMacro{Qwen25-Coder_camel_case-pascal_case_unchanged}{295}
\DefMacro{Qwen25-Coder_camel_case-pascal_case_changed}{108}
\DefMacro{Qwen25-Coder_camel_case-pascal_case_merged}{1}
\DefMacro{Qwen25-Coder_camel_case-pascal_case_split}{79}
\DefMacro{Qwen25-Coder_camel_case-pascal_case_mixed}{28}
\DefMacro{deepseek-coder_camel_case-pascal_case_all}{406}
\DefMacro{deepseek-coder_camel_case-pascal_case_unchanged}{246}
\DefMacro{deepseek-coder_camel_case-pascal_case_changed}{160}
\DefMacro{deepseek-coder_camel_case-pascal_case_merged}{17}
\DefMacro{deepseek-coder_camel_case-pascal_case_split}{129}
\DefMacro{deepseek-coder_camel_case-pascal_case_mixed}{14}
\DefMacro{CodeQwen15_camel_case-pascal_case_all}{406}
\DefMacro{CodeQwen15_camel_case-pascal_case_unchanged}{258}
\DefMacro{CodeQwen15_camel_case-pascal_case_changed}{148}
\DefMacro{CodeQwen15_camel_case-pascal_case_merged}{15}
\DefMacro{CodeQwen15_camel_case-pascal_case_split}{122}
\DefMacro{CodeQwen15_camel_case-pascal_case_mixed}{11}

\DefMacro{Llama-3_snake_case-screaming_snake_case_all}{343}
\DefMacro{Llama-3_snake_case-screaming_snake_case_unchanged}{112}
\DefMacro{Llama-3_snake_case-screaming_snake_case_changed}{231}
\DefMacro{Llama-3_snake_case-screaming_snake_case_merged}{0}
\DefMacro{Llama-3_snake_case-screaming_snake_case_split}{231}
\DefMacro{Llama-3_snake_case-screaming_snake_case_mixed}{0}
\DefMacro{Qwen25-Coder_snake_case-screaming_snake_case_all}{343}
\DefMacro{Qwen25-Coder_snake_case-screaming_snake_case_unchanged}{112}
\DefMacro{Qwen25-Coder_snake_case-screaming_snake_case_changed}{231}
\DefMacro{Qwen25-Coder_snake_case-screaming_snake_case_merged}{0}
\DefMacro{Qwen25-Coder_snake_case-screaming_snake_case_split}{231}
\DefMacro{Qwen25-Coder_snake_case-screaming_snake_case_mixed}{0}
\DefMacro{deepseek-coder_snake_case-screaming_snake_case_all}{343}
\DefMacro{deepseek-coder_snake_case-screaming_snake_case_unchanged}{30}
\DefMacro{deepseek-coder_snake_case-screaming_snake_case_changed}{313}
\DefMacro{deepseek-coder_snake_case-screaming_snake_case_merged}{0}
\DefMacro{deepseek-coder_snake_case-screaming_snake_case_split}{309}
\DefMacro{deepseek-coder_snake_case-screaming_snake_case_mixed}{4}
\DefMacro{CodeQwen15_snake_case-screaming_snake_case_all}{343}
\DefMacro{CodeQwen15_snake_case-screaming_snake_case_unchanged}{21}
\DefMacro{CodeQwen15_snake_case-screaming_snake_case_changed}{322}
\DefMacro{CodeQwen15_snake_case-screaming_snake_case_merged}{1}
\DefMacro{CodeQwen15_snake_case-screaming_snake_case_split}{308}
\DefMacro{CodeQwen15_snake_case-screaming_snake_case_mixed}{13}

\DefMacro{Llama-3_op_dash_all}{114}
\DefMacro{Llama-3_op_dash_unchanged}{94}
\DefMacro{Llama-3_op_dash_changed}{20}
\DefMacro{Llama-3_op_dash_merged}{8}
\DefMacro{Llama-3_op_dash_split}{11}
\DefMacro{Llama-3_op_dash_mixed}{1}
\DefMacro{Qwen25-Coder_op_dash_all}{114}
\DefMacro{Qwen25-Coder_op_dash_unchanged}{94}
\DefMacro{Qwen25-Coder_op_dash_changed}{20}
\DefMacro{Qwen25-Coder_op_dash_merged}{8}
\DefMacro{Qwen25-Coder_op_dash_split}{11}
\DefMacro{Qwen25-Coder_op_dash_mixed}{1}
\DefMacro{deepseek-coder_op_dash_all}{114}
\DefMacro{deepseek-coder_op_dash_unchanged}{107}
\DefMacro{deepseek-coder_op_dash_changed}{7}
\DefMacro{deepseek-coder_op_dash_merged}{7}
\DefMacro{deepseek-coder_op_dash_split}{0}
\DefMacro{deepseek-coder_op_dash_mixed}{0}
\DefMacro{CodeQwen15_op_dash_all}{114}
\DefMacro{CodeQwen15_op_dash_unchanged}{109}
\DefMacro{CodeQwen15_op_dash_changed}{5}
\DefMacro{CodeQwen15_op_dash_merged}{5}
\DefMacro{CodeQwen15_op_dash_split}{0}
\DefMacro{CodeQwen15_op_dash_mixed}{0}

\DefMacro{Llama-3_op_lsquarebracket_all}{155}
\DefMacro{Llama-3_op_lsquarebracket_unchanged}{145}
\DefMacro{Llama-3_op_lsquarebracket_changed}{10}
\DefMacro{Llama-3_op_lsquarebracket_merged}{7}
\DefMacro{Llama-3_op_lsquarebracket_split}{1}
\DefMacro{Llama-3_op_lsquarebracket_mixed}{2}
\DefMacro{Qwen25-Coder_op_lsquarebracket_all}{155}
\DefMacro{Qwen25-Coder_op_lsquarebracket_unchanged}{145}
\DefMacro{Qwen25-Coder_op_lsquarebracket_changed}{10}
\DefMacro{Qwen25-Coder_op_lsquarebracket_merged}{7}
\DefMacro{Qwen25-Coder_op_lsquarebracket_split}{1}
\DefMacro{Qwen25-Coder_op_lsquarebracket_mixed}{2}
\DefMacro{deepseek-coder_op_lsquarebracket_all}{155}
\DefMacro{deepseek-coder_op_lsquarebracket_unchanged}{141}
\DefMacro{deepseek-coder_op_lsquarebracket_changed}{14}
\DefMacro{deepseek-coder_op_lsquarebracket_merged}{3}
\DefMacro{deepseek-coder_op_lsquarebracket_split}{6}
\DefMacro{deepseek-coder_op_lsquarebracket_mixed}{5}
\DefMacro{CodeQwen15_op_lsquarebracket_all}{155}
\DefMacro{CodeQwen15_op_lsquarebracket_unchanged}{125}
\DefMacro{CodeQwen15_op_lsquarebracket_changed}{30}
\DefMacro{CodeQwen15_op_lsquarebracket_merged}{22}
\DefMacro{CodeQwen15_op_lsquarebracket_split}{6}
\DefMacro{CodeQwen15_op_lsquarebracket_mixed}{2}

\DefMacro{Llama-3_lparentheses_rparentheses_all}{1185}
\DefMacro{Llama-3_lparentheses_rparentheses_unchanged}{982}
\DefMacro{Llama-3_lparentheses_rparentheses_changed}{203}
\DefMacro{Llama-3_lparentheses_rparentheses_merged}{11}
\DefMacro{Llama-3_lparentheses_rparentheses_split}{169}
\DefMacro{Llama-3_lparentheses_rparentheses_mixed}{23}
\DefMacro{Qwen25-Coder_lparentheses_rparentheses_all}{1185}
\DefMacro{Qwen25-Coder_lparentheses_rparentheses_unchanged}{982}
\DefMacro{Qwen25-Coder_lparentheses_rparentheses_changed}{203}
\DefMacro{Qwen25-Coder_lparentheses_rparentheses_merged}{11}
\DefMacro{Qwen25-Coder_lparentheses_rparentheses_split}{169}
\DefMacro{Qwen25-Coder_lparentheses_rparentheses_mixed}{23}
\DefMacro{deepseek-coder_lparentheses_rparentheses_all}{1185}
\DefMacro{deepseek-coder_lparentheses_rparentheses_unchanged}{519}
\DefMacro{deepseek-coder_lparentheses_rparentheses_changed}{666}
\DefMacro{deepseek-coder_lparentheses_rparentheses_merged}{0}
\DefMacro{deepseek-coder_lparentheses_rparentheses_split}{637}
\DefMacro{deepseek-coder_lparentheses_rparentheses_mixed}{29}
\DefMacro{CodeQwen15_lparentheses_rparentheses_all}{1185}
\DefMacro{CodeQwen15_lparentheses_rparentheses_unchanged}{599}
\DefMacro{CodeQwen15_lparentheses_rparentheses_changed}{586}
\DefMacro{CodeQwen15_lparentheses_rparentheses_merged}{9}
\DefMacro{CodeQwen15_lparentheses_rparentheses_split}{540}
\DefMacro{CodeQwen15_lparentheses_rparentheses_mixed}{37}

\DefMacro{Llama-3_rparentheses_colon_all}{587}
\DefMacro{Llama-3_rparentheses_colon_unchanged}{570}
\DefMacro{Llama-3_rparentheses_colon_changed}{17}
\DefMacro{Llama-3_rparentheses_colon_merged}{15}
\DefMacro{Llama-3_rparentheses_colon_split}{2}
\DefMacro{Llama-3_rparentheses_colon_mixed}{0}
\DefMacro{Qwen25-Coder_rparentheses_colon_all}{587}
\DefMacro{Qwen25-Coder_rparentheses_colon_unchanged}{570}
\DefMacro{Qwen25-Coder_rparentheses_colon_changed}{17}
\DefMacro{Qwen25-Coder_rparentheses_colon_merged}{15}
\DefMacro{Qwen25-Coder_rparentheses_colon_split}{2}
\DefMacro{Qwen25-Coder_rparentheses_colon_mixed}{0}
\DefMacro{deepseek-coder_rparentheses_colon_all}{587}
\DefMacro{deepseek-coder_rparentheses_colon_unchanged}{542}
\DefMacro{deepseek-coder_rparentheses_colon_changed}{45}
\DefMacro{deepseek-coder_rparentheses_colon_merged}{34}
\DefMacro{deepseek-coder_rparentheses_colon_split}{0}
\DefMacro{deepseek-coder_rparentheses_colon_mixed}{11}
\DefMacro{CodeQwen15_rparentheses_colon_all}{587}
\DefMacro{CodeQwen15_rparentheses_colon_unchanged}{562}
\DefMacro{CodeQwen15_rparentheses_colon_changed}{25}
\DefMacro{CodeQwen15_rparentheses_colon_merged}{20}
\DefMacro{CodeQwen15_rparentheses_colon_split}{0}
\DefMacro{CodeQwen15_rparentheses_colon_mixed}{5}

\DefMacro{Llama-3_camel_case-screaming_snake_case_all}{403}
\DefMacro{Llama-3_camel_case-screaming_snake_case_unchanged}{126}
\DefMacro{Llama-3_camel_case-screaming_snake_case_changed}{277}
\DefMacro{Llama-3_camel_case-screaming_snake_case_merged}{0}
\DefMacro{Llama-3_camel_case-screaming_snake_case_split}{274}
\DefMacro{Llama-3_camel_case-screaming_snake_case_mixed}{3}
\DefMacro{Qwen25-Coder_camel_case-screaming_snake_case_all}{403}
\DefMacro{Qwen25-Coder_camel_case-screaming_snake_case_unchanged}{126}
\DefMacro{Qwen25-Coder_camel_case-screaming_snake_case_changed}{277}
\DefMacro{Qwen25-Coder_camel_case-screaming_snake_case_merged}{0}
\DefMacro{Qwen25-Coder_camel_case-screaming_snake_case_split}{274}
\DefMacro{Qwen25-Coder_camel_case-screaming_snake_case_mixed}{3}
\DefMacro{deepseek-coder_camel_case-screaming_snake_case_all}{406}
\DefMacro{deepseek-coder_camel_case-screaming_snake_case_unchanged}{50}
\DefMacro{deepseek-coder_camel_case-screaming_snake_case_changed}{356}
\DefMacro{deepseek-coder_camel_case-screaming_snake_case_merged}{3}
\DefMacro{deepseek-coder_camel_case-screaming_snake_case_split}{324}
\DefMacro{deepseek-coder_camel_case-screaming_snake_case_mixed}{29}
\DefMacro{CodeQwen15_camel_case-screaming_snake_case_all}{406}
\DefMacro{CodeQwen15_camel_case-screaming_snake_case_unchanged}{38}
\DefMacro{CodeQwen15_camel_case-screaming_snake_case_changed}{368}
\DefMacro{CodeQwen15_camel_case-screaming_snake_case_merged}{0}
\DefMacro{CodeQwen15_camel_case-screaming_snake_case_split}{339}
\DefMacro{CodeQwen15_camel_case-screaming_snake_case_mixed}{29}

\DefMacro{Llama-3_op_semicolon_all}{774}
\DefMacro{Llama-3_op_semicolon_unchanged}{596}
\DefMacro{Llama-3_op_semicolon_changed}{178}
\DefMacro{Llama-3_op_semicolon_merged}{5}
\DefMacro{Llama-3_op_semicolon_split}{167}
\DefMacro{Llama-3_op_semicolon_mixed}{6}
\DefMacro{Qwen25-Coder_op_semicolon_all}{774}
\DefMacro{Qwen25-Coder_op_semicolon_unchanged}{596}
\DefMacro{Qwen25-Coder_op_semicolon_changed}{178}
\DefMacro{Qwen25-Coder_op_semicolon_merged}{5}
\DefMacro{Qwen25-Coder_op_semicolon_split}{167}
\DefMacro{Qwen25-Coder_op_semicolon_mixed}{6}
\DefMacro{deepseek-coder_op_semicolon_all}{774}
\DefMacro{deepseek-coder_op_semicolon_unchanged}{737}
\DefMacro{deepseek-coder_op_semicolon_changed}{37}
\DefMacro{deepseek-coder_op_semicolon_merged}{24}
\DefMacro{deepseek-coder_op_semicolon_split}{0}
\DefMacro{deepseek-coder_op_semicolon_mixed}{13}
\DefMacro{CodeQwen15_op_semicolon_all}{774}
\DefMacro{CodeQwen15_op_semicolon_unchanged}{583}
\DefMacro{CodeQwen15_op_semicolon_changed}{191}
\DefMacro{CodeQwen15_op_semicolon_merged}{18}
\DefMacro{CodeQwen15_op_semicolon_split}{156}
\DefMacro{CodeQwen15_op_semicolon_mixed}{17}

\DefMacro{Llama-3_rsquarebracket_rparentheses_all}{243}
\DefMacro{Llama-3_rsquarebracket_rparentheses_unchanged}{216}
\DefMacro{Llama-3_rsquarebracket_rparentheses_changed}{27}
\DefMacro{Llama-3_rsquarebracket_rparentheses_merged}{14}
\DefMacro{Llama-3_rsquarebracket_rparentheses_split}{11}
\DefMacro{Llama-3_rsquarebracket_rparentheses_mixed}{2}
\DefMacro{Qwen25-Coder_rsquarebracket_rparentheses_all}{243}
\DefMacro{Qwen25-Coder_rsquarebracket_rparentheses_unchanged}{216}
\DefMacro{Qwen25-Coder_rsquarebracket_rparentheses_changed}{27}
\DefMacro{Qwen25-Coder_rsquarebracket_rparentheses_merged}{14}
\DefMacro{Qwen25-Coder_rsquarebracket_rparentheses_split}{11}
\DefMacro{Qwen25-Coder_rsquarebracket_rparentheses_mixed}{2}
\DefMacro{deepseek-coder_rsquarebracket_rparentheses_all}{243}
\DefMacro{deepseek-coder_rsquarebracket_rparentheses_unchanged}{165}
\DefMacro{deepseek-coder_rsquarebracket_rparentheses_changed}{78}
\DefMacro{deepseek-coder_rsquarebracket_rparentheses_merged}{22}
\DefMacro{deepseek-coder_rsquarebracket_rparentheses_split}{53}
\DefMacro{deepseek-coder_rsquarebracket_rparentheses_mixed}{3}
\DefMacro{CodeQwen15_rsquarebracket_rparentheses_all}{243}
\DefMacro{CodeQwen15_rsquarebracket_rparentheses_unchanged}{165}
\DefMacro{CodeQwen15_rsquarebracket_rparentheses_changed}{78}
\DefMacro{CodeQwen15_rsquarebracket_rparentheses_merged}{23}
\DefMacro{CodeQwen15_rsquarebracket_rparentheses_split}{53}
\DefMacro{CodeQwen15_rsquarebracket_rparentheses_mixed}{2}

\DefMacro{Llama-3_op_all_all}{1546}
\DefMacro{Llama-3_op_all_unchanged}{827}
\DefMacro{Llama-3_op_all_changed}{719}
\DefMacro{Llama-3_op_all_merged}{154}
\DefMacro{Llama-3_op_all_split}{396}
\DefMacro{Llama-3_op_all_mixed}{169}
\DefMacro{Qwen25-Coder_op_all_all}{1546}
\DefMacro{Qwen25-Coder_op_all_unchanged}{827}
\DefMacro{Qwen25-Coder_op_all_changed}{719}
\DefMacro{Qwen25-Coder_op_all_merged}{154}
\DefMacro{Qwen25-Coder_op_all_split}{395}
\DefMacro{Qwen25-Coder_op_all_mixed}{170}
\DefMacro{deepseek-coder_op_all_all}{1546}
\DefMacro{deepseek-coder_op_all_unchanged}{711}
\DefMacro{deepseek-coder_op_all_changed}{835}
\DefMacro{deepseek-coder_op_all_merged}{142}
\DefMacro{deepseek-coder_op_all_split}{469}
\DefMacro{deepseek-coder_op_all_mixed}{224}
\DefMacro{CodeQwen15_op_all_all}{1546}
\DefMacro{CodeQwen15_op_all_unchanged}{694}
\DefMacro{CodeQwen15_op_all_changed}{852}
\DefMacro{CodeQwen15_op_all_merged}{174}
\DefMacro{CodeQwen15_op_all_split}{393}
\DefMacro{CodeQwen15_op_all_mixed}{285}

\DefMacro{Llama-3_rparentheses_rparentheses_all}{992}
\DefMacro{Llama-3_rparentheses_rparentheses_unchanged}{943}
\DefMacro{Llama-3_rparentheses_rparentheses_changed}{49}
\DefMacro{Llama-3_rparentheses_rparentheses_merged}{40}
\DefMacro{Llama-3_rparentheses_rparentheses_split}{5}
\DefMacro{Llama-3_rparentheses_rparentheses_mixed}{4}
\DefMacro{Qwen25-Coder_rparentheses_rparentheses_all}{992}
\DefMacro{Qwen25-Coder_rparentheses_rparentheses_unchanged}{943}
\DefMacro{Qwen25-Coder_rparentheses_rparentheses_changed}{49}
\DefMacro{Qwen25-Coder_rparentheses_rparentheses_merged}{40}
\DefMacro{Qwen25-Coder_rparentheses_rparentheses_split}{5}
\DefMacro{Qwen25-Coder_rparentheses_rparentheses_mixed}{4}
\DefMacro{deepseek-coder_rparentheses_rparentheses_all}{992}
\DefMacro{deepseek-coder_rparentheses_rparentheses_unchanged}{873}
\DefMacro{deepseek-coder_rparentheses_rparentheses_changed}{119}
\DefMacro{deepseek-coder_rparentheses_rparentheses_merged}{34}
\DefMacro{deepseek-coder_rparentheses_rparentheses_split}{12}
\DefMacro{deepseek-coder_rparentheses_rparentheses_mixed}{73}
\DefMacro{CodeQwen15_rparentheses_rparentheses_all}{992}
\DefMacro{CodeQwen15_rparentheses_rparentheses_unchanged}{948}
\DefMacro{CodeQwen15_rparentheses_rparentheses_changed}{44}
\DefMacro{CodeQwen15_rparentheses_rparentheses_merged}{26}
\DefMacro{CodeQwen15_rparentheses_rparentheses_split}{12}
\DefMacro{CodeQwen15_rparentheses_rparentheses_mixed}{6}

\DefMacro{Llama-3_naming_all}{2238}
\DefMacro{Llama-3_naming_unchanged}{1292}
\DefMacro{Llama-3_naming_changed}{946}
\DefMacro{Llama-3_naming_merged}{105}
\DefMacro{Llama-3_naming_split}{767}
\DefMacro{Llama-3_naming_mixed}{74}
\DefMacro{Llama-3_spacing_all}{12804}
\DefMacro{Llama-3_spacing_unchanged}{9315}
\DefMacro{Llama-3_spacing_changed}{3489}
\DefMacro{Llama-3_spacing_merged}{660}
\DefMacro{Llama-3_spacing_split}{2394}
\DefMacro{Llama-3_spacing_mixed}{435}

\DefMacro{Qwen25-Coder_naming_all}{2238}
\DefMacro{Qwen25-Coder_naming_unchanged}{1292}
\DefMacro{Qwen25-Coder_naming_changed}{946}
\DefMacro{Qwen25-Coder_naming_merged}{105}
\DefMacro{Qwen25-Coder_naming_split}{767}
\DefMacro{Qwen25-Coder_naming_mixed}{74}
\DefMacro{Qwen25-Coder_spacing_all}{12804}
\DefMacro{Qwen25-Coder_spacing_unchanged}{9315}
\DefMacro{Qwen25-Coder_spacing_changed}{3489}
\DefMacro{Qwen25-Coder_spacing_merged}{660}
\DefMacro{Qwen25-Coder_spacing_split}{2391}
\DefMacro{Qwen25-Coder_spacing_mixed}{438}

\DefMacro{deepseek-coder_naming_all}{2247}
\DefMacro{deepseek-coder_naming_unchanged}{999}
\DefMacro{deepseek-coder_naming_changed}{1248}
\DefMacro{deepseek-coder_naming_merged}{123}
\DefMacro{deepseek-coder_naming_split}{996}
\DefMacro{deepseek-coder_naming_mixed}{129}
\DefMacro{deepseek-coder_spacing_all}{12804}
\DefMacro{deepseek-coder_spacing_unchanged}{8381}
\DefMacro{deepseek-coder_spacing_changed}{4423}
\DefMacro{deepseek-coder_spacing_merged}{608}
\DefMacro{deepseek-coder_spacing_split}{3091}
\DefMacro{deepseek-coder_spacing_mixed}{724}

\DefMacro{CodeQwen15_naming_all}{2247}
\DefMacro{CodeQwen15_naming_unchanged}{954}
\DefMacro{CodeQwen15_naming_changed}{1293}
\DefMacro{CodeQwen15_naming_merged}{136}
\DefMacro{CodeQwen15_naming_split}{1043}
\DefMacro{CodeQwen15_naming_mixed}{114}
\DefMacro{CodeQwen15_spacing_all}{12804}
\DefMacro{CodeQwen15_spacing_unchanged}{8720}
\DefMacro{CodeQwen15_spacing_changed}{4084}
\DefMacro{CodeQwen15_spacing_merged}{725}
\DefMacro{CodeQwen15_spacing_split}{2504}
\DefMacro{CodeQwen15_spacing_mixed}{855}

\DefMacro{llama-medium_java_baseline}{43.15}
\DefMacro{llama-medium_java_camel-case-snake-case}{43.54}
\DefMacro{llama-medium_java_camel-case-snake-case_delta}{+0.39}
\DefMacro{llama-medium_java_camel-case-snake-case_trend}{$\uparrow$}
\DefMacro{llama-medium_java_camel-case-pascal-case}{43.54}
\DefMacro{llama-medium_java_camel-case-pascal-case_delta}{+0.39}
\DefMacro{llama-medium_java_camel-case-pascal-case_trend}{$\uparrow$}
\DefMacro{llama-medium_java_camel-case-screaming-snake-case}{44.19}
\DefMacro{llama-medium_java_camel-case-screaming-snake-case_delta}{+1.04}
\DefMacro{llama-medium_java_camel-case-screaming-snake-case_trend}{$\uparrow$}
\DefMacro{llama-medium_java_rparentheses-semicolon}{44.57}
\DefMacro{llama-medium_java_rparentheses-semicolon_delta}{+1.42}
\DefMacro{llama-medium_java_rparentheses-semicolon_trend}{$\uparrow$}
\DefMacro{llama-medium_java_period-name}{36.82}
\DefMacro{llama-medium_java_period-name_delta}{-6.33}
\DefMacro{llama-medium_java_period-name_trend}{$\downarrow$}
\DefMacro{llama-medium_java_lparentheses-rparentheses}{41.09}
\DefMacro{llama-medium_java_lparentheses-rparentheses_delta}{-2.06}
\DefMacro{llama-medium_java_lparentheses-rparentheses_trend}{$\downarrow$}
\DefMacro{llama-medium_java_lparentheses-name}{40.83}
\DefMacro{llama-medium_java_lparentheses-name_delta}{-2.32}
\DefMacro{llama-medium_java_lparentheses-name_trend}{$\downarrow$}
\DefMacro{llama-medium_java_rparentheses-rparentheses}{42.64}
\DefMacro{llama-medium_java_rparentheses-rparentheses_delta}{-0.51}
\DefMacro{llama-medium_java_rparentheses-rparentheses_trend}{$\downarrow$}
\DefMacro{llama-medium_java_period-asterisk}{40.96}
\DefMacro{llama-medium_java_period-asterisk_delta}{-2.19}
\DefMacro{llama-medium_java_period-asterisk_trend}{$\downarrow$}
\DefMacro{llama-medium_java_double-plus-rparentheses}{43.28}
\DefMacro{llama-medium_java_double-plus-rparentheses_delta}{+0.13}
\DefMacro{llama-medium_java_double-plus-rparentheses_trend}{$\uparrow$}
\DefMacro{llama-medium_java_rparentheses-period}{43.02}
\DefMacro{llama-medium_java_rparentheses-period_delta}{-0.13}
\DefMacro{llama-medium_java_rparentheses-period_trend}{$\downarrow$}
\DefMacro{llama-medium_java_op-semicolon}{43.02}
\DefMacro{llama-medium_java_op-semicolon_delta}{-0.13}
\DefMacro{llama-medium_java_op-semicolon_trend}{$\downarrow$}
\DefMacro{llama-medium_java_op-lparentheses}{43.02}
\DefMacro{llama-medium_java_op-lparentheses_delta}{-0.13}
\DefMacro{llama-medium_java_op-lparentheses_trend}{$\downarrow$}
\DefMacro{llama-medium_java_op-name}{37.34}
\DefMacro{llama-medium_java_op-name_delta}{-5.81}
\DefMacro{llama-medium_java_op-name_trend}{$\downarrow$}
\DefMacro{llama-medium_java_op-all}{34.88}
\DefMacro{llama-medium_java_op-all_delta}{-8.27}
\DefMacro{llama-medium_java_op-all_trend}{$\downarrow$}

\DefMacro{llama-small_java_baseline}{32.04}
\DefMacro{llama-small_java_camel-case-snake-case}{32.69}
\DefMacro{llama-small_java_camel-case-snake-case_delta}{+0.65}
\DefMacro{llama-small_java_camel-case-snake-case_trend}{$\uparrow$}
\DefMacro{llama-small_java_camel-case-pascal-case}{32.17}
\DefMacro{llama-small_java_camel-case-pascal-case_delta}{+0.13}
\DefMacro{llama-small_java_camel-case-pascal-case_trend}{$\uparrow$}
\DefMacro{llama-small_java_camel-case-screaming-snake-case}{32.56}
\DefMacro{llama-small_java_camel-case-screaming-snake-case_delta}{+0.52}
\DefMacro{llama-small_java_camel-case-screaming-snake-case_trend}{$\uparrow$}
\DefMacro{llama-small_java_rparentheses-semicolon}{32.69}
\DefMacro{llama-small_java_rparentheses-semicolon_delta}{+0.65}
\DefMacro{llama-small_java_rparentheses-semicolon_trend}{$\uparrow$}
\DefMacro{llama-small_java_period-name}{30.62}
\DefMacro{llama-small_java_period-name_delta}{-1.42}
\DefMacro{llama-small_java_period-name_trend}{$\downarrow$}
\DefMacro{llama-small_java_lparentheses-rparentheses}{29.84}
\DefMacro{llama-small_java_lparentheses-rparentheses_delta}{-2.20}
\DefMacro{llama-small_java_lparentheses-rparentheses_trend}{$\downarrow$}
\DefMacro{llama-small_java_lparentheses-name}{30.88}
\DefMacro{llama-small_java_lparentheses-name_delta}{-1.16}
\DefMacro{llama-small_java_lparentheses-name_trend}{$\downarrow$}
\DefMacro{llama-small_java_rparentheses-rparentheses}{32.43}
\DefMacro{llama-small_java_rparentheses-rparentheses_delta}{+0.39}
\DefMacro{llama-small_java_rparentheses-rparentheses_trend}{$\uparrow$}
\DefMacro{llama-small_java_period-asterisk}{32.30}
\DefMacro{llama-small_java_period-asterisk_delta}{+0.26}
\DefMacro{llama-small_java_period-asterisk_trend}{$\uparrow$}
\DefMacro{llama-small_java_double-plus-rparentheses}{31.91}
\DefMacro{llama-small_java_double-plus-rparentheses_delta}{-0.13}
\DefMacro{llama-small_java_double-plus-rparentheses_trend}{$\downarrow$}
\DefMacro{llama-small_java_rparentheses-period}{31.65}
\DefMacro{llama-small_java_rparentheses-period_delta}{-0.39}
\DefMacro{llama-small_java_rparentheses-period_trend}{$\downarrow$}
\DefMacro{llama-small_java_op-semicolon}{30.49}
\DefMacro{llama-small_java_op-semicolon_delta}{-1.55}
\DefMacro{llama-small_java_op-semicolon_trend}{$\downarrow$}
\DefMacro{llama-small_java_op-lparentheses}{31.52}
\DefMacro{llama-small_java_op-lparentheses_delta}{-0.52}
\DefMacro{llama-small_java_op-lparentheses_trend}{$\downarrow$}
\DefMacro{llama-small_java_op-name}{28.68}
\DefMacro{llama-small_java_op-name_delta}{-3.36}
\DefMacro{llama-small_java_op-name_trend}{$\downarrow$}
\DefMacro{llama-small_java_op-all}{25.97}
\DefMacro{llama-small_java_op-all_delta}{-6.07}
\DefMacro{llama-small_java_op-all_trend}{$\downarrow$}

\DefMacro{llama-large_java_baseline}{57.24}
\DefMacro{llama-large_java_camel-case-snake-case}{57.49}
\DefMacro{llama-large_java_camel-case-snake-case_delta}{+0.25}
\DefMacro{llama-large_java_camel-case-snake-case_trend}{$\uparrow$}
\DefMacro{llama-large_java_camel-case-pascal-case}{56.85}
\DefMacro{llama-large_java_camel-case-pascal-case_delta}{-0.39}
\DefMacro{llama-large_java_camel-case-pascal-case_trend}{$\downarrow$}
\DefMacro{llama-large_java_camel-case-screaming-snake-case}{56.20}
\DefMacro{llama-large_java_camel-case-screaming-snake-case_delta}{-1.04}
\DefMacro{llama-large_java_camel-case-screaming-snake-case_trend}{$\downarrow$}
\DefMacro{llama-large_java_rparentheses-semicolon}{55.17}
\DefMacro{llama-large_java_rparentheses-semicolon_delta}{-2.07}
\DefMacro{llama-large_java_rparentheses-semicolon_trend}{$\downarrow$}
\DefMacro{llama-large_java_period-name}{57.24}
\DefMacro{llama-large_java_period-name_delta}{+0.00}
\DefMacro{llama-large_java_period-name_trend}{}
\DefMacro{llama-large_java_lparentheses-rparentheses}{54.13}
\DefMacro{llama-large_java_lparentheses-rparentheses_delta}{-3.11}
\DefMacro{llama-large_java_lparentheses-rparentheses_trend}{$\downarrow$}
\DefMacro{llama-large_java_lparentheses-name}{55.94}
\DefMacro{llama-large_java_lparentheses-name_delta}{-1.30}
\DefMacro{llama-large_java_lparentheses-name_trend}{$\downarrow$}
\DefMacro{llama-large_java_rparentheses-rparentheses}{56.59}
\DefMacro{llama-large_java_rparentheses-rparentheses_delta}{-0.65}
\DefMacro{llama-large_java_rparentheses-rparentheses_trend}{$\downarrow$}
\DefMacro{llama-large_java_period-asterisk}{58.66}
\DefMacro{llama-large_java_period-asterisk_delta}{+1.42}
\DefMacro{llama-large_java_period-asterisk_trend}{$\uparrow$}
\DefMacro{llama-large_java_double-plus-rparentheses}{57.24}
\DefMacro{llama-large_java_double-plus-rparentheses_delta}{+0.00}
\DefMacro{llama-large_java_double-plus-rparentheses_trend}{}
\DefMacro{llama-large_java_rparentheses-period}{56.20}
\DefMacro{llama-large_java_rparentheses-period_delta}{-1.04}
\DefMacro{llama-large_java_rparentheses-period_trend}{$\downarrow$}
\DefMacro{llama-large_java_op-semicolon}{56.07}
\DefMacro{llama-large_java_op-semicolon_delta}{-1.17}
\DefMacro{llama-large_java_op-semicolon_trend}{$\downarrow$}
\DefMacro{llama-large_java_op-lparentheses}{57.62}
\DefMacro{llama-large_java_op-lparentheses_delta}{+0.38}
\DefMacro{llama-large_java_op-lparentheses_trend}{$\uparrow$}
\DefMacro{llama-large_java_op-name}{56.07}
\DefMacro{llama-large_java_op-name_delta}{-1.17}
\DefMacro{llama-large_java_op-name_trend}{$\downarrow$}
\DefMacro{llama-large_java_op-all}{56.85}
\DefMacro{llama-large_java_op-all_delta}{-0.39}
\DefMacro{llama-large_java_op-all_trend}{$\downarrow$}

\DefMacro{qwen-small_java_baseline}{33.59}
\DefMacro{qwen-small_java_camel-case-snake-case}{35.27}
\DefMacro{qwen-small_java_camel-case-snake-case_delta}{+1.68}
\DefMacro{qwen-small_java_camel-case-snake-case_trend}{$\uparrow$}
\DefMacro{qwen-small_java_camel-case-pascal-case}{35.27}
\DefMacro{qwen-small_java_camel-case-pascal-case_delta}{+1.68}
\DefMacro{qwen-small_java_camel-case-pascal-case_trend}{$\uparrow$}
\DefMacro{qwen-small_java_camel-case-screaming-snake-case}{35.53}
\DefMacro{qwen-small_java_camel-case-screaming-snake-case_delta}{+1.94}
\DefMacro{qwen-small_java_camel-case-screaming-snake-case_trend}{$\uparrow$}
\DefMacro{qwen-small_java_rparentheses-semicolon}{35.14}
\DefMacro{qwen-small_java_rparentheses-semicolon_delta}{+1.55}
\DefMacro{qwen-small_java_rparentheses-semicolon_trend}{$\uparrow$}
\DefMacro{qwen-small_java_period-name}{33.46}
\DefMacro{qwen-small_java_period-name_delta}{-0.13}
\DefMacro{qwen-small_java_period-name_trend}{$\downarrow$}
\DefMacro{qwen-small_java_lparentheses-rparentheses}{32.17}
\DefMacro{qwen-small_java_lparentheses-rparentheses_delta}{-1.42}
\DefMacro{qwen-small_java_lparentheses-rparentheses_trend}{$\downarrow$}
\DefMacro{qwen-small_java_lparentheses-name}{34.88}
\DefMacro{qwen-small_java_lparentheses-name_delta}{+1.29}
\DefMacro{qwen-small_java_lparentheses-name_trend}{$\uparrow$}
\DefMacro{qwen-small_java_rparentheses-rparentheses}{33.46}
\DefMacro{qwen-small_java_rparentheses-rparentheses_delta}{-0.13}
\DefMacro{qwen-small_java_rparentheses-rparentheses_trend}{$\downarrow$}
\DefMacro{qwen-small_java_period-asterisk}{33.46}
\DefMacro{qwen-small_java_period-asterisk_delta}{-0.13}
\DefMacro{qwen-small_java_period-asterisk_trend}{$\downarrow$}
\DefMacro{qwen-small_java_double-plus-rparentheses}{34.11}
\DefMacro{qwen-small_java_double-plus-rparentheses_delta}{+0.52}
\DefMacro{qwen-small_java_double-plus-rparentheses_trend}{$\uparrow$}
\DefMacro{qwen-small_java_rparentheses-period}{34.37}
\DefMacro{qwen-small_java_rparentheses-period_delta}{+0.78}
\DefMacro{qwen-small_java_rparentheses-period_trend}{$\uparrow$}
\DefMacro{qwen-small_java_op-semicolon}{34.75}
\DefMacro{qwen-small_java_op-semicolon_delta}{+1.16}
\DefMacro{qwen-small_java_op-semicolon_trend}{$\uparrow$}
\DefMacro{qwen-small_java_op-lparentheses}{33.20}
\DefMacro{qwen-small_java_op-lparentheses_delta}{-0.39}
\DefMacro{qwen-small_java_op-lparentheses_trend}{$\downarrow$}
\DefMacro{qwen-small_java_op-name}{35.66}
\DefMacro{qwen-small_java_op-name_delta}{+2.07}
\DefMacro{qwen-small_java_op-name_trend}{$\uparrow$}
\DefMacro{qwen-small_java_op-all}{34.11}
\DefMacro{qwen-small_java_op-all_delta}{+0.52}
\DefMacro{qwen-small_java_op-all_trend}{$\uparrow$}

\DefMacro{qwen-large_java_baseline}{70.41}
\DefMacro{qwen-large_java_camel-case-snake-case}{70.28}
\DefMacro{qwen-large_java_camel-case-snake-case_delta}{-0.13}
\DefMacro{qwen-large_java_camel-case-snake-case_trend}{$\downarrow$}
\DefMacro{qwen-large_java_camel-case-pascal-case}{70.41}
\DefMacro{qwen-large_java_camel-case-pascal-case_delta}{+0.00}
\DefMacro{qwen-large_java_camel-case-pascal-case_trend}{}
\DefMacro{qwen-large_java_camel-case-screaming-snake-case}{69.12}
\DefMacro{qwen-large_java_camel-case-screaming-snake-case_delta}{-1.29}
\DefMacro{qwen-large_java_camel-case-screaming-snake-case_trend}{$\downarrow$}
\DefMacro{qwen-large_java_rparentheses-semicolon}{71.58}
\DefMacro{qwen-large_java_rparentheses-semicolon_delta}{+1.17}
\DefMacro{qwen-large_java_rparentheses-semicolon_trend}{$\uparrow$}
\DefMacro{qwen-large_java_period-name}{70.28}
\DefMacro{qwen-large_java_period-name_delta}{-0.13}
\DefMacro{qwen-large_java_period-name_trend}{$\downarrow$}
\DefMacro{qwen-large_java_lparentheses-rparentheses}{71.19}
\DefMacro{qwen-large_java_lparentheses-rparentheses_delta}{+0.78}
\DefMacro{qwen-large_java_lparentheses-rparentheses_trend}{$\uparrow$}
\DefMacro{qwen-large_java_lparentheses-name}{71.96}
\DefMacro{qwen-large_java_lparentheses-name_delta}{+1.55}
\DefMacro{qwen-large_java_lparentheses-name_trend}{$\uparrow$}
\DefMacro{qwen-large_java_rparentheses-rparentheses}{69.77}
\DefMacro{qwen-large_java_rparentheses-rparentheses_delta}{-0.64}
\DefMacro{qwen-large_java_rparentheses-rparentheses_trend}{$\downarrow$}
\DefMacro{qwen-large_java_period-asterisk}{69.51}
\DefMacro{qwen-large_java_period-asterisk_delta}{-0.90}
\DefMacro{qwen-large_java_period-asterisk_trend}{$\downarrow$}
\DefMacro{qwen-large_java_double-plus-rparentheses}{71.45}
\DefMacro{qwen-large_java_double-plus-rparentheses_delta}{+1.04}
\DefMacro{qwen-large_java_double-plus-rparentheses_trend}{$\uparrow$}
\DefMacro{qwen-large_java_rparentheses-period}{70.41}
\DefMacro{qwen-large_java_rparentheses-period_delta}{+0.00}
\DefMacro{qwen-large_java_rparentheses-period_trend}{}
\DefMacro{qwen-large_java_op-semicolon}{67.05}
\DefMacro{qwen-large_java_op-semicolon_delta}{-3.36}
\DefMacro{qwen-large_java_op-semicolon_trend}{$\downarrow$}
\DefMacro{qwen-large_java_op-lparentheses}{70.28}
\DefMacro{qwen-large_java_op-lparentheses_delta}{-0.13}
\DefMacro{qwen-large_java_op-lparentheses_trend}{$\downarrow$}
\DefMacro{qwen-large_java_op-name}{70.03}
\DefMacro{qwen-large_java_op-name_delta}{-0.38}
\DefMacro{qwen-large_java_op-name_trend}{$\downarrow$}
\DefMacro{qwen-large_java_op-all}{70.28}
\DefMacro{qwen-large_java_op-all_delta}{-0.13}
\DefMacro{qwen-large_java_op-all_trend}{$\downarrow$}

\DefMacro{qwen-medium_java_baseline}{57.36}
\DefMacro{qwen-medium_java_camel-case-snake-case}{57.62}
\DefMacro{qwen-medium_java_camel-case-snake-case_delta}{+0.26}
\DefMacro{qwen-medium_java_camel-case-snake-case_trend}{$\uparrow$}
\DefMacro{qwen-medium_java_camel-case-pascal-case}{57.75}
\DefMacro{qwen-medium_java_camel-case-pascal-case_delta}{+0.39}
\DefMacro{qwen-medium_java_camel-case-pascal-case_trend}{$\uparrow$}
\DefMacro{qwen-medium_java_camel-case-screaming-snake-case}{58.01}
\DefMacro{qwen-medium_java_camel-case-screaming-snake-case_delta}{+0.65}
\DefMacro{qwen-medium_java_camel-case-screaming-snake-case_trend}{$\uparrow$}
\DefMacro{qwen-medium_java_rparentheses-semicolon}{56.33}
\DefMacro{qwen-medium_java_rparentheses-semicolon_delta}{-1.03}
\DefMacro{qwen-medium_java_rparentheses-semicolon_trend}{$\downarrow$}
\DefMacro{qwen-medium_java_period-name}{56.72}
\DefMacro{qwen-medium_java_period-name_delta}{-0.64}
\DefMacro{qwen-medium_java_period-name_trend}{$\downarrow$}
\DefMacro{qwen-medium_java_lparentheses-rparentheses}{56.85}
\DefMacro{qwen-medium_java_lparentheses-rparentheses_delta}{-0.51}
\DefMacro{qwen-medium_java_lparentheses-rparentheses_trend}{$\downarrow$}
\DefMacro{qwen-medium_java_lparentheses-name}{57.36}
\DefMacro{qwen-medium_java_lparentheses-name_delta}{+0.00}
\DefMacro{qwen-medium_java_lparentheses-name_trend}{}
\DefMacro{qwen-medium_java_rparentheses-rparentheses}{57.36}
\DefMacro{qwen-medium_java_rparentheses-rparentheses_delta}{+0.00}
\DefMacro{qwen-medium_java_rparentheses-rparentheses_trend}{}
\DefMacro{qwen-medium_java_period-asterisk}{58.14}
\DefMacro{qwen-medium_java_period-asterisk_delta}{+0.78}
\DefMacro{qwen-medium_java_period-asterisk_trend}{$\uparrow$}
\DefMacro{qwen-medium_java_double-plus-rparentheses}{56.72}
\DefMacro{qwen-medium_java_double-plus-rparentheses_delta}{-0.64}
\DefMacro{qwen-medium_java_double-plus-rparentheses_trend}{$\downarrow$}
\DefMacro{qwen-medium_java_rparentheses-period}{56.72}
\DefMacro{qwen-medium_java_rparentheses-period_delta}{-0.64}
\DefMacro{qwen-medium_java_rparentheses-period_trend}{$\downarrow$}
\DefMacro{qwen-medium_java_op-semicolon}{55.81}
\DefMacro{qwen-medium_java_op-semicolon_delta}{-1.55}
\DefMacro{qwen-medium_java_op-semicolon_trend}{$\downarrow$}
\DefMacro{qwen-medium_java_op-lparentheses}{57.49}
\DefMacro{qwen-medium_java_op-lparentheses_delta}{+0.13}
\DefMacro{qwen-medium_java_op-lparentheses_trend}{$\uparrow$}
\DefMacro{qwen-medium_java_op-name}{55.43}
\DefMacro{qwen-medium_java_op-name_delta}{-1.93}
\DefMacro{qwen-medium_java_op-name_trend}{$\downarrow$}
\DefMacro{qwen-medium_java_op-all}{56.07}
\DefMacro{qwen-medium_java_op-all_delta}{-1.29}
\DefMacro{qwen-medium_java_op-all_trend}{$\downarrow$}

\DefMacro{dscoder-small_java_baseline}{38.50}
\DefMacro{dscoder-small_java_camel-case-snake-case}{37.98}
\DefMacro{dscoder-small_java_camel-case-snake-case_delta}{-0.52}
\DefMacro{dscoder-small_java_camel-case-snake-case_trend}{$\downarrow$}
\DefMacro{dscoder-small_java_camel-case-pascal-case}{39.02}
\DefMacro{dscoder-small_java_camel-case-pascal-case_delta}{+0.52}
\DefMacro{dscoder-small_java_camel-case-pascal-case_trend}{$\uparrow$}
\DefMacro{dscoder-small_java_camel-case-screaming-snake-case}{38.37}
\DefMacro{dscoder-small_java_camel-case-screaming-snake-case_delta}{-0.13}
\DefMacro{dscoder-small_java_camel-case-screaming-snake-case_trend}{$\downarrow$}
\DefMacro{dscoder-small_java_rparentheses-semicolon}{37.34}
\DefMacro{dscoder-small_java_rparentheses-semicolon_delta}{-1.16}
\DefMacro{dscoder-small_java_rparentheses-semicolon_trend}{$\downarrow$}
\DefMacro{dscoder-small_java_period-name}{37.34}
\DefMacro{dscoder-small_java_period-name_delta}{-1.16}
\DefMacro{dscoder-small_java_period-name_trend}{$\downarrow$}
\DefMacro{dscoder-small_java_lparentheses-rparentheses}{37.86}
\DefMacro{dscoder-small_java_lparentheses-rparentheses_delta}{-0.64}
\DefMacro{dscoder-small_java_lparentheses-rparentheses_trend}{$\downarrow$}
\DefMacro{dscoder-small_java_lparentheses-name}{36.43}
\DefMacro{dscoder-small_java_lparentheses-name_delta}{-2.07}
\DefMacro{dscoder-small_java_lparentheses-name_trend}{$\downarrow$}
\DefMacro{dscoder-small_java_rparentheses-rparentheses}{37.47}
\DefMacro{dscoder-small_java_rparentheses-rparentheses_delta}{-1.03}
\DefMacro{dscoder-small_java_rparentheses-rparentheses_trend}{$\downarrow$}
\DefMacro{dscoder-small_java_period-asterisk}{36.95}
\DefMacro{dscoder-small_java_period-asterisk_delta}{-1.55}
\DefMacro{dscoder-small_java_period-asterisk_trend}{$\downarrow$}
\DefMacro{dscoder-small_java_double-plus-rparentheses}{38.63}
\DefMacro{dscoder-small_java_double-plus-rparentheses_delta}{+0.13}
\DefMacro{dscoder-small_java_double-plus-rparentheses_trend}{$\uparrow$}
\DefMacro{dscoder-small_java_rparentheses-period}{37.34}
\DefMacro{dscoder-small_java_rparentheses-period_delta}{-1.16}
\DefMacro{dscoder-small_java_rparentheses-period_trend}{$\downarrow$}
\DefMacro{dscoder-small_java_op-semicolon}{38.63}
\DefMacro{dscoder-small_java_op-semicolon_delta}{+0.13}
\DefMacro{dscoder-small_java_op-semicolon_trend}{$\uparrow$}
\DefMacro{dscoder-small_java_op-lparentheses}{37.98}
\DefMacro{dscoder-small_java_op-lparentheses_delta}{-0.52}
\DefMacro{dscoder-small_java_op-lparentheses_trend}{$\downarrow$}
\DefMacro{dscoder-small_java_op-name}{35.40}
\DefMacro{dscoder-small_java_op-name_delta}{-3.10}
\DefMacro{dscoder-small_java_op-name_trend}{$\downarrow$}
\DefMacro{dscoder-small_java_op-all}{33.98}
\DefMacro{dscoder-small_java_op-all_delta}{-4.52}
\DefMacro{dscoder-small_java_op-all_trend}{$\downarrow$}

\DefMacro{dscoder-large_java_baseline}{57.36}
\DefMacro{dscoder-large_java_camel-case-snake-case}{57.11}
\DefMacro{dscoder-large_java_camel-case-snake-case_delta}{-0.25}
\DefMacro{dscoder-large_java_camel-case-snake-case_trend}{$\downarrow$}
\DefMacro{dscoder-large_java_camel-case-pascal-case}{57.36}
\DefMacro{dscoder-large_java_camel-case-pascal-case_delta}{+0.00}
\DefMacro{dscoder-large_java_camel-case-pascal-case_trend}{}
\DefMacro{dscoder-large_java_camel-case-screaming-snake-case}{56.46}
\DefMacro{dscoder-large_java_camel-case-screaming-snake-case_delta}{-0.90}
\DefMacro{dscoder-large_java_camel-case-screaming-snake-case_trend}{$\downarrow$}
\DefMacro{dscoder-large_java_rparentheses-semicolon}{57.11}
\DefMacro{dscoder-large_java_rparentheses-semicolon_delta}{-0.25}
\DefMacro{dscoder-large_java_rparentheses-semicolon_trend}{$\downarrow$}
\DefMacro{dscoder-large_java_period-name}{59.43}
\DefMacro{dscoder-large_java_period-name_delta}{+2.07}
\DefMacro{dscoder-large_java_period-name_trend}{$\uparrow$}
\DefMacro{dscoder-large_java_lparentheses-rparentheses}{57.62}
\DefMacro{dscoder-large_java_lparentheses-rparentheses_delta}{+0.26}
\DefMacro{dscoder-large_java_lparentheses-rparentheses_trend}{$\uparrow$}
\DefMacro{dscoder-large_java_lparentheses-name}{58.66}
\DefMacro{dscoder-large_java_lparentheses-name_delta}{+1.30}
\DefMacro{dscoder-large_java_lparentheses-name_trend}{$\uparrow$}
\DefMacro{dscoder-large_java_rparentheses-rparentheses}{56.98}
\DefMacro{dscoder-large_java_rparentheses-rparentheses_delta}{-0.38}
\DefMacro{dscoder-large_java_rparentheses-rparentheses_trend}{$\downarrow$}
\DefMacro{dscoder-large_java_period-asterisk}{57.75}
\DefMacro{dscoder-large_java_period-asterisk_delta}{+0.39}
\DefMacro{dscoder-large_java_period-asterisk_trend}{$\uparrow$}
\DefMacro{dscoder-large_java_double-plus-rparentheses}{58.27}
\DefMacro{dscoder-large_java_double-plus-rparentheses_delta}{+0.91}
\DefMacro{dscoder-large_java_double-plus-rparentheses_trend}{$\uparrow$}
\DefMacro{dscoder-large_java_rparentheses-period}{57.88}
\DefMacro{dscoder-large_java_rparentheses-period_delta}{+0.52}
\DefMacro{dscoder-large_java_rparentheses-period_trend}{$\uparrow$}
\DefMacro{dscoder-large_java_op-semicolon}{58.53}
\DefMacro{dscoder-large_java_op-semicolon_delta}{+1.17}
\DefMacro{dscoder-large_java_op-semicolon_trend}{$\uparrow$}
\DefMacro{dscoder-large_java_op-lparentheses}{57.49}
\DefMacro{dscoder-large_java_op-lparentheses_delta}{+0.13}
\DefMacro{dscoder-large_java_op-lparentheses_trend}{$\uparrow$}
\DefMacro{dscoder-large_java_op-name}{58.91}
\DefMacro{dscoder-large_java_op-name_delta}{+1.55}
\DefMacro{dscoder-large_java_op-name_trend}{$\uparrow$}
\DefMacro{dscoder-large_java_op-all}{56.33}
\DefMacro{dscoder-large_java_op-all_delta}{-1.03}
\DefMacro{dscoder-large_java_op-all_trend}{$\downarrow$}

\DefMacro{dscoder-medium_java_baseline}{58.01}
\DefMacro{dscoder-medium_java_camel-case-snake-case}{57.36}
\DefMacro{dscoder-medium_java_camel-case-snake-case_delta}{-0.65}
\DefMacro{dscoder-medium_java_camel-case-snake-case_trend}{$\downarrow$}
\DefMacro{dscoder-medium_java_camel-case-pascal-case}{58.14}
\DefMacro{dscoder-medium_java_camel-case-pascal-case_delta}{+0.13}
\DefMacro{dscoder-medium_java_camel-case-pascal-case_trend}{$\uparrow$}
\DefMacro{dscoder-medium_java_camel-case-screaming-snake-case}{56.33}
\DefMacro{dscoder-medium_java_camel-case-screaming-snake-case_delta}{-1.68}
\DefMacro{dscoder-medium_java_camel-case-screaming-snake-case_trend}{$\downarrow$}
\DefMacro{dscoder-medium_java_rparentheses-semicolon}{57.11}
\DefMacro{dscoder-medium_java_rparentheses-semicolon_delta}{-0.90}
\DefMacro{dscoder-medium_java_rparentheses-semicolon_trend}{$\downarrow$}
\DefMacro{dscoder-medium_java_period-name}{55.43}
\DefMacro{dscoder-medium_java_period-name_delta}{-2.58}
\DefMacro{dscoder-medium_java_period-name_trend}{$\downarrow$}
\DefMacro{dscoder-medium_java_lparentheses-rparentheses}{57.11}
\DefMacro{dscoder-medium_java_lparentheses-rparentheses_delta}{-0.90}
\DefMacro{dscoder-medium_java_lparentheses-rparentheses_trend}{$\downarrow$}
\DefMacro{dscoder-medium_java_lparentheses-name}{57.49}
\DefMacro{dscoder-medium_java_lparentheses-name_delta}{-0.52}
\DefMacro{dscoder-medium_java_lparentheses-name_trend}{$\downarrow$}
\DefMacro{dscoder-medium_java_rparentheses-rparentheses}{58.27}
\DefMacro{dscoder-medium_java_rparentheses-rparentheses_delta}{+0.26}
\DefMacro{dscoder-medium_java_rparentheses-rparentheses_trend}{$\uparrow$}
\DefMacro{dscoder-medium_java_period-asterisk}{56.59}
\DefMacro{dscoder-medium_java_period-asterisk_delta}{-1.42}
\DefMacro{dscoder-medium_java_period-asterisk_trend}{$\downarrow$}
\DefMacro{dscoder-medium_java_double-plus-rparentheses}{57.49}
\DefMacro{dscoder-medium_java_double-plus-rparentheses_delta}{-0.52}
\DefMacro{dscoder-medium_java_double-plus-rparentheses_trend}{$\downarrow$}
\DefMacro{dscoder-medium_java_rparentheses-period}{58.66}
\DefMacro{dscoder-medium_java_rparentheses-period_delta}{+0.65}
\DefMacro{dscoder-medium_java_rparentheses-period_trend}{$\uparrow$}
\DefMacro{dscoder-medium_java_op-semicolon}{55.94}
\DefMacro{dscoder-medium_java_op-semicolon_delta}{-2.07}
\DefMacro{dscoder-medium_java_op-semicolon_trend}{$\downarrow$}
\DefMacro{dscoder-medium_java_op-lparentheses}{58.53}
\DefMacro{dscoder-medium_java_op-lparentheses_delta}{+0.52}
\DefMacro{dscoder-medium_java_op-lparentheses_trend}{$\uparrow$}
\DefMacro{dscoder-medium_java_op-name}{55.04}
\DefMacro{dscoder-medium_java_op-name_delta}{-2.97}
\DefMacro{dscoder-medium_java_op-name_trend}{$\downarrow$}
\DefMacro{dscoder-medium_java_op-all}{53.10}
\DefMacro{dscoder-medium_java_op-all_delta}{-4.91}
\DefMacro{dscoder-medium_java_op-all_trend}{$\downarrow$}

\DefMacro{average_java_baseline}{49.74}
\DefMacro{average_java_camel-case-snake-case}{49.93}
\DefMacro{average_java_camel-case-snake-case_delta}{+0.19}
\DefMacro{average_java_camel-case-snake-case_trend}{$\uparrow$}
\DefMacro{average_java_camel-case-pascal-case}{50.06}
\DefMacro{average_java_camel-case-pascal-case_delta}{+0.32}
\DefMacro{average_java_camel-case-pascal-case_trend}{$\uparrow$}
\DefMacro{average_java_camel-case-screaming-snake-case}{49.64}
\DefMacro{average_java_camel-case-screaming-snake-case_delta}{-0.10}
\DefMacro{average_java_camel-case-screaming-snake-case_trend}{$\downarrow$}
\DefMacro{average_java_rparentheses-semicolon}{49.67}
\DefMacro{average_java_rparentheses-semicolon_delta}{-0.07}
\DefMacro{average_java_rparentheses-semicolon_trend}{$\downarrow$}
\DefMacro{average_java_period-name}{48.59}
\DefMacro{average_java_period-name_delta}{-1.15}
\DefMacro{average_java_period-name_trend}{$\downarrow$}
\DefMacro{average_java_lparentheses-rparentheses}{48.65}
\DefMacro{average_java_lparentheses-rparentheses_delta}{-1.09}
\DefMacro{average_java_lparentheses-rparentheses_trend}{$\downarrow$}
\DefMacro{average_java_lparentheses-name}{49.38}
\DefMacro{average_java_lparentheses-name_delta}{-0.36}
\DefMacro{average_java_lparentheses-name_trend}{$\downarrow$}
\DefMacro{average_java_rparentheses-rparentheses}{49.44}
\DefMacro{average_java_rparentheses-rparentheses_delta}{-0.30}
\DefMacro{average_java_rparentheses-rparentheses_trend}{$\downarrow$}
\DefMacro{average_java_period-asterisk}{49.37}
\DefMacro{average_java_period-asterisk_delta}{-0.37}
\DefMacro{average_java_period-asterisk_trend}{$\downarrow$}
\DefMacro{average_java_double-plus-rparentheses}{49.90}
\DefMacro{average_java_double-plus-rparentheses_delta}{+0.16}
\DefMacro{average_java_double-plus-rparentheses_trend}{$\uparrow$}
\DefMacro{average_java_rparentheses-period}{49.58}
\DefMacro{average_java_rparentheses-period_delta}{-0.16}
\DefMacro{average_java_rparentheses-period_trend}{$\downarrow$}
\DefMacro{average_java_op-semicolon}{48.92}
\DefMacro{average_java_op-semicolon_delta}{-0.82}
\DefMacro{average_java_op-semicolon_trend}{$\downarrow$}
\DefMacro{average_java_op-lparentheses}{49.68}
\DefMacro{average_java_op-lparentheses_delta}{-0.06}
\DefMacro{average_java_op-lparentheses_trend}{$\downarrow$}
\DefMacro{average_java_op-name}{48.06}
\DefMacro{average_java_op-name_delta}{-1.68}
\DefMacro{average_java_op-name_trend}{$\downarrow$}
\DefMacro{average_java_op-all}{46.84}
\DefMacro{average_java_op-all_delta}{-2.90}
\DefMacro{average_java_op-all_trend}{$\downarrow$}

\DefMacro{llama-medium_python_baseline}{49.87}
\DefMacro{llama-medium_python_snake-case-camel-case}{51.04}
\DefMacro{llama-medium_python_snake-case-camel-case_delta}{+1.17}
\DefMacro{llama-medium_python_snake-case-camel-case_trend}{$\uparrow$}
\DefMacro{llama-medium_python_snake-case-pascal-case}{50.91}
\DefMacro{llama-medium_python_snake-case-pascal-case_delta}{+1.04}
\DefMacro{llama-medium_python_snake-case-pascal-case_trend}{$\uparrow$}
\DefMacro{llama-medium_python_snake-case-screaming-snake-case}{50.65}
\DefMacro{llama-medium_python_snake-case-screaming-snake-case_delta}{+0.78}
\DefMacro{llama-medium_python_snake-case-screaming-snake-case_trend}{$\uparrow$}
\DefMacro{llama-medium_python_lparentheses-name}{49.87}
\DefMacro{llama-medium_python_lparentheses-name_delta}{+0.00}
\DefMacro{llama-medium_python_lparentheses-name_trend}{}
\DefMacro{llama-medium_python_lparentheses-rparentheses}{49.22}
\DefMacro{llama-medium_python_lparentheses-rparentheses_delta}{-0.65}
\DefMacro{llama-medium_python_lparentheses-rparentheses_trend}{$\downarrow$}
\DefMacro{llama-medium_python_rparentheses-colon}{50.65}
\DefMacro{llama-medium_python_rparentheses-colon_delta}{+0.78}
\DefMacro{llama-medium_python_rparentheses-colon_trend}{$\uparrow$}
\DefMacro{llama-medium_python_rparentheses-rparentheses}{50.13}
\DefMacro{llama-medium_python_rparentheses-rparentheses_delta}{+0.26}
\DefMacro{llama-medium_python_rparentheses-rparentheses_trend}{$\uparrow$}
\DefMacro{llama-medium_python_lsquarebracket-name}{49.35}
\DefMacro{llama-medium_python_lsquarebracket-name_delta}{-0.52}
\DefMacro{llama-medium_python_lsquarebracket-name_trend}{$\downarrow$}
\DefMacro{llama-medium_python_period-name}{50.26}
\DefMacro{llama-medium_python_period-name_delta}{+0.39}
\DefMacro{llama-medium_python_period-name_trend}{$\uparrow$}
\DefMacro{llama-medium_python_rsquarebracket-rparentheses}{50.65}
\DefMacro{llama-medium_python_rsquarebracket-rparentheses_delta}{+0.78}
\DefMacro{llama-medium_python_rsquarebracket-rparentheses_trend}{$\uparrow$}
\DefMacro{llama-medium_python_op-lsquarebracket}{50.65}
\DefMacro{llama-medium_python_op-lsquarebracket_delta}{+0.78}
\DefMacro{llama-medium_python_op-lsquarebracket_trend}{$\uparrow$}
\DefMacro{llama-medium_python_op-rsquarebracket}{50.78}
\DefMacro{llama-medium_python_op-rsquarebracket_delta}{+0.91}
\DefMacro{llama-medium_python_op-rsquarebracket_trend}{$\uparrow$}
\DefMacro{llama-medium_python_op-dash}{50.39}
\DefMacro{llama-medium_python_op-dash_delta}{+0.52}
\DefMacro{llama-medium_python_op-dash_trend}{$\uparrow$}
\DefMacro{llama-medium_python_op-name}{50.39}
\DefMacro{llama-medium_python_op-name_delta}{+0.52}
\DefMacro{llama-medium_python_op-name_trend}{$\uparrow$}
\DefMacro{llama-medium_python_op-all}{49.87}
\DefMacro{llama-medium_python_op-all_delta}{+0.00}
\DefMacro{llama-medium_python_op-all_trend}{}

\DefMacro{llama-small_python_baseline}{39.12}
\DefMacro{llama-small_python_snake-case-camel-case}{40.03}
\DefMacro{llama-small_python_snake-case-camel-case_delta}{+0.91}
\DefMacro{llama-small_python_snake-case-camel-case_trend}{$\uparrow$}
\DefMacro{llama-small_python_snake-case-pascal-case}{37.56}
\DefMacro{llama-small_python_snake-case-pascal-case_delta}{-1.56}
\DefMacro{llama-small_python_snake-case-pascal-case_trend}{$\downarrow$}
\DefMacro{llama-small_python_snake-case-screaming-snake-case}{38.08}
\DefMacro{llama-small_python_snake-case-screaming-snake-case_delta}{-1.04}
\DefMacro{llama-small_python_snake-case-screaming-snake-case_trend}{$\downarrow$}
\DefMacro{llama-small_python_lparentheses-name}{40.16}
\DefMacro{llama-small_python_lparentheses-name_delta}{+1.04}
\DefMacro{llama-small_python_lparentheses-name_trend}{$\uparrow$}
\DefMacro{llama-small_python_lparentheses-rparentheses}{38.73}
\DefMacro{llama-small_python_lparentheses-rparentheses_delta}{-0.39}
\DefMacro{llama-small_python_lparentheses-rparentheses_trend}{$\downarrow$}
\DefMacro{llama-small_python_rparentheses-colon}{38.47}
\DefMacro{llama-small_python_rparentheses-colon_delta}{-0.65}
\DefMacro{llama-small_python_rparentheses-colon_trend}{$\downarrow$}
\DefMacro{llama-small_python_rparentheses-rparentheses}{37.95}
\DefMacro{llama-small_python_rparentheses-rparentheses_delta}{-1.17}
\DefMacro{llama-small_python_rparentheses-rparentheses_trend}{$\downarrow$}
\DefMacro{llama-small_python_lsquarebracket-name}{40.03}
\DefMacro{llama-small_python_lsquarebracket-name_delta}{+0.91}
\DefMacro{llama-small_python_lsquarebracket-name_trend}{$\uparrow$}
\DefMacro{llama-small_python_period-name}{39.12}
\DefMacro{llama-small_python_period-name_delta}{+0.00}
\DefMacro{llama-small_python_period-name_trend}{}
\DefMacro{llama-small_python_rsquarebracket-rparentheses}{39.77}
\DefMacro{llama-small_python_rsquarebracket-rparentheses_delta}{+0.65}
\DefMacro{llama-small_python_rsquarebracket-rparentheses_trend}{$\uparrow$}
\DefMacro{llama-small_python_op-lsquarebracket}{39.64}
\DefMacro{llama-small_python_op-lsquarebracket_delta}{+0.52}
\DefMacro{llama-small_python_op-lsquarebracket_trend}{$\uparrow$}
\DefMacro{llama-small_python_op-rsquarebracket}{38.60}
\DefMacro{llama-small_python_op-rsquarebracket_delta}{-0.52}
\DefMacro{llama-small_python_op-rsquarebracket_trend}{$\downarrow$}
\DefMacro{llama-small_python_op-dash}{39.38}
\DefMacro{llama-small_python_op-dash_delta}{+0.26}
\DefMacro{llama-small_python_op-dash_trend}{$\uparrow$}
\DefMacro{llama-small_python_op-name}{40.41}
\DefMacro{llama-small_python_op-name_delta}{+1.29}
\DefMacro{llama-small_python_op-name_trend}{$\uparrow$}
\DefMacro{llama-small_python_op-all}{37.44}
\DefMacro{llama-small_python_op-all_delta}{-1.68}
\DefMacro{llama-small_python_op-all_trend}{$\downarrow$}

\DefMacro{llama-large_python_baseline}{69.04}
\DefMacro{llama-large_python_snake-case-camel-case}{68.91}
\DefMacro{llama-large_python_snake-case-camel-case_delta}{-0.13}
\DefMacro{llama-large_python_snake-case-camel-case_trend}{$\downarrow$}
\DefMacro{llama-large_python_snake-case-pascal-case}{68.65}
\DefMacro{llama-large_python_snake-case-pascal-case_delta}{-0.39}
\DefMacro{llama-large_python_snake-case-pascal-case_trend}{$\downarrow$}
\DefMacro{llama-large_python_snake-case-screaming-snake-case}{66.19}
\DefMacro{llama-large_python_snake-case-screaming-snake-case_delta}{-2.85}
\DefMacro{llama-large_python_snake-case-screaming-snake-case_trend}{$\downarrow$}
\DefMacro{llama-large_python_lparentheses-name}{69.04}
\DefMacro{llama-large_python_lparentheses-name_delta}{+0.00}
\DefMacro{llama-large_python_lparentheses-name_trend}{}
\DefMacro{llama-large_python_lparentheses-rparentheses}{68.39}
\DefMacro{llama-large_python_lparentheses-rparentheses_delta}{-0.65}
\DefMacro{llama-large_python_lparentheses-rparentheses_trend}{$\downarrow$}
\DefMacro{llama-large_python_rparentheses-colon}{69.17}
\DefMacro{llama-large_python_rparentheses-colon_delta}{+0.13}
\DefMacro{llama-large_python_rparentheses-colon_trend}{$\uparrow$}
\DefMacro{llama-large_python_rparentheses-rparentheses}{69.30}
\DefMacro{llama-large_python_rparentheses-rparentheses_delta}{+0.26}
\DefMacro{llama-large_python_rparentheses-rparentheses_trend}{$\uparrow$}
\DefMacro{llama-large_python_lsquarebracket-name}{68.26}
\DefMacro{llama-large_python_lsquarebracket-name_delta}{-0.78}
\DefMacro{llama-large_python_lsquarebracket-name_trend}{$\downarrow$}
\DefMacro{llama-large_python_period-name}{67.49}
\DefMacro{llama-large_python_period-name_delta}{-1.55}
\DefMacro{llama-large_python_period-name_trend}{$\downarrow$}
\DefMacro{llama-large_python_rsquarebracket-rparentheses}{69.30}
\DefMacro{llama-large_python_rsquarebracket-rparentheses_delta}{+0.26}
\DefMacro{llama-large_python_rsquarebracket-rparentheses_trend}{$\uparrow$}
\DefMacro{llama-large_python_op-lsquarebracket}{68.78}
\DefMacro{llama-large_python_op-lsquarebracket_delta}{-0.26}
\DefMacro{llama-large_python_op-lsquarebracket_trend}{$\downarrow$}
\DefMacro{llama-large_python_op-rsquarebracket}{68.91}
\DefMacro{llama-large_python_op-rsquarebracket_delta}{-0.13}
\DefMacro{llama-large_python_op-rsquarebracket_trend}{$\downarrow$}
\DefMacro{llama-large_python_op-dash}{68.65}
\DefMacro{llama-large_python_op-dash_delta}{-0.39}
\DefMacro{llama-large_python_op-dash_trend}{$\downarrow$}
\DefMacro{llama-large_python_op-name}{67.62}
\DefMacro{llama-large_python_op-name_delta}{-1.42}
\DefMacro{llama-large_python_op-name_trend}{$\downarrow$}
\DefMacro{llama-large_python_op-all}{67.62}
\DefMacro{llama-large_python_op-all_delta}{-1.42}
\DefMacro{llama-large_python_op-all_trend}{$\downarrow$}

\DefMacro{qwen-small_python_baseline}{40.67}
\DefMacro{qwen-small_python_snake-case-camel-case}{39.77}
\DefMacro{qwen-small_python_snake-case-camel-case_delta}{-0.90}
\DefMacro{qwen-small_python_snake-case-camel-case_trend}{$\downarrow$}
\DefMacro{qwen-small_python_snake-case-pascal-case}{39.25}
\DefMacro{qwen-small_python_snake-case-pascal-case_delta}{-1.42}
\DefMacro{qwen-small_python_snake-case-pascal-case_trend}{$\downarrow$}
\DefMacro{qwen-small_python_snake-case-screaming-snake-case}{39.38}
\DefMacro{qwen-small_python_snake-case-screaming-snake-case_delta}{-1.29}
\DefMacro{qwen-small_python_snake-case-screaming-snake-case_trend}{$\downarrow$}
\DefMacro{qwen-small_python_lparentheses-name}{39.64}
\DefMacro{qwen-small_python_lparentheses-name_delta}{-1.03}
\DefMacro{qwen-small_python_lparentheses-name_trend}{$\downarrow$}
\DefMacro{qwen-small_python_lparentheses-rparentheses}{39.38}
\DefMacro{qwen-small_python_lparentheses-rparentheses_delta}{-1.29}
\DefMacro{qwen-small_python_lparentheses-rparentheses_trend}{$\downarrow$}
\DefMacro{qwen-small_python_rparentheses-colon}{40.67}
\DefMacro{qwen-small_python_rparentheses-colon_delta}{+0.00}
\DefMacro{qwen-small_python_rparentheses-colon_trend}{}
\DefMacro{qwen-small_python_rparentheses-rparentheses}{40.54}
\DefMacro{qwen-small_python_rparentheses-rparentheses_delta}{-0.13}
\DefMacro{qwen-small_python_rparentheses-rparentheses_trend}{$\downarrow$}
\DefMacro{qwen-small_python_lsquarebracket-name}{40.67}
\DefMacro{qwen-small_python_lsquarebracket-name_delta}{+0.00}
\DefMacro{qwen-small_python_lsquarebracket-name_trend}{}
\DefMacro{qwen-small_python_period-name}{39.77}
\DefMacro{qwen-small_python_period-name_delta}{-0.90}
\DefMacro{qwen-small_python_period-name_trend}{$\downarrow$}
\DefMacro{qwen-small_python_rsquarebracket-rparentheses}{40.54}
\DefMacro{qwen-small_python_rsquarebracket-rparentheses_delta}{-0.13}
\DefMacro{qwen-small_python_rsquarebracket-rparentheses_trend}{$\downarrow$}
\DefMacro{qwen-small_python_op-lsquarebracket}{40.41}
\DefMacro{qwen-small_python_op-lsquarebracket_delta}{-0.26}
\DefMacro{qwen-small_python_op-lsquarebracket_trend}{$\downarrow$}
\DefMacro{qwen-small_python_op-rsquarebracket}{40.80}
\DefMacro{qwen-small_python_op-rsquarebracket_delta}{+0.13}
\DefMacro{qwen-small_python_op-rsquarebracket_trend}{$\uparrow$}
\DefMacro{qwen-small_python_op-dash}{40.54}
\DefMacro{qwen-small_python_op-dash_delta}{-0.13}
\DefMacro{qwen-small_python_op-dash_trend}{$\downarrow$}
\DefMacro{qwen-small_python_op-name}{39.38}
\DefMacro{qwen-small_python_op-name_delta}{-1.29}
\DefMacro{qwen-small_python_op-name_trend}{$\downarrow$}
\DefMacro{qwen-small_python_op-all}{38.34}
\DefMacro{qwen-small_python_op-all_delta}{-2.33}
\DefMacro{qwen-small_python_op-all_trend}{$\downarrow$}

\DefMacro{qwen-large_python_baseline}{76.17}
\DefMacro{qwen-large_python_snake-case-camel-case}{77.85}
\DefMacro{qwen-large_python_snake-case-camel-case_delta}{+1.68}
\DefMacro{qwen-large_python_snake-case-camel-case_trend}{$\uparrow$}
\DefMacro{qwen-large_python_snake-case-pascal-case}{77.72}
\DefMacro{qwen-large_python_snake-case-pascal-case_delta}{+1.55}
\DefMacro{qwen-large_python_snake-case-pascal-case_trend}{$\uparrow$}
\DefMacro{qwen-large_python_snake-case-screaming-snake-case}{76.81}
\DefMacro{qwen-large_python_snake-case-screaming-snake-case_delta}{+0.64}
\DefMacro{qwen-large_python_snake-case-screaming-snake-case_trend}{$\uparrow$}
\DefMacro{qwen-large_python_lparentheses-name}{76.68}
\DefMacro{qwen-large_python_lparentheses-name_delta}{+0.51}
\DefMacro{qwen-large_python_lparentheses-name_trend}{$\uparrow$}
\DefMacro{qwen-large_python_lparentheses-rparentheses}{74.09}
\DefMacro{qwen-large_python_lparentheses-rparentheses_delta}{-2.08}
\DefMacro{qwen-large_python_lparentheses-rparentheses_trend}{$\downarrow$}
\DefMacro{qwen-large_python_rparentheses-colon}{77.46}
\DefMacro{qwen-large_python_rparentheses-colon_delta}{+1.29}
\DefMacro{qwen-large_python_rparentheses-colon_trend}{$\uparrow$}
\DefMacro{qwen-large_python_rparentheses-rparentheses}{76.55}
\DefMacro{qwen-large_python_rparentheses-rparentheses_delta}{+0.38}
\DefMacro{qwen-large_python_rparentheses-rparentheses_trend}{$\uparrow$}
\DefMacro{qwen-large_python_lsquarebracket-name}{76.42}
\DefMacro{qwen-large_python_lsquarebracket-name_delta}{+0.25}
\DefMacro{qwen-large_python_lsquarebracket-name_trend}{$\uparrow$}
\DefMacro{qwen-large_python_period-name}{76.30}
\DefMacro{qwen-large_python_period-name_delta}{+0.13}
\DefMacro{qwen-large_python_period-name_trend}{$\uparrow$}
\DefMacro{qwen-large_python_rsquarebracket-rparentheses}{73.19}
\DefMacro{qwen-large_python_rsquarebracket-rparentheses_delta}{-2.98}
\DefMacro{qwen-large_python_rsquarebracket-rparentheses_trend}{$\downarrow$}
\DefMacro{qwen-large_python_op-lsquarebracket}{75.91}
\DefMacro{qwen-large_python_op-lsquarebracket_delta}{-0.26}
\DefMacro{qwen-large_python_op-lsquarebracket_trend}{$\downarrow$}
\DefMacro{qwen-large_python_op-rsquarebracket}{76.94}
\DefMacro{qwen-large_python_op-rsquarebracket_delta}{+0.77}
\DefMacro{qwen-large_python_op-rsquarebracket_trend}{$\uparrow$}
\DefMacro{qwen-large_python_op-dash}{76.68}
\DefMacro{qwen-large_python_op-dash_delta}{+0.51}
\DefMacro{qwen-large_python_op-dash_trend}{$\uparrow$}
\DefMacro{qwen-large_python_op-name}{76.55}
\DefMacro{qwen-large_python_op-name_delta}{+0.38}
\DefMacro{qwen-large_python_op-name_trend}{$\uparrow$}
\DefMacro{qwen-large_python_op-all}{75.13}
\DefMacro{qwen-large_python_op-all_delta}{-1.04}
\DefMacro{qwen-large_python_op-all_trend}{$\downarrow$}

\DefMacro{qwen-medium_python_baseline}{64.51}
\DefMacro{qwen-medium_python_snake-case-camel-case}{65.03}
\DefMacro{qwen-medium_python_snake-case-camel-case_delta}{+0.52}
\DefMacro{qwen-medium_python_snake-case-camel-case_trend}{$\uparrow$}
\DefMacro{qwen-medium_python_snake-case-pascal-case}{64.77}
\DefMacro{qwen-medium_python_snake-case-pascal-case_delta}{+0.26}
\DefMacro{qwen-medium_python_snake-case-pascal-case_trend}{$\uparrow$}
\DefMacro{qwen-medium_python_snake-case-screaming-snake-case}{64.51}
\DefMacro{qwen-medium_python_snake-case-screaming-snake-case_delta}{+0.00}
\DefMacro{qwen-medium_python_snake-case-screaming-snake-case_trend}{}
\DefMacro{qwen-medium_python_lparentheses-name}{63.08}
\DefMacro{qwen-medium_python_lparentheses-name_delta}{-1.43}
\DefMacro{qwen-medium_python_lparentheses-name_trend}{$\downarrow$}
\DefMacro{qwen-medium_python_lparentheses-rparentheses}{63.73}
\DefMacro{qwen-medium_python_lparentheses-rparentheses_delta}{-0.78}
\DefMacro{qwen-medium_python_lparentheses-rparentheses_trend}{$\downarrow$}
\DefMacro{qwen-medium_python_rparentheses-colon}{63.99}
\DefMacro{qwen-medium_python_rparentheses-colon_delta}{-0.52}
\DefMacro{qwen-medium_python_rparentheses-colon_trend}{$\downarrow$}
\DefMacro{qwen-medium_python_rparentheses-rparentheses}{64.90}
\DefMacro{qwen-medium_python_rparentheses-rparentheses_delta}{+0.39}
\DefMacro{qwen-medium_python_rparentheses-rparentheses_trend}{$\uparrow$}
\DefMacro{qwen-medium_python_lsquarebracket-name}{63.34}
\DefMacro{qwen-medium_python_lsquarebracket-name_delta}{-1.17}
\DefMacro{qwen-medium_python_lsquarebracket-name_trend}{$\downarrow$}
\DefMacro{qwen-medium_python_period-name}{62.69}
\DefMacro{qwen-medium_python_period-name_delta}{-1.82}
\DefMacro{qwen-medium_python_period-name_trend}{$\downarrow$}
\DefMacro{qwen-medium_python_rsquarebracket-rparentheses}{64.51}
\DefMacro{qwen-medium_python_rsquarebracket-rparentheses_delta}{+0.00}
\DefMacro{qwen-medium_python_rsquarebracket-rparentheses_trend}{}
\DefMacro{qwen-medium_python_op-lsquarebracket}{64.77}
\DefMacro{qwen-medium_python_op-lsquarebracket_delta}{+0.26}
\DefMacro{qwen-medium_python_op-lsquarebracket_trend}{$\uparrow$}
\DefMacro{qwen-medium_python_op-rsquarebracket}{64.12}
\DefMacro{qwen-medium_python_op-rsquarebracket_delta}{-0.39}
\DefMacro{qwen-medium_python_op-rsquarebracket_trend}{$\downarrow$}
\DefMacro{qwen-medium_python_op-dash}{64.51}
\DefMacro{qwen-medium_python_op-dash_delta}{+0.00}
\DefMacro{qwen-medium_python_op-dash_trend}{}
\DefMacro{qwen-medium_python_op-name}{61.92}
\DefMacro{qwen-medium_python_op-name_delta}{-2.59}
\DefMacro{qwen-medium_python_op-name_trend}{$\downarrow$}
\DefMacro{qwen-medium_python_op-all}{63.08}
\DefMacro{qwen-medium_python_op-all_delta}{-1.43}
\DefMacro{qwen-medium_python_op-all_trend}{$\downarrow$}

\DefMacro{dscoder-small_python_baseline}{44.82}
\DefMacro{dscoder-small_python_snake-case-camel-case}{44.30}
\DefMacro{dscoder-small_python_snake-case-camel-case_delta}{-0.52}
\DefMacro{dscoder-small_python_snake-case-camel-case_trend}{$\downarrow$}
\DefMacro{dscoder-small_python_snake-case-pascal-case}{42.88}
\DefMacro{dscoder-small_python_snake-case-pascal-case_delta}{-1.94}
\DefMacro{dscoder-small_python_snake-case-pascal-case_trend}{$\downarrow$}
\DefMacro{dscoder-small_python_snake-case-screaming-snake-case}{42.23}
\DefMacro{dscoder-small_python_snake-case-screaming-snake-case_delta}{-2.59}
\DefMacro{dscoder-small_python_snake-case-screaming-snake-case_trend}{$\downarrow$}
\DefMacro{dscoder-small_python_lparentheses-name}{43.65}
\DefMacro{dscoder-small_python_lparentheses-name_delta}{-1.17}
\DefMacro{dscoder-small_python_lparentheses-name_trend}{$\downarrow$}
\DefMacro{dscoder-small_python_lparentheses-rparentheses}{45.08}
\DefMacro{dscoder-small_python_lparentheses-rparentheses_delta}{+0.26}
\DefMacro{dscoder-small_python_lparentheses-rparentheses_trend}{$\uparrow$}
\DefMacro{dscoder-small_python_rparentheses-colon}{44.56}
\DefMacro{dscoder-small_python_rparentheses-colon_delta}{-0.26}
\DefMacro{dscoder-small_python_rparentheses-colon_trend}{$\downarrow$}
\DefMacro{dscoder-small_python_rparentheses-rparentheses}{44.30}
\DefMacro{dscoder-small_python_rparentheses-rparentheses_delta}{-0.52}
\DefMacro{dscoder-small_python_rparentheses-rparentheses_trend}{$\downarrow$}
\DefMacro{dscoder-small_python_lsquarebracket-name}{44.30}
\DefMacro{dscoder-small_python_lsquarebracket-name_delta}{-0.52}
\DefMacro{dscoder-small_python_lsquarebracket-name_trend}{$\downarrow$}
\DefMacro{dscoder-small_python_period-name}{44.17}
\DefMacro{dscoder-small_python_period-name_delta}{-0.65}
\DefMacro{dscoder-small_python_period-name_trend}{$\downarrow$}
\DefMacro{dscoder-small_python_rsquarebracket-rparentheses}{44.82}
\DefMacro{dscoder-small_python_rsquarebracket-rparentheses_delta}{+0.00}
\DefMacro{dscoder-small_python_rsquarebracket-rparentheses_trend}{}
\DefMacro{dscoder-small_python_op-lsquarebracket}{43.65}
\DefMacro{dscoder-small_python_op-lsquarebracket_delta}{-1.17}
\DefMacro{dscoder-small_python_op-lsquarebracket_trend}{$\downarrow$}
\DefMacro{dscoder-small_python_op-rsquarebracket}{44.43}
\DefMacro{dscoder-small_python_op-rsquarebracket_delta}{-0.39}
\DefMacro{dscoder-small_python_op-rsquarebracket_trend}{$\downarrow$}
\DefMacro{dscoder-small_python_op-dash}{44.69}
\DefMacro{dscoder-small_python_op-dash_delta}{-0.13}
\DefMacro{dscoder-small_python_op-dash_trend}{$\downarrow$}
\DefMacro{dscoder-small_python_op-name}{42.62}
\DefMacro{dscoder-small_python_op-name_delta}{-2.20}
\DefMacro{dscoder-small_python_op-name_trend}{$\downarrow$}
\DefMacro{dscoder-small_python_op-all}{42.49}
\DefMacro{dscoder-small_python_op-all_delta}{-2.33}
\DefMacro{dscoder-small_python_op-all_trend}{$\downarrow$}

\DefMacro{dscoder-large_python_baseline}{68.13}
\DefMacro{dscoder-large_python_snake-case-camel-case}{68.39}
\DefMacro{dscoder-large_python_snake-case-camel-case_delta}{+0.26}
\DefMacro{dscoder-large_python_snake-case-camel-case_trend}{$\uparrow$}
\DefMacro{dscoder-large_python_snake-case-pascal-case}{68.39}
\DefMacro{dscoder-large_python_snake-case-pascal-case_delta}{+0.26}
\DefMacro{dscoder-large_python_snake-case-pascal-case_trend}{$\uparrow$}
\DefMacro{dscoder-large_python_snake-case-screaming-snake-case}{67.62}
\DefMacro{dscoder-large_python_snake-case-screaming-snake-case_delta}{-0.51}
\DefMacro{dscoder-large_python_snake-case-screaming-snake-case_trend}{$\downarrow$}
\DefMacro{dscoder-large_python_lparentheses-name}{67.23}
\DefMacro{dscoder-large_python_lparentheses-name_delta}{-0.90}
\DefMacro{dscoder-large_python_lparentheses-name_trend}{$\downarrow$}
\DefMacro{dscoder-large_python_lparentheses-rparentheses}{67.49}
\DefMacro{dscoder-large_python_lparentheses-rparentheses_delta}{-0.64}
\DefMacro{dscoder-large_python_lparentheses-rparentheses_trend}{$\downarrow$}
\DefMacro{dscoder-large_python_rparentheses-colon}{67.10}
\DefMacro{dscoder-large_python_rparentheses-colon_delta}{-1.03}
\DefMacro{dscoder-large_python_rparentheses-colon_trend}{$\downarrow$}
\DefMacro{dscoder-large_python_rparentheses-rparentheses}{67.10}
\DefMacro{dscoder-large_python_rparentheses-rparentheses_delta}{-1.03}
\DefMacro{dscoder-large_python_rparentheses-rparentheses_trend}{$\downarrow$}
\DefMacro{dscoder-large_python_lsquarebracket-name}{67.23}
\DefMacro{dscoder-large_python_lsquarebracket-name_delta}{-0.90}
\DefMacro{dscoder-large_python_lsquarebracket-name_trend}{$\downarrow$}
\DefMacro{dscoder-large_python_period-name}{67.23}
\DefMacro{dscoder-large_python_period-name_delta}{-0.90}
\DefMacro{dscoder-large_python_period-name_trend}{$\downarrow$}
\DefMacro{dscoder-large_python_rsquarebracket-rparentheses}{67.36}
\DefMacro{dscoder-large_python_rsquarebracket-rparentheses_delta}{-0.77}
\DefMacro{dscoder-large_python_rsquarebracket-rparentheses_trend}{$\downarrow$}
\DefMacro{dscoder-large_python_op-lsquarebracket}{67.75}
\DefMacro{dscoder-large_python_op-lsquarebracket_delta}{-0.38}
\DefMacro{dscoder-large_python_op-lsquarebracket_trend}{$\downarrow$}
\DefMacro{dscoder-large_python_op-rsquarebracket}{66.71}
\DefMacro{dscoder-large_python_op-rsquarebracket_delta}{-1.42}
\DefMacro{dscoder-large_python_op-rsquarebracket_trend}{$\downarrow$}
\DefMacro{dscoder-large_python_op-dash}{67.62}
\DefMacro{dscoder-large_python_op-dash_delta}{-0.51}
\DefMacro{dscoder-large_python_op-dash_trend}{$\downarrow$}
\DefMacro{dscoder-large_python_op-name}{66.32}
\DefMacro{dscoder-large_python_op-name_delta}{-1.81}
\DefMacro{dscoder-large_python_op-name_trend}{$\downarrow$}
\DefMacro{dscoder-large_python_op-all}{67.36}
\DefMacro{dscoder-large_python_op-all_delta}{-0.77}
\DefMacro{dscoder-large_python_op-all_trend}{$\downarrow$}

\DefMacro{dscoder-medium_python_baseline}{61.92}
\DefMacro{dscoder-medium_python_snake-case-camel-case}{61.53}
\DefMacro{dscoder-medium_python_snake-case-camel-case_delta}{-0.39}
\DefMacro{dscoder-medium_python_snake-case-camel-case_trend}{$\downarrow$}
\DefMacro{dscoder-medium_python_snake-case-pascal-case}{61.53}
\DefMacro{dscoder-medium_python_snake-case-pascal-case_delta}{-0.39}
\DefMacro{dscoder-medium_python_snake-case-pascal-case_trend}{$\downarrow$}
\DefMacro{dscoder-medium_python_snake-case-screaming-snake-case}{61.14}
\DefMacro{dscoder-medium_python_snake-case-screaming-snake-case_delta}{-0.78}
\DefMacro{dscoder-medium_python_snake-case-screaming-snake-case_trend}{$\downarrow$}
\DefMacro{dscoder-medium_python_lparentheses-name}{61.27}
\DefMacro{dscoder-medium_python_lparentheses-name_delta}{-0.65}
\DefMacro{dscoder-medium_python_lparentheses-name_trend}{$\downarrow$}
\DefMacro{dscoder-medium_python_lparentheses-rparentheses}{61.66}
\DefMacro{dscoder-medium_python_lparentheses-rparentheses_delta}{-0.26}
\DefMacro{dscoder-medium_python_lparentheses-rparentheses_trend}{$\downarrow$}
\DefMacro{dscoder-medium_python_rparentheses-colon}{62.05}
\DefMacro{dscoder-medium_python_rparentheses-colon_delta}{+0.13}
\DefMacro{dscoder-medium_python_rparentheses-colon_trend}{$\uparrow$}
\DefMacro{dscoder-medium_python_rparentheses-rparentheses}{62.05}
\DefMacro{dscoder-medium_python_rparentheses-rparentheses_delta}{+0.13}
\DefMacro{dscoder-medium_python_rparentheses-rparentheses_trend}{$\uparrow$}
\DefMacro{dscoder-medium_python_lsquarebracket-name}{62.69}
\DefMacro{dscoder-medium_python_lsquarebracket-name_delta}{+0.77}
\DefMacro{dscoder-medium_python_lsquarebracket-name_trend}{$\uparrow$}
\DefMacro{dscoder-medium_python_period-name}{61.66}
\DefMacro{dscoder-medium_python_period-name_delta}{-0.26}
\DefMacro{dscoder-medium_python_period-name_trend}{$\downarrow$}
\DefMacro{dscoder-medium_python_rsquarebracket-rparentheses}{61.92}
\DefMacro{dscoder-medium_python_rsquarebracket-rparentheses_delta}{+0.00}
\DefMacro{dscoder-medium_python_rsquarebracket-rparentheses_trend}{}
\DefMacro{dscoder-medium_python_op-lsquarebracket}{62.44}
\DefMacro{dscoder-medium_python_op-lsquarebracket_delta}{+0.52}
\DefMacro{dscoder-medium_python_op-lsquarebracket_trend}{$\uparrow$}
\DefMacro{dscoder-medium_python_op-rsquarebracket}{62.69}
\DefMacro{dscoder-medium_python_op-rsquarebracket_delta}{+0.77}
\DefMacro{dscoder-medium_python_op-rsquarebracket_trend}{$\uparrow$}
\DefMacro{dscoder-medium_python_op-dash}{62.56}
\DefMacro{dscoder-medium_python_op-dash_delta}{+0.64}
\DefMacro{dscoder-medium_python_op-dash_trend}{$\uparrow$}
\DefMacro{dscoder-medium_python_op-name}{60.49}
\DefMacro{dscoder-medium_python_op-name_delta}{-1.43}
\DefMacro{dscoder-medium_python_op-name_trend}{$\downarrow$}
\DefMacro{dscoder-medium_python_op-all}{62.05}
\DefMacro{dscoder-medium_python_op-all_delta}{+0.13}
\DefMacro{dscoder-medium_python_op-all_trend}{$\uparrow$}

\DefMacro{average_python_baseline}{57.14}
\DefMacro{average_python_snake-case-camel-case}{57.43}
\DefMacro{average_python_snake-case-camel-case_delta}{+0.29}
\DefMacro{average_python_snake-case-camel-case_trend}{$\uparrow$}
\DefMacro{average_python_snake-case-pascal-case}{56.85}
\DefMacro{average_python_snake-case-pascal-case_delta}{-0.29}
\DefMacro{average_python_snake-case-pascal-case_trend}{$\downarrow$}
\DefMacro{average_python_snake-case-screaming-snake-case}{56.29}
\DefMacro{average_python_snake-case-screaming-snake-case_delta}{-0.85}
\DefMacro{average_python_snake-case-screaming-snake-case_trend}{$\downarrow$}
\DefMacro{average_python_lparentheses-name}{56.74}
\DefMacro{average_python_lparentheses-name_delta}{-0.40}
\DefMacro{average_python_lparentheses-name_trend}{$\downarrow$}
\DefMacro{average_python_lparentheses-rparentheses}{56.42}
\DefMacro{average_python_lparentheses-rparentheses_delta}{-0.72}
\DefMacro{average_python_lparentheses-rparentheses_trend}{$\downarrow$}
\DefMacro{average_python_rparentheses-colon}{57.12}
\DefMacro{average_python_rparentheses-colon_delta}{-0.02}
\DefMacro{average_python_rparentheses-colon_trend}{$\downarrow$}
\DefMacro{average_python_rparentheses-rparentheses}{56.98}
\DefMacro{average_python_rparentheses-rparentheses_delta}{-0.16}
\DefMacro{average_python_rparentheses-rparentheses_trend}{$\downarrow$}
\DefMacro{average_python_lsquarebracket-name}{56.92}
\DefMacro{average_python_lsquarebracket-name_delta}{-0.22}
\DefMacro{average_python_lsquarebracket-name_trend}{$\downarrow$}
\DefMacro{average_python_period-name}{56.52}
\DefMacro{average_python_period-name_delta}{-0.62}
\DefMacro{average_python_period-name_trend}{$\downarrow$}
\DefMacro{average_python_rsquarebracket-rparentheses}{56.90}
\DefMacro{average_python_rsquarebracket-rparentheses_delta}{-0.24}
\DefMacro{average_python_rsquarebracket-rparentheses_trend}{$\downarrow$}
\DefMacro{average_python_op-lsquarebracket}{57.11}
\DefMacro{average_python_op-lsquarebracket_delta}{-0.03}
\DefMacro{average_python_op-lsquarebracket_trend}{$\downarrow$}
\DefMacro{average_python_op-rsquarebracket}{57.11}
\DefMacro{average_python_op-rsquarebracket_delta}{-0.03}
\DefMacro{average_python_op-rsquarebracket_trend}{$\downarrow$}
\DefMacro{average_python_op-dash}{57.22}
\DefMacro{average_python_op-dash_delta}{+0.08}
\DefMacro{average_python_op-dash_trend}{$\uparrow$}
\DefMacro{average_python_op-name}{56.19}
\DefMacro{average_python_op-name_delta}{-0.95}
\DefMacro{average_python_op-name_trend}{$\downarrow$}
\DefMacro{average_python_op-all}{55.93}
\DefMacro{average_python_op-all_delta}{-1.21}
\DefMacro{average_python_op-all_trend}{$\downarrow$}

\DefMacro{llama-medium_all-naming_sensitivity}{10.68} 
\DefMacro{llama-medium_all-spacing_sensitivity}{10.99} 
\DefMacro{llama-medium_fragment-changed-naming_sensitivity}{12.37} 
\DefMacro{llama-medium_fragment-changed-spacing_sensitivity}{14.45} 
\DefMacro{llama-medium_fragment-unchanged-naming_sensitivity}{9.44} 
\DefMacro{llama-medium_fragment-unchanged-spacing_sensitivity}{9.69} 
\DefMacro{llama-medium_merging-naming_sensitivity}{8.57} 
\DefMacro{llama-medium_merging-spacing_sensitivity}{13.03} 
\DefMacro{llama-medium_splitting-naming_sensitivity}{13.30} 
\DefMacro{llama-medium_splitting-spacing_sensitivity}{14.83} 
\DefMacro{llama-medium_mixing-naming_sensitivity}{8.11} 
\DefMacro{llama-medium_mixing-spacing_sensitivity}{14.48} 

\DefMacro{llama-small_all-naming_sensitivity}{11.48} 
\DefMacro{llama-small_all-spacing_sensitivity}{10.22} 
\DefMacro{llama-small_fragment-changed-naming_sensitivity}{12.58} 
\DefMacro{llama-small_fragment-changed-spacing_sensitivity}{12.61} 
\DefMacro{llama-small_fragment-unchanged-naming_sensitivity}{10.68} 
\DefMacro{llama-small_fragment-unchanged-spacing_sensitivity}{9.32} 
\DefMacro{llama-small_merging-naming_sensitivity}{10.48} 
\DefMacro{llama-small_merging-spacing_sensitivity}{11.06} 
\DefMacro{llama-small_splitting-naming_sensitivity}{13.17} 
\DefMacro{llama-small_splitting-spacing_sensitivity}{13.37} 
\DefMacro{llama-small_mixing-naming_sensitivity}{9.46} 
\DefMacro{llama-small_mixing-spacing_sensitivity}{10.80} 

\DefMacro{llama-large_all-naming_sensitivity}{9.43} 
\DefMacro{llama-large_all-spacing_sensitivity}{8.51} 
\DefMacro{llama-large_fragment-changed-naming_sensitivity}{11.21} 
\DefMacro{llama-large_fragment-changed-spacing_sensitivity}{11.89} 
\DefMacro{llama-large_fragment-unchanged-naming_sensitivity}{8.13} 
\DefMacro{llama-large_fragment-unchanged-spacing_sensitivity}{7.24} 
\DefMacro{llama-large_merging-naming_sensitivity}{9.52} 
\DefMacro{llama-large_merging-spacing_sensitivity}{10.00} 
\DefMacro{llama-large_splitting-naming_sensitivity}{11.73} 
\DefMacro{llama-large_splitting-spacing_sensitivity}{11.53} 
\DefMacro{llama-large_mixing-naming_sensitivity}{8.11} 
\DefMacro{llama-large_mixing-spacing_sensitivity}{16.78} 

\DefMacro{qwen-medium_all-naming_sensitivity}{7.95} 
\DefMacro{qwen-medium_all-spacing_sensitivity}{8.87} 
\DefMacro{qwen-medium_fragment-changed-naming_sensitivity}{8.77} 
\DefMacro{qwen-medium_fragment-changed-spacing_sensitivity}{12.47} 
\DefMacro{qwen-medium_fragment-unchanged-naming_sensitivity}{7.35} 
\DefMacro{qwen-medium_fragment-unchanged-spacing_sensitivity}{7.53} 
\DefMacro{qwen-medium_merging-naming_sensitivity}{5.71} 
\DefMacro{qwen-medium_merging-spacing_sensitivity}{12.42} 
\DefMacro{qwen-medium_splitting-naming_sensitivity}{9.00} 
\DefMacro{qwen-medium_splitting-spacing_sensitivity}{10.71} 
\DefMacro{qwen-medium_mixing-naming_sensitivity}{10.81} 
\DefMacro{qwen-medium_mixing-spacing_sensitivity}{22.15} 

\DefMacro{qwen-small_all-naming_sensitivity}{7.73} 
\DefMacro{qwen-small_all-spacing_sensitivity}{7.07} 
\DefMacro{qwen-small_fragment-changed-naming_sensitivity}{8.77} 
\DefMacro{qwen-small_fragment-changed-spacing_sensitivity}{9.80} 
\DefMacro{qwen-small_fragment-unchanged-naming_sensitivity}{6.97} 
\DefMacro{qwen-small_fragment-unchanged-spacing_sensitivity}{6.04} 
\DefMacro{qwen-small_merging-naming_sensitivity}{8.57} 
\DefMacro{qwen-small_merging-spacing_sensitivity}{8.33} 
\DefMacro{qwen-small_splitting-naming_sensitivity}{9.00} 
\DefMacro{qwen-small_splitting-spacing_sensitivity}{9.62} 
\DefMacro{qwen-small_mixing-naming_sensitivity}{6.76} 
\DefMacro{qwen-small_mixing-spacing_sensitivity}{13.01} 

\DefMacro{qwen-large_all-naming_sensitivity}{8.27} 
\DefMacro{qwen-large_all-spacing_sensitivity}{5.71} 
\DefMacro{qwen-large_fragment-changed-naming_sensitivity}{10.57} 
\DefMacro{qwen-large_fragment-changed-spacing_sensitivity}{7.37} 
\DefMacro{qwen-large_fragment-unchanged-naming_sensitivity}{6.58} 
\DefMacro{qwen-large_fragment-unchanged-spacing_sensitivity}{5.09} 
\DefMacro{qwen-large_merging-naming_sensitivity}{11.43} 
\DefMacro{qwen-large_merging-spacing_sensitivity}{7.42} 
\DefMacro{qwen-large_splitting-naming_sensitivity}{10.82} 
\DefMacro{qwen-large_splitting-spacing_sensitivity}{6.48} 
\DefMacro{qwen-large_mixing-naming_sensitivity}{6.76} 
\DefMacro{qwen-large_mixing-spacing_sensitivity}{12.10} 

\DefMacro{dscoder-small_all-naming_sensitivity}{9.88} 
\DefMacro{dscoder-small_all-spacing_sensitivity}{8.36} 
\DefMacro{dscoder-small_fragment-changed-naming_sensitivity}{11.14} 
\DefMacro{dscoder-small_fragment-changed-spacing_sensitivity}{10.47} 
\DefMacro{dscoder-small_fragment-unchanged-naming_sensitivity}{8.31} 
\DefMacro{dscoder-small_fragment-unchanged-spacing_sensitivity}{7.25} 
\DefMacro{dscoder-small_merging-naming_sensitivity}{4.88} 
\DefMacro{dscoder-small_merging-spacing_sensitivity}{8.72} 
\DefMacro{dscoder-small_splitting-naming_sensitivity}{11.95} 
\DefMacro{dscoder-small_splitting-spacing_sensitivity}{10.45} 
\DefMacro{dscoder-small_mixing-naming_sensitivity}{10.85} 
\DefMacro{dscoder-small_mixing-spacing_sensitivity}{12.02} 

\DefMacro{dscoder-medium_all-naming_sensitivity}{8.95} 
\DefMacro{dscoder-medium_all-spacing_sensitivity}{8.71} 
\DefMacro{dscoder-medium_fragment-changed-naming_sensitivity}{9.78} 
\DefMacro{dscoder-medium_fragment-changed-spacing_sensitivity}{10.85} 
\DefMacro{dscoder-medium_fragment-unchanged-naming_sensitivity}{7.91} 
\DefMacro{dscoder-medium_fragment-unchanged-spacing_sensitivity}{7.58} 
\DefMacro{dscoder-medium_merging-naming_sensitivity}{7.32} 
\DefMacro{dscoder-medium_merging-spacing_sensitivity}{10.36} 
\DefMacro{dscoder-medium_splitting-naming_sensitivity}{10.54} 
\DefMacro{dscoder-medium_splitting-spacing_sensitivity}{10.19} 
\DefMacro{dscoder-medium_mixing-naming_sensitivity}{6.20} 
\DefMacro{dscoder-medium_mixing-spacing_sensitivity}{14.09} 

\DefMacro{dscoder-large_all-naming_sensitivity}{8.95} 
\DefMacro{dscoder-large_all-spacing_sensitivity}{6.26} 
\DefMacro{dscoder-large_fragment-changed-naming_sensitivity}{10.82} 
\DefMacro{dscoder-large_fragment-changed-spacing_sensitivity}{7.12} 
\DefMacro{dscoder-large_fragment-unchanged-naming_sensitivity}{6.61} 
\DefMacro{dscoder-large_fragment-unchanged-spacing_sensitivity}{5.80} 
\DefMacro{dscoder-large_merging-naming_sensitivity}{10.57} 
\DefMacro{dscoder-large_merging-spacing_sensitivity}{6.41} 
\DefMacro{dscoder-large_splitting-naming_sensitivity}{10.64} 
\DefMacro{dscoder-large_splitting-spacing_sensitivity}{6.44} 
\DefMacro{dscoder-large_mixing-naming_sensitivity}{12.40} 
\DefMacro{dscoder-large_mixing-spacing_sensitivity}{10.64} 

\DefMacro{codeqwen_all-naming_sensitivity}{8.86} 
\DefMacro{codeqwen_all-spacing_sensitivity}{7.03} 
\DefMacro{codeqwen_fragment-changed-naming_sensitivity}{9.82} 
\DefMacro{codeqwen_fragment-changed-spacing_sensitivity}{9.16} 
\DefMacro{codeqwen_fragment-unchanged-naming_sensitivity}{7.55} 
\DefMacro{codeqwen_fragment-unchanged-spacing_sensitivity}{6.03} 
\DefMacro{codeqwen_merging-naming_sensitivity}{5.15} 
\DefMacro{codeqwen_merging-spacing_sensitivity}{8.97} 
\DefMacro{codeqwen_splitting-naming_sensitivity}{10.07} 
\DefMacro{codeqwen_splitting-spacing_sensitivity}{8.51} 
\DefMacro{codeqwen_mixing-naming_sensitivity}{13.16} 
\DefMacro{codeqwen_mixing-spacing_sensitivity}{11.23} 

\DefMacro{qwen-base_all-naming_sensitivity}{5.32} 
\DefMacro{qwen-base_all-spacing_sensitivity}{6.32} 
\DefMacro{qwen-base_fragment-changed-naming_sensitivity}{5.60} 
\DefMacro{qwen-base_fragment-changed-spacing_sensitivity}{8.17} 
\DefMacro{qwen-base_fragment-unchanged-naming_sensitivity}{5.11} 
\DefMacro{qwen-base_fragment-unchanged-spacing_sensitivity}{5.63} 
\DefMacro{qwen-base_merging-naming_sensitivity}{1.90} 
\DefMacro{qwen-base_merging-spacing_sensitivity}{8.79} 
\DefMacro{qwen-base_splitting-naming_sensitivity}{5.74} 
\DefMacro{qwen-base_splitting-spacing_sensitivity}{7.53} 
\DefMacro{qwen-base_mixing-naming_sensitivity}{9.46} 
\DefMacro{qwen-base_mixing-spacing_sensitivity}{10.73}

\DefMacro{python_lparentheses-name_github-frequency}{360450048}
\DefMacro{python_lparentheses-name_spaced_github-frequency}{3960928}
\DefMacro{python_lparentheses-name_github_frequency_ratio}{1.1}
\DefMacro{python_lparentheses-name_github-frequency_w}{105M}
\DefMacro{python_lparentheses-name_spaced_github-frequency_w}{2.9M}
\DefMacro{python_lparentheses-name_github-frequency_ratio_w}{2.76}

\DefMacro{python_lparentheses-rparentheses_github-frequency}{14884864}
\DefMacro{python_lparentheses-rparentheses_spaced_github-frequency}{55216}
\DefMacro{python_lparentheses-rparentheses_github_frequency_ratio}{0.37}
\DefMacro{python_lparentheses-rparentheses_github-frequency_w}{78.1M}
\DefMacro{python_lparentheses-rparentheses_spaced_github-frequency_w}{71.7K}
\DefMacro{python_lparentheses-rparentheses_github-frequency_ratio_w}{0.09}

\DefMacro{python_rparentheses-colon_github-frequency}{16793600}
\DefMacro{python_rparentheses-colon_spaced_github-frequency}{312832}
\DefMacro{python_rparentheses-colon_github_frequency_ratio}{1.86}
\DefMacro{python_rparentheses-colon_github-frequency_w}{76M}
\DefMacro{python_rparentheses-colon_spaced_github-frequency_w}{1.4M}
\DefMacro{python_rparentheses-colon_github-frequency_ratio_w}{1.84}

\DefMacro{python_rparentheses-rparentheses_github-frequency}{12480512}
\DefMacro{python_rparentheses-rparentheses_spaced_github-frequency}{393216}
\DefMacro{python_rparentheses-rparentheses_github_frequency_ratio}{3.15}
\DefMacro{python_rparentheses-rparentheses_github-frequency_w}{59M}
\DefMacro{python_rparentheses-rparentheses_spaced_github-frequency_w}{2.4M}
\DefMacro{python_rparentheses-rparentheses_github-frequency_ratio_w}{4.07}

\DefMacro{python_lsquarebracket-name_github-frequency}{79758720}
\DefMacro{python_lsquarebracket-name_spaced_github-frequency}{1370816}
\DefMacro{python_lsquarebracket-name_github_frequency_ratio}{1.72}
\DefMacro{python_lsquarebracket-name_github-frequency_w}{61.1M}
\DefMacro{python_lsquarebracket-name_spaced_github-frequency_w}{1.1M}
\DefMacro{python_lsquarebracket-name_github-frequency_ratio_w}{1.83}

\DefMacro{python_period-name_github-frequency}{0}
\DefMacro{python_period-name_spaced_github-frequency}{0}
\DefMacro{python_period-name_github_frequency_ratio}{0}
\DefMacro{python_period-name_github-frequency_w}{107M}
\DefMacro{python_period-name_spaced_github-frequency_w}{40.6M}
\DefMacro{python_period-name_github-frequency_ratio_w}{37.94}

\DefMacro{python_rsquarebracket-rparentheses_github-frequency}{11280384}
\DefMacro{python_rsquarebracket-rparentheses_spaced_github-frequency}{190976}
\DefMacro{python_rsquarebracket-rparentheses_github_frequency_ratio}{1.69}
\DefMacro{python_rsquarebracket-rparentheses_github-frequency_w}{44.6M}
\DefMacro{python_rsquarebracket-rparentheses_spaced_github-frequency_w}{1.7M}
\DefMacro{python_rsquarebracket-rparentheses_github-frequency_ratio_w}{3.81}

\DefMacro{python_op-lsquarebracket_github-frequency}{23875530}
\DefMacro{python_op-lsquarebracket_spaced_github-frequency}{17750079}
\DefMacro{python_op-lsquarebracket_github_frequency_ratio}{74.34}

\DefMacro{python_op-rsquarebracket_github-frequency}{25100786}
\DefMacro{python_op-rsquarebracket_spaced_github-frequency}{614603}
\DefMacro{python_op-rsquarebracket_github_frequency_ratio}{2.45}

\DefMacro{python_op-dash_github-frequency}{11922469}
\DefMacro{python_op-dash_spaced_github-frequency}{13654816}
\DefMacro{python_op-dash_github_frequency_ratio}{114.53}

\DefMacro{python_op-name_github-frequency}{657323369}
\DefMacro{python_op-name_spaced_github-frequency}{816290410}
\DefMacro{python_op-name_github_frequency_ratio}{124.18}

\DefMacro{java_rparentheses-semicolon_github-frequency}{8527872}
\DefMacro{java_rparentheses-semicolon_spaced_github-frequency}{189440}
\DefMacro{java_rparentheses-semicolon_github_frequency_ratio}{2.22}
\DefMacro{java_rparentheses-semicolon_github-frequency_w}{161M}
\DefMacro{java_rparentheses-semicolon_spaced_github-frequency_w}{924K}
\DefMacro{java_rparentheses-semicolon_github-frequency_ratio_w}{0.57}

\DefMacro{java_period-name_github-frequency}{0}
\DefMacro{java_period-name_spaced_github-frequency}{0}
\DefMacro{java_period-name_github_frequency_ratio}{0}
\DefMacro{java_period-name_github-frequency_w}{175M}
\DefMacro{java_period-name_spaced_github-frequency_w}{45.9M}
\DefMacro{java_period-name_github-frequency_ratio_w}{16.22}

\DefMacro{java_lparentheses-rparentheses_github-frequency}{16015360}
\DefMacro{java_lparentheses-rparentheses_spaced_github-frequency}{43776}
\DefMacro{java_lparentheses-rparentheses_github_frequency_ratio}{0.27}
\DefMacro{java_lparentheses-rparentheses_github-frequency_w}{144M}
\DefMacro{java_lparentheses-rparentheses_spaced_github-frequency_w}{195K}
\DefMacro{java_lparentheses-rparentheses_github-frequency_ratio_w}{0.14}

\DefMacro{java_lparentheses-name_github-frequency}{52593280}
\DefMacro{java_lparentheses-name_spaced_github-frequency}{2026912}
\DefMacro{java_lparentheses-name_github_frequency_ratio}{3.85}
\DefMacro{java_lparentheses-name_github-frequency_w}{172M}
\DefMacro{java_lparentheses-name_spaced_github-frequency_w}{6.6M}
\DefMacro{java_lparentheses-name_github-frequency_ratio_w}{3.84}

\DefMacro{java_rparentheses-rparentheses_github-frequency}{6696960}
\DefMacro{java_rparentheses-rparentheses_spaced_github-frequency}{520192}
\DefMacro{java_rparentheses-rparentheses_github_frequency_ratio}{7.77}
\DefMacro{java_rparentheses-rparentheses_github-frequency_w}{102M}
\DefMacro{java_rparentheses-rparentheses_spaced_github-frequency_w}{3.4M}
\DefMacro{java_rparentheses-rparentheses_github-frequency_ratio_w}{3.33}

\DefMacro{java_period-asterisk_github-frequency}{176656}
\DefMacro{java_period-asterisk_spaced_github-frequency}{911360}
\DefMacro{java_period-asterisk_github_frequency_ratio}{515.9}
\DefMacro{java_period-asterisk_github-frequency_w}{34.2M}
\DefMacro{java_period-asterisk_spaced_github-frequency_w}{7.3M}
\DefMacro{java_period-asterisk_github-frequency_ratio_w}{21.35}

\DefMacro{java_double-plus-rparentheses_github-frequency}{1597440}
\DefMacro{java_double-plus-rparentheses_spaced_github-frequency}{52320}
\DefMacro{java_double-plus-rparentheses_github_frequency_ratio}{3.28}
\DefMacro{java_double-plus-rparentheses_github-frequency_w}{22.9M}
\DefMacro{java_double-plus-rparentheses_spaced_github-frequency_w}{664K}
\DefMacro{java_double-plus-rparentheses_github-frequency_ratio_w}{2.90}

\DefMacro{java_rparentheses-period_github-frequency}{4411392}
\DefMacro{java_rparentheses-period_spaced_github-frequency}{21280}
\DefMacro{java_rparentheses-period_github_frequency_ratio}{0.48}
\DefMacro{java_rparentheses-period_github-frequency_w}{78.9M}
\DefMacro{java_rparentheses-period_spaced_github-frequency_w}{45.7K}
\DefMacro{java_rparentheses-period_github-frequency_ratio_w}{0.06}

\DefMacro{java_op-semicolon_github-frequency}{10082513}
\DefMacro{java_op-semicolon_spaced_github-frequency}{212409}
\DefMacro{java_op-semicolon_github_frequency_ratio}{2.11}

\DefMacro{java_op-lparentheses_github-frequency}{2421040}
\DefMacro{java_op-lparentheses_spaced_github-frequency}{5540404}
\DefMacro{java_op-lparentheses_github_frequency_ratio}{228.84}

\DefMacro{java_op-name_github-frequency}{56309838}
\DefMacro{java_op-name_spaced_github-frequency}{12791608}
\DefMacro{java_op-name_github_frequency_ratio}{22.72}

\DefMacro{camel-snake-example}{\texttt{sortedLst}$\rightarrow$\texttt{sorted\_lst}}
\DefMacro{camel-snake-example-tok}{[\texttt{'\textvisiblespace sorted','L','st'}]$\rightarrow$[\texttt{'\textvisiblespace sorted','\_lst'}]}
\DefMacro{camel-snake-example-lang}{\UseMacro{only-java}\UseMacro{camel-snake-example-tok}}
\DefMacro{camel-snake-example-colorful}{\TokenOld{\textvisiblespace sorted}\TokenOld{L}\TokenOld{st}$\rightarrow$\TokenNew{\textvisiblespace sorted}\TokenNew{\_lst}}

\DefMacro{camel-pascal-example}{\texttt{cloestPair}$\rightarrow$\texttt{ClosestPair}}
\DefMacro{camel-pascal-example-tok}{[\texttt{'\textvisiblespace cloestPair'}]$\rightarrow$[\texttt{'\textvisiblespace Close','st','Pair'}]}
\DefMacro{camel-pascal-example-lang}{\UseMacro{only-java}\UseMacro{camel-pascal-example-tok}}
\DefMacro{camel-pascal-example-colorful}{\TokenOld{\textvisiblespace cloestPair}$\rightarrow$\TokenNew{\textvisiblespace Close}\TokenNew{st}\TokenNew{Pair}}

\DefMacro{camel-screaming-snake-example}{\texttt{possibleSolutions}$\rightarrow$\texttt{POSSIBLE\_SOLUTIONS}}
\DefMacro{camel-screaming-snake-example-tok}{[\texttt{'\textvisiblespace possible','S','olutions'}]$\rightarrow$[\texttt{'\textvisiblespace POSS','IBLE','\_S','OLUTION','S'}]}
\DefMacro{camel-screaming-snake-example-lang}{\UseMacro{only-java}\UseMacro{camel-screaming-snake-example-tok}}
\DefMacro{camel-screaming-snake-example-colorful}{\TokenOld{\textvisiblespace possible}\TokenOld{S}\TokenOld{olutions}$\rightarrow$\TokenNew{\textvisiblespace POSS}\TokenNew{IBLE}\TokenNew{\_S}\TokenNew{OLUTION}\TokenNew{S}}

\DefMacro{snake-camel-example}{\texttt{input\_clipboard}$\rightarrow$\texttt{inputClipboard}}
\DefMacro{snake-camel-example-tok}{[\texttt{'\textvisiblespace input','\_clip','board'}]$\rightarrow$[\texttt{'\textvisiblespace input','Clipboard'}]}
\DefMacro{snake-camel-example-lang}{\UseMacro{only-python}\UseMacro{snake-camel-example-tok}}
\DefMacro{snake-camel-example-colorful}{\TokenOld{\textvisiblespace input}\TokenOld{\_clip}\TokenOld{board}$\rightarrow$\TokenNew{\textvisiblespace input}\TokenNew{Clipboard}}

\DefMacro{snake-pascal-example}{\texttt{string\_xor}$\rightarrow$\texttt{StringXor}}
\DefMacro{snake-pascal-example-tok}{[\texttt{'\textvisiblespace string','\_xor'}]$\rightarrow$[\texttt{'\textvisiblespace String','X','or'}]}
\DefMacro{snake-pascal-example-lang}{\UseMacro{only-python}\UseMacro{snake-pascal-example-tok}}
\DefMacro{snake-pascal-example-colorful}{\TokenOld{\textvisiblespace string}\TokenOld{\_xor}$\rightarrow$\TokenNew{\textvisiblespace String}\TokenNew{X}\TokenNew{or}}

\DefMacro{snake-screaming-snake-example}{\texttt{triangle\_area}$\rightarrow$\texttt{TRIANGLE\_AREA}}
\DefMacro{snake-screaming-snake-example-tok}{[\texttt{'\textvisiblespace triangle','\_area'}]$\rightarrow$[\texttt{'\textvisiblespace TRI','ANGLE','\_AREA'}]}
\DefMacro{snake-screaming-snake-example-lang}{\UseMacro{only-python}\UseMacro{snake-screaming-snake-example-tok}}
\DefMacro{snake-screaming-snake-example-colorful}{\TokenOld{\textvisiblespace triangle}\TokenOld{\_area}$\rightarrow$\TokenNew{\textvisiblespace TRI}\TokenNew{ANGLE}\TokenNew{\_AREA}}

\DefMacro{op-dash-example}{\texttt{[::-}$\rightarrow$\texttt{[::\textvisiblespace -}}
\DefMacro{op-dash-example-tok}{[\texttt{'[::-','1', ']'}]$\rightarrow$[\texttt{'[','::','\textvisiblespace -','1', ']'}]}
\DefMacro{op-dash-example-lang}{\UseMacro{only-python}\UseMacro{op-dash-example-tok}}
\DefMacro{op-dash-example-colorful}{\TokenOld{[::-}\TokenCtx{1}\TokenCtx{]}$\rightarrow$\TokenNew{[}\TokenNew{::}\TokenNew{\textvisiblespace -}\TokenCtx{1}\TokenCtx{]}}

\DefMacro{op-lsquarebracket-example}{\texttt{)[}$\rightarrow$\texttt{)\textvisiblespace [}}
\DefMacro{op-lsquarebracket-example-tok}{[\texttt{'(N','))',')[', '2', ':]\textbackslash n'}]$\rightarrow$[\texttt{'(N', ')))', '\textvisiblespace [', '2', ':]\textbackslash n'}]}
\DefMacro{op-lsquarebracket-example-lang}{\UseMacro{only-python}\UseMacro{op-lsquarebracket-example-tok}}
\DefMacro{op-lsquarebracket-example-colorful}{%
\TokenOld{))}\TokenOld{)[}\TokenCtx{2}\TokenCtx{:]\textbackslash n}%
$\rightarrow$%
\TokenNew{)))}\TokenNew{\textvisiblespace [}\TokenCtx{2}\TokenCtx{:]\textbackslash n}%
}

\DefMacro{rparentheses-period-example}{\texttt{).}$\rightarrow$\texttt{)\textvisiblespace .}}
\DefMacro{rparentheses-period-example-tok}{[\texttt{')','.'}]$\rightarrow$[\texttt{')', '\textvisiblespace .'}]}
\DefMacro{rparentheses-period-example-lang}{\UseMacro{only-java}\UseMacro{rparentheses-period-example-tok}}
\DefMacro{rparentheses-period-example-colorful}{%
\TokenOld{\textvisiblespace\textquotesingle.}\TokenOld{\textquotesingle).}\TokenCtx{replace}%
$\rightarrow$%
\TokenNew{\textvisiblespace\textquotesingle.\textquotesingle)}\TokenNew{\textvisiblespace .}\TokenCtx{replace}%
}

\DefMacro{rsquarebracket-rparentheses-example}{\texttt{]):\textbackslash n}$\rightarrow$\texttt{])\textvisiblespace :\textbackslash n}}
\DefMacro{rsquarebracket-rparentheses-example-tok}{[\texttt{':', ']):\textbackslash n'}]$\rightarrow$[\texttt{':]', '\textvisiblespace):\textbackslash n'}]}
\DefMacro{rsquarebracket-rparentheses-example-lang}{\UseMacro{only-python}\UseMacro{rsquarebracket-rparentheses-example-tok}}
\DefMacro{rsquarebracket-rparentheses-example-colorful}{%
\TokenOld{:}\TokenOld{]):\textbackslash n}%
$\rightarrow$%
\TokenNew{:]}\TokenNew{\textvisiblespace ):\textbackslash n}%
}

\DefMacro{op-rsquarebracket-example}{\texttt{[[]}$\rightarrow$\texttt{[[\textvisiblespace ]}}
\DefMacro{op-rsquarebracket-example-tok}{[\texttt{'\textvisiblespace [[]'}]$\rightarrow$[\texttt{'\textvisiblespace [[','\textvisiblespace]'}]}
\DefMacro{op-rsquarebracket-example-lang}{\UseMacro{only-python}\UseMacro{op-rsquarebracket-example-tok}}
\DefMacro{op-rsquarebracket-example-colorful}{%
\TokenCtx{=}\TokenOld{\textvisiblespace [[]}%
$\rightarrow$%
\TokenCtx{=}\TokenNew{\textvisiblespace [[}\TokenNew{\textvisiblespace ]}%
}

\DefMacro{op-lparentheses-example}{\texttt{((!}$\rightarrow$\texttt{(\textvisiblespace (!}}
\DefMacro{op-lparentheses-example-tok}{[\texttt{'((','!'}]$\rightarrow$[\texttt{'(', '\textvisiblespace (!'}]}
\DefMacro{op-lparentheses-example-lang}{\UseMacro{only-java}\UseMacro{op-lparentheses-example-tok}}
\DefMacro{op-lparentheses-example-colorful}{%
\TokenOld{((}\TokenOld{!}\TokenCtx{is}\TokenCtx{True}%
$\rightarrow$%
\TokenNew{(}\TokenNew{\textvisiblespace (!}\TokenCtx{is}\TokenCtx{True}%
}

\DefMacro{lsquarebracket-name-example}{\texttt{([vowels}$\rightarrow$\texttt{([\textvisiblespace vowels}}
\DefMacro{lsquarebracket-name-example-tok}{[\texttt{'([', 'v', 'ow', 'els'}]$\rightarrow$[\texttt{'([', '\textvisiblespace vowels'}]}
\DefMacro{lsquarebracket-name-example-lang}{\UseMacro{only-python}\UseMacro{lsquarebracket-name-example-tok}}
\DefMacro{lsquarebracket-name-example-colorful}{%
\TokenCtx{([}\TokenOld{v}\TokenOld{ow}\TokenOld{els}%
$\rightarrow$%
\TokenCtx{([}\TokenNew{\textvisiblespace vowels}%
}

\DefMacro{double-plus-rparentheses-example}{\texttt{++)}$\rightarrow$\texttt{++\textvisiblespace )}}
\DefMacro{double-plus-rparentheses-example-tok}{[\texttt{'++)'}]$\rightarrow$[\texttt{'++', '\textvisiblespace )'}]}
\DefMacro{double-plus-rparentheses-example-lang}{\UseMacro{only-java}\UseMacro{double-plus-rparentheses-example-tok}}
\DefMacro{double-plus-rparentheses-example-colorful}{%
\TokenCtx{\textvisiblespace i}\TokenOld{++)}%
$\rightarrow$%
\TokenCtx{\textvisiblespace i}\TokenNew{++}\TokenNew{\textvisiblespace )}%
}

\DefMacro{period-asterisk-example}{\texttt{.*}$\rightarrow$\texttt{.\textvisiblespace *}}
\DefMacro{period-asterisk-example-tok}{[\texttt{'\textvisiblespace java', '.util', '.*;\textbackslash n'}]$\rightarrow$[\texttt{'\textvisiblespace java', '.util', '.','\textvisiblespace *',';\textbackslash n'}]}
\DefMacro{period-asterisk-example-lang}{\UseMacro{only-java}\UseMacro{period-asterisk-example-tok}}
\DefMacro{period-asterisk-example-colorful}{%
\TokenOld{.*;\textbackslash n}%
$\rightarrow$%
\TokenNew{.}\TokenNew{\textvisiblespace *}\TokenNew{;\textbackslash n}%
}

\DefMacro{rparentheses-colon-example}{\texttt{():}$\rightarrow$\texttt{()\textvisiblespace :}}
\DefMacro{rparentheses-colon-example-tok}{[\texttt{'def', '\textvisiblespace main', '(): \textbackslash n'}]$\rightarrow$[\texttt{'def', '\textvisiblespace main', '()', '\textvisiblespace :\textbackslash n'}]}
\DefMacro{rparentheses-colon-example-lang}{\UseMacro{only-python}\UseMacro{rparentheses-colon-example-tok}}
\DefMacro{rparentheses-colon-example-colorful}{%
\TokenCtx{\textvisiblespace main}\TokenOld{():\textbackslash n}%
$\rightarrow$%
\TokenCtx{\textvisiblespace main}\TokenNew{()}\TokenNew{\textvisiblespace :\textbackslash n}%
}

\DefMacro{rparentheses-semicolon-example}{\texttt{<>();}$\rightarrow$\texttt{<>()\textvisiblespace;}}
\DefMacro{rparentheses-semicolon-example-tok}{[\texttt{'<>();\textbackslash n'}]$\rightarrow$[\texttt{'<', '>()', '\textvisiblespace ;\textbackslash n'}]}
\DefMacro{rparentheses-semicolon-example-lang}{\UseMacro{only-java}\UseMacro{rparentheses-semicolon-example-tok}}
\DefMacro{rparentheses-semicolon-example-colorful}{%
\TokenOld{<>();\textbackslash n}%
$\rightarrow$%
\TokenNew{<}\TokenNew{>()}\TokenNew{\textvisiblespace ;\textbackslash n}%
}

\DefMacro{period-name-example}{\texttt{.factor}$\rightarrow$\texttt{.\textvisiblespace factor}}
\DefMacro{period-name-example-tok}{[\texttt{'.factor', 'ial'}]$\rightarrow$[\texttt{'.', '\textvisiblespace factorial'}]}
\DefMacro{period-name-example-lang}{\UseMacro{both}\UseMacro{period-name-example-tok}}
\DefMacro{period-name-example-colorful}{%
\TokenOld{.factor}\TokenOld{ial}%
$\rightarrow$%
\TokenNew{.}\TokenNew{\textvisiblespace factorial}%
}

\DefMacro{lparentheses-rparentheses-example}{\texttt{()}$\rightarrow$\texttt{()\textvisiblespace }}
\DefMacro{lparentheses-rparentheses-example-tok}{[\texttt{'()'}]$\rightarrow$[\texttt{'(', '\textvisiblespace)'}]}
\DefMacro{lparentheses-rparentheses-example-lang}{\UseMacro{both}\UseMacro{lparentheses-rparentheses-example-tok}}
\DefMacro{lparentheses-rparentheses-example-colorful}{%
\TokenCtx{alpha}\TokenOld{()}%
$\rightarrow$%
\TokenCtx{alpha}\TokenNew{(}\TokenNew{\textvisiblespace )}%
}

\DefMacro{lparentheses-name-example}{\texttt{(String}$\rightarrow$\texttt{(\textvisiblespace String}}
\DefMacro{lparentheses-name-example-tok}{[\texttt{'(String'}]$\rightarrow$[\texttt{'(', '\textvisiblespace String'}]}
\DefMacro{lparentheses-name-example-lang}{\UseMacro{both}\UseMacro{lparentheses-name-example-tok}}
\DefMacro{lparentheses-name-example-colorful}{%
\TokenOld{(String}%
$\rightarrow$%
\TokenNew{(}\TokenNew{\textvisiblespace String}%
}

\DefMacro{rparentheses-rparentheses-example}{\texttt{()}$\rightarrow$\texttt{(\textvisiblespace )}}
\DefMacro{rparentheses-rparentheses-example-tok}{[\texttt{'())'}]$\rightarrow$[\texttt{'()', '\textvisiblespace )'}]}
\DefMacro{rparentheses-rparentheses-example-lang}{\UseMacro{both}\UseMacro{rparentheses-rparentheses-example-tok}}
\DefMacro{rparentheses-rparentheses-example-colorful}{%
\TokenCtx{.toCharArray}\TokenOld{())}%
$\rightarrow$%
\TokenCtx{.toCharArray}\TokenNew{()}\TokenNew{\textvisiblespace )}%
}

\DefMacro{op-semicolon-example}{\texttt{++;}$\rightarrow$\texttt{++\textvisiblespace ;}}
\DefMacro{op-semicolon-example-tok}{[\texttt{'++;'}]$\rightarrow$[\texttt{'++', '\textvisiblespace ;'}]}
\DefMacro{op-semicolon-example-lang}{\UseMacro{only-java}\UseMacro{op-semicolon-example-tok}}
\DefMacro{op-semicolon-example-colorful}{%
\TokenCtx{Ac}\TokenOld{++;}%
$\rightarrow$%
\TokenCtx{Ac}\TokenNew{++}\TokenNew{\textvisiblespace ;}%
}

\DefMacro{op-name-example}{\texttt{[i:i+len(substring)]]}$\rightarrow$\texttt{[ i: i+ len( substring)]}}
\DefMacro{op-name-example-tok}{[\texttt{'[i', ':i', '+len', '(sub', 'string', ')]'}]$\rightarrow$[\texttt{'[', '\textvisiblespace i', ':', '\textvisiblespace i', '+', '\textvisiblespace len', '(', 'Ġsubstring', ')]'}]}
\DefMacro{op-name-example-lang}{\UseMacro{both}\UseMacro{op-name-example-tok}}
\DefMacro{op-name-example-colorful}{%
\TokenOld{:i}\TokenOld{+len}\TokenOld{(sub}\TokenOld{string}%
$\rightarrow$%
\TokenNew{:}\TokenNew{\textvisiblespace i}\TokenNew{+}\TokenNew{\textvisiblespace len}\TokenNew{(}\TokenNew{\textvisiblespace substring}%
}

\DefMacro{op-all-example-tok}{[\texttt{'(l', ':', '\textvisiblespace list', '):\textbackslash n'}]$\rightarrow$[\texttt{(', '\textvisiblespace l', ':', '\textvisiblespace list', ')', '\textvisiblespace :\textbackslash n'}]}
\DefMacro{op-all-example-lang}{\UseMacro{both}\UseMacro{op-all-example-tok}}
\DefMacro{op-all-example-colorful}{%
\TokenOld{(l}\TokenCtx{:}\TokenCtx{\textvisiblespace list}\TokenOld{):\textbackslash n}%
$\rightarrow$%
\TokenNew{(}\TokenNew{\textvisiblespace l}\TokenCtx{:}\TokenCtx{\textvisiblespace list}\TokenNew{)}\TokenNew{\textvisiblespace :\textbackslash n}%
}

\DefMacro{only-java}{$\times$}
\DefMacro{only-python}{$\square$}
\DefMacro{both}{$\boxtimes$}

\newcommand{\NumBenchmarks}{eight\xspace}
\newcommand{\NumModels}{nine\xspace}
\newcommand{\NumNamingRules}{six\xspace}
\newcommand{\NumSpacingRules}{eighteen\xspace}

\title{\Title}

\author{
Yinxi Li, Yuntian Deng, Pengyu Nie \\
  University of Waterloo \\
  \texttt{\{yinxi.li, yuntian, pynie\}@uwaterloo.ca}
}

\usepackage[nameinlink]{cleveref}


\begin{document}
\maketitle

\begin{abstract}

Large language models (LLMs) for code rely on \subword tokenizers, such as byte-pair encoding (BPE), learned from mixed natural language text and programming language code but driven by statistics rather than grammar.
As a result, semantically identical code snippets can be tokenized differently depending on superficial factors such as whitespace or identifier naming.
To measure the impact of this misalignment, we introduce \Tool, a framework that applies semantic-preserving rewrite rules to create code variants differing only in tokenization.
Across \NumModels code LLMs, including large ones with over 30B parameters, even minor formatting changes can cause substantial shifts in model behavior.
Layer-wise analysis shows that the issue originates in early embeddings, where \subword segmentation fails to capture grammar token boundaries.
Our findings identify misaligned tokenization as a hidden obstacle to reliable code understanding and generation, highlighting the need for grammar-aware tokenization for future code LLMs.

\end{abstract}


\section{Introduction}
\label{sec:intro}


\begin{figure}[!t]
\begin{center}
\begin{minipage}{\linewidth}
\includegraphics[width=.9\linewidth]{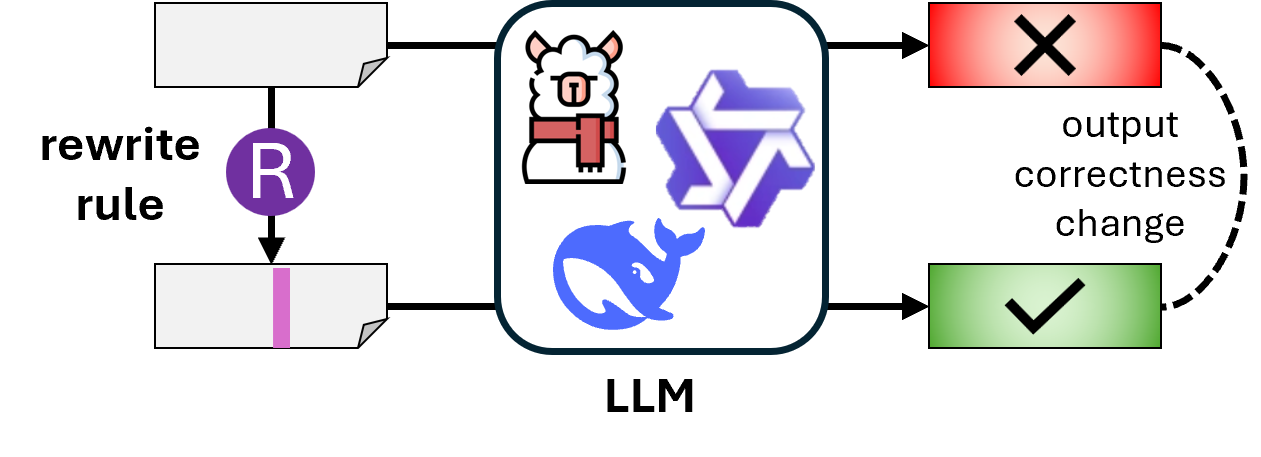}
\vspace{-3pt}
\subcaption{Workflow of \Tool, our framework for quantifying LLM sensitivity to semantic-preserving code \rewriterules.}
\label{fig:motivation:workflow}
\end{minipage}\\ \vspace{3pt}
\begin{minipage}{\linewidth}
\includegraphics[width=\linewidth]{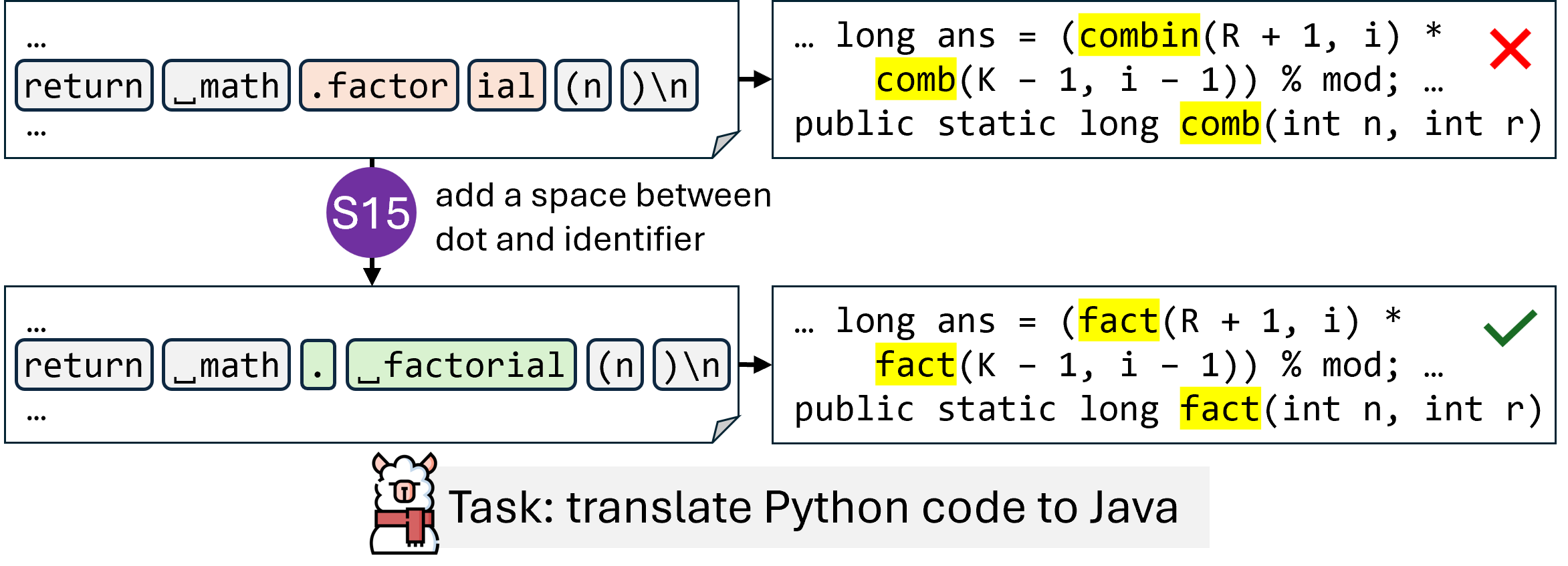}
\subcaption{Example of tokenization misalignment.
Adding a space between dot (``\texttt{.}'') and ``\texttt{factorial}'' causes a significant change in token sequences, from [``\texttt{.factor}'', ``\texttt{ial}''] to [``\texttt{.}'', ``\texttt{\textvisiblespace factorial}''].
Consequently, the LLM's code translation prediction shifts from incorrect (naming the factorial function as ``\texttt{comb}'' and later referring to it as ``\texttt{combin}'') to correct.
}
\label{fig:motivation:example1}
\end{minipage}

%
\caption{\Tool workflow and example.}
\label{fig:motivation}
\end{center}
\end{figure}

Large language models (LLMs) have become powerful tools for programming tasks~\citep{ChenETAL21Evaluating, nye2021workscratchpadsintermediatecomputation, yang2024sweagent, guo2024deepseekcoderlargelanguagemodel, meta2025cwm}.
Before any modeling occurs, code is first tokenized into discrete units using a \pretrained \subword tokenizer such as byte-pair encoding (BPE)~\citep{SennrichETAL16BPE}.
However, the tokens that LLMs see, which are based on \subword frequencies, are often very different from the tokens defined by programming language (PL) grammar.
Whereas PLs have clear syntactic boundaries (e.g., keywords, identifiers, operators), \subword tokenizers merge character sequences statistically, sometimes splitting identifiers at arbitrary points or combining unrelated symbols into a single token.
This misalignment between \subwords and syntax means that LLMs do not always process code in the units that programmers or compilers would expect.

As an example, the presence of a space before an identifier can lead to completely different token sequences, and thus different predictions, despite identical program semantics (\Cref{fig:motivation}).
While such differences may appear superficial, they raise a deeper concern about how robustly code LLMs represent grammar and meaning.
If tokenization determines how code is segmented and embedded, even small discrepancies could propagate through the model and alter its predictions.
This motivates the central question of our study: 


\begin{tcolorbox}[notitle,boxrule=0pt,colback=gray!30,colframe=gray!20]
\emph{Does the misalignment between \subword tokenization and PL grammar limit LLMs' ability to understand and generate code?}
\end{tcolorbox}

To study this question, we introduce \Tool, a framework that applies semantic-preserving \rewriterules, such as changing whitespace or identifier casing style, to create pairs of programs that are semantically equivalent but tokenized differently.
We evaluate \NumModels code LLMs across three representative programming tasks---bug fixing, code summarization, and code translation---and measure whether model outputs remain functionally equivalent when tokenization changes.

Our experiments show that even minor tokenization variations can substantially impact model behavior.
For example, the most performant LLM in our experiment, \UseMacro{qwen-large}, changes its prediction 6.09\% of the times when the input tokenization changes (and up to 60\% under a single \rewriterule).
Layer-wise analysis further indicates that the effect originates in early layers, where subword segmentation fails to align with grammatical token boundaries.
Together, these findings suggest that tokenizer design remains a critical yet under-explored factor in developing robust and grammar-aware code LLMs.

\noindent
The main contributions of this work include:

\begin{itemize}[topsep=3pt,itemsep=1ex,partopsep=0ex,parsep=0ex,leftmargin=*]
\item We identify and formalize the misaligned tokenization problem in code LLMs.
\item We introduce \Tool, a framework for quantifying model sensitivity to semantic-preserving code rewrites that alter tokenization.
\item We conduct a large-scale empirical study showing that misaligned tokenization affects all evaluated models and persists with scaling.
\item We open-source our framework and data to facilitate future research on grammar-aware and domain-adaptive tokenization.
\end{itemize}

\noindent
Our code and data are available at:\\
\url{https://github.com/uw-swag/tokdrift}

\section{Background}
\label{sec:background}

\subsection{LLM Tokenization}
\label{sec:background:llm-tokenization}

Tokenization is the first step in processing input for LLMs, converting raw text into a sequence of discrete tokens. Each token corresponds to a model time step and has a dedicated embedding. 
%
Modern LLMs use learned tokenization strategies that eliminate the out-of-vocabulary problem by starting from minimal units, such as characters or bytes, and learning how to merge them into longer fragments based on frequency in a large corpus. 
Popular approaches like BPE~\citep{SennrichETAL16BPE} and WordPiece~\citep{37842,devlin-etal-2019-bert} follow this general principle, differing mainly in their merge heuristics. Often, pre-tokenization steps like splitting at whitespace are applied before learning to prevent tokens from spanning across word boundaries.

\begin{figure}[t]
\centering
\includegraphics[width=.82\linewidth]{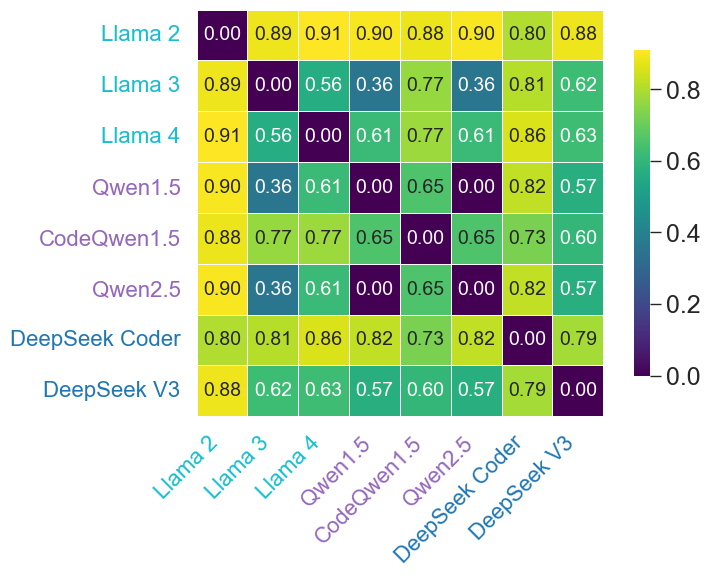}
\vspace{-10pt}
\caption{Heatmap of (code) LLMs' vocabulary distances~\cite{AmbaHombaiahETAL21Dynamic}.}
\label{fig:vocb-heatmap-selected}
\vspace{-10pt}
\end{figure}

The tokenizers used by different LLMs can vary significantly due to differences in pre-tokenization rules, token learning algorithms, and pretraining corpora. As shown in \Cref{fig:vocb-heatmap-selected}, even models from the same family often share less than half of their vocabulary, such as Llama 3 vs. Llama 4. 
The main exception occurs when model developers intentionally reuse the same tokenizer across variants, such as Qwen2.5 and Qwen2.5-Coder, which share an identical vocabulary and tokenizer configuration.


\begin{table*}[!t]
\begin{center}
\caption{Benchmarks in our experiments. We manually examine the benchmarks to follow the naming conventions, and to fix/exclude invalid tests and samples, see details in \Cref{sec:appendix-methodology:benchmarks-normalization}.}
\label{tab:benchmarks}
\vspace{-8pt}
\resizebox{.9\textwidth}{!}{
\begin{tabular}{@{}l|c|c|ccc@{}}
\toprule
\textbf{Benchmark} & \textbf{Source} & \textbf{Task} & \textbf{Input PL} & \textbf{Output PL} & \textbf{\# Samples} \\
\midrule
\UseMacro{humaneval-fix-py} & \multirow{4}{*}{\makecell[c]{HumanEvalPack\\ \cite{MuennighoffETAL23OctoPack}}} & \multirow{2}{*}{bug fixing} & Python & Python & 164 \\
\UseMacro{humaneval-fix-java} &  &  & Java & Java & 164 \\ \cline{1-1} \cline{3-6}
\UseMacro{humaneval-explain-py} &  & \multirow{2}{*}{code summarization} & Python & Python & 164 \\
\UseMacro{humaneval-explain-java} &  &  & Java & Java & 164 \\ \cline{1-6}
\UseMacro{avatar-py2java} & \multirow{2}{*}{\makecell[c]{Avatar\\ \cite{AhmadETAL23Avatar, pan2024lost}}} & \multirow{4}{*}{code translation} & Python & Java & 244 \\
\UseMacro{avatar-java2py} & & & Java & Python & 246 \\ \cline{1-2} \cline{4-6}
\UseMacro{codenet-py2java} & \multirow{2}{*}{\makecell[c]{CodeNet\\ \cite{PuriETAL21CodeNet, pan2024lost}}} & & Python & Java & 200 \\
\UseMacro{codenet-java2py} & & & Java & Python & 200 \\
\bottomrule
\end{tabular}

}
\end{center}
\vspace{-10pt}
\end{table*}

\subsection{PL Tokenization}
\label{sec:background:pl-tokenization}



Tokenization in PLs, often called \emph{lexing}, is the first step of code parsing: it transforms a stream of characters into a sequence of tokens according to a PL's grammar. These tokens are then passed to a parser, which constructs an abstract syntax tree (AST) to represent the program's structure.

While exact rules vary by language, most PLs share a common set of token types, including:
\emph{identifiers} (e.g., variable or function names),
\emph{operators} (e.g., \texttt{+}, \texttt{*}),
\emph{keywords} (e.g., \texttt{if}, \texttt{return}),
\emph{literals} (e.g., numeric or string constants), and
\emph{whitespace}, which is typically used to separate tokens but is otherwise ignored.

Unlike LLM tokenization, PL tokenization in compilers and interpreters is deterministic. 
For example, the snippet \texttt{x+1} is always tokenized into three tokens: an identifier (\texttt{x}), an operator (\texttt{+}), and a literal (\texttt{1}). 
Formatting changes, such as adding spaces, do not affect the token sequence as long as the code remains syntactically valid.

This behavior misaligns with LLM tokenizers: while PL tokenizers produce stable, grammar-aware units, LLM tokenizers frequently break code structure, resulting in inconsistent or fragmented representations of semantically identical programs. 
In this work, we refer to grammar-aware tokens as \emph{PL tokens}, and contrast them with the \emph{LLM tokens} produced by learned \subword tokenizers.
\section{\Tool Framework}
\label{sec:methodology}

\Cref{fig:motivation:workflow} illustrates the overall workflow of \Tool, our framework for quantifying model sensitivity to semantic-preserving code rewrites that alter tokenization.
In a nutshell, \Tool systematically compares the LLM outputs given the baseline input tokens and variant input tokens (after applying \rewriterules) through a large set of experiments.
Each experiment is performed on a specific benchmark, and tests the sensitivity of a given LLM against a specific \rewriterule.


\subsection{Benchmarks}
\label{sec:methodology:benchmarks}

We searched for recent popular coding LLM benchmarks where:
(1)~the input includes a code snippet, since \rewriterules cannot be applied on natural language;
(2)~the output is evaluated with an automated functional correctness metric.
We focused on two popular PLs, Java and Python.

Based on these criteria, we selected \NumBenchmarks benchmarks covering three tasks, listed in \Cref{tab:benchmarks}.
Bug fixing~\cite{tufano2019learning} transforms a buggy code snippet into a correct one.
Code summarization~\cite{hu2018deep,panthaplackel2020learning} aims at summarizing a code snippet into natural language description;
following HumanEvalPack's setup~\cite{MuennighoffETAL23OctoPack}, the description is fed back to LLM to generate code for measuring correctness.
Code translation~\cite{AhmadETAL23Avatar, PuriETAL21CodeNet} is the task of translating a code snippet from one PL to another.
All benchmarks use tests to evaluate the correctness of outputs.

\subsection{Models}
\label{sec:methodology:models}

\begin{table}[t!]
\begin{small}
\begin{center}
\caption{Models used in our experiments.}
\label{tab:models}
\vspace{-5pt}
\begin{tabular}{@{}lccc@{}}
\toprule
\textbf{Series} & \textbf{S} & \textbf{M} & \textbf{L} \\
\midrule
\llama & 3B & 8B & 70B \\
\qwen & 1.5B & 7B & 32B \\
\dscoder & 1.3B & 6.7B & 33B \\
\bottomrule
\end{tabular}

\end{center}
\end{small}
\vspace{-10pt}
\end{table}

\begin{table*}[t!]
\begin{center}
\caption{Rewrite rules supported by \Tool, inspired by naming conventions (starting with N) and spacing conventions (starting with S).
Each \rewriterule may apply to Java (marked by J), Python (marked by P), or both.}
\label{tab:rewrite-rules}
\vspace{-5pt}
\resizebox{\textwidth}{!}{

\begin{tabular}{@{}l@{\hspace{3pt}}c@{}cll@{}}
\toprule
\textbf{No.} & \textbf{PL} & \textbf{Rewrite Rule} & \textbf{Description} & \textbf{Example} \\
\midrule
\UseMacro{camel-case-snake-case-no} & \LangJava & \UseMacro{N1-short} 
& \multirow{6}{*}{\makecell[l]{Convert identifiers from the most common\\ casing style in the input PL to alternative ones}} & \UseMacro{camel-snake-example-colorful} \\
\UseMacro{camel-case-pascal-case-no} & \LangJava & \UseMacro{N2-short} 
&& \UseMacro{camel-pascal-example-colorful} \\
\UseMacro{camel-case-screaming-snake-case-no} & \LangJava & \UseMacro{N3-short} 
&& \UseMacro{camel-screaming-snake-example-colorful} \\
\UseMacro{snake-case-camel-case-no} & \LangPython & \UseMacro{N4-short} 
&& \UseMacro{snake-camel-example-colorful} \\
\UseMacro{snake-case-pascal-case-no} & \LangPython & \UseMacro{N5-short} 
&& \UseMacro{snake-pascal-example-colorful} \\
\UseMacro{snake-case-screaming-snake-case-no} & \LangPython & \UseMacro{N6-short} 
&& \UseMacro{snake-screaming-snake-example-colorful} \\
\midrule
\UseMacro{op-dash-no} & \LangPython & \UseMacro{S1-short} 
& {\footnotesize Add space between operator and minus sign} & \UseMacro{op-dash-example-colorful} \\
\UseMacro{op-lsquarebracket-no} & \LangPython & \UseMacro{S2-short} 
& {\footnotesize Add space between operator and left square bracket} & \UseMacro{op-lsquarebracket-example-colorful} \\
\UseMacro{rparentheses-period-no} & \LangJava & \UseMacro{S3-short} 
& {\footnotesize Add space between right parentheses and period} & \UseMacro{rparentheses-period-example-colorful} \\
\UseMacro{rsquarebracket-rparentheses-no} & \LangPython & \UseMacro{S4-short} 
& {\footnotesize Add space between right square bracket and right parentheses} & \UseMacro{rsquarebracket-rparentheses-example-colorful} \\
\UseMacro{op-rsquarebracket-no} & \LangPython & \UseMacro{S5-short} 
& {\footnotesize Add space between operator and right square bracket} & \UseMacro{op-rsquarebracket-example-colorful} \\
\UseMacro{op-lparentheses-no} & \LangJava & \UseMacro{S6-short} 
& {\footnotesize Add space between operator and left parentheses} & \UseMacro{op-lparentheses-example-colorful} \\
\UseMacro{lsquarebracket-name-no} & \LangPython & \UseMacro{S7-short} 
& {\footnotesize Add space between left square bracket and identifier} & \UseMacro{lsquarebracket-name-example-colorful} \\
\UseMacro{double-plus-rparentheses-no} & \LangJava & \UseMacro{S8-short} 
& {\footnotesize Add space between increment operator and right parentheses} & \UseMacro{double-plus-rparentheses-example-colorful} \\
\UseMacro{period-asterisk-no} & \LangJava & \UseMacro{S9-short} 
& {\footnotesize Add space between period and asterisk} & \UseMacro{period-asterisk-example-colorful} \\
\UseMacro{rparentheses-colon-no} & \LangPython & \UseMacro{S10-short} 
& {\footnotesize Add space between right parentheses and colon} & \UseMacro{rparentheses-colon-example-colorful} \\
\UseMacro{rparentheses-semicolon-no} & \LangJava & \UseMacro{S11-short} 
& {\footnotesize Add space between right parentheses and semicolon} & \UseMacro{rparentheses-semicolon-example-colorful} \\
\UseMacro{op-semicolon-no} & \LangJava & \UseMacro{S12-short} 
& {\footnotesize Add space between operator and semicolon} & \UseMacro{op-semicolon-example-colorful} \\
\UseMacro{rparentheses-rparentheses-no} & \LangBoth & \UseMacro{S13-short} 
& {\footnotesize Add space between two right parentheses} & \UseMacro{rparentheses-rparentheses-example-colorful} \\
\UseMacro{lparentheses-rparentheses-no} & \LangBoth & \UseMacro{S14-short} 
& {\footnotesize Add space between two left parentheses} & \UseMacro{lparentheses-rparentheses-example-colorful} \\
\UseMacro{period-name-no} & \LangBoth & \UseMacro{S15-short} 
& {\footnotesize Add space between period and identifier} & \UseMacro{period-name-example-colorful} \\
\UseMacro{lparentheses-name-no} & \LangBoth & \UseMacro{S16-short} 
& {\footnotesize Add space between left parentheses and identifier} & \UseMacro{lparentheses-name-example-colorful} \\
\UseMacro{op-name-no} & \LangBoth & \UseMacro{S17-short} 
& {\footnotesize Add space between operator and identifier} & \UseMacro{op-name-example-colorful} \\
\UseMacro{op-all-no} & \LangBoth & \UseMacro{S18-short} 
& {\footnotesize Add space between operator and identifier/operator} & \UseMacro{op-all-example-colorful} \\
\bottomrule
\end{tabular}
                    
}
\end{center}
\vspace{-10pt}
\end{table*}

\Cref{tab:models} lists the models used in \Tool.
We selected three series of popular open-source LLMs (using the coding-specific variants if available), namely \llama, \qwen, and \dscoder.
To cover the model size spectrum, we used small ($\sim$1B parameters), medium ($\sim$7B), and large ($>$30B) variants in each series.
All models are instruction-tuned.
We perform greedy decoding to generate deterministic outputs (see experimental environment details in \Cref{sec:appendix-methodology:experimental-environment}).

\subsection{Rewrite Rules}
\label{sec:methodology:rewrite}

\Cref{tab:rewrite-rules} lists the \rewriterules used in \Tool.
Each \rewriterule converts all occurrences of the left-hand side substring to the right-hand side substring.
According to the grammars of the two PLs we experiment on (and generally for most modern PLs), these rewrite rules are semantically-preserving by design.
We apply one \rewriterule at a time to investigate their impact in isolation.

The \NumNamingRules \rewriterules starting with ``N'' are inspired by naming conventions.
Identifiers usually follow one of the four casing styles:
\aCamelCase (for variables/functions in Java), \aPascalCase (for classes in Java/Python), \aSnakeCase (for variables/functions in Python), and \aScreamingSnakeCase (for constants in Java/Python).
Since variables/functions are most common among identifiers, we design \rewriterules to alter their casing style.
Specifically, \UseMacro{camel-case-snake-case-no}, \UseMacro{camel-case-pascal-case-no}, \UseMacro{camel-case-screaming-snake-case-no} convert \aCamelCase identifiers in Java to the other three casing styles,
while \UseMacro{snake-case-camel-case-no}, \UseMacro{snake-case-pascal-case-no}, \UseMacro{snake-case-screaming-snake-case-no} convert \aSnakeCase identifiers in Python. 
These \rewriterules challenge LLMs' robustness to different naming styles.

The \NumSpacingRules \rewriterules starting with ``S'' are inspired by spacing conventions.
Whitespace around most operators usually carries no semantic meaning and is optional.
Thus, the spacing-related \rewriterules identifies two consecutive tokens (one of them is an operator) and inserts a space in between.
Specifically, we look for combinations where one of them is a specific operator or any kind of operator (represented by \aOP), and the other one is another specific operator or an identifier (represented by \aID).
Exploring all combinations would be infeasible, thus we select the top-10 frequently appearing combinations in the benchmarks for each PL.
In addition, we add \UseMacro{op-name-no} and \UseMacro{op-all-no} as ``wildcard'' rules to cover all cases where an \aOP is followed by an \aID or \aID/\aOP for both PLs.
These \rewriterules challenge LLM and its tokenizer's robustness to different formatting styles.
Notably, in most LLMs with a pre-tokenization step of splitting before whitespace, these \rewriterules will lead to more \llmtokens.


\begin{table*}[t!]
\begin{center}
\caption{\Acc and \accdelta (in parenthesis) of each model on each \rewriterule.}
\vspace{-5pt}
\label{tab:accuracy}
\resizebox{\linewidth}{!}{%

\resizebox{\textwidth}{!}{%
\begin{tabular}{l|cccccccccc}
\toprule
\UseMacro{TH-variant} & \UseMacro{llama-small-short} & \UseMacro{llama-medium-short} & \UseMacro{llama-large-short} & \UseMacro{qwen-small-short} & \UseMacro{qwen-medium-short} & \UseMacro{qwen-large-short} & \UseMacro{deepseek-coder-1.3b-short} & \UseMacro{deepseek-coder-6.7b-short} & \UseMacro{deepseek-coder-33b-short} & \UseMacro{TH-accuracy-average} \\
\midrule
\multicolumn{11}{c}{\UseMacro{TH-java-results}} \\
\midrule
\rowcolor[HTML]{C0C0C0} \UseMacro{baseline} & \UseMacro{llama-small_java_baseline}\phantom{ \tiny{(+0.00)}} & \UseMacro{llama-medium_java_baseline}\phantom{ \tiny{(+0.00)}} & \UseMacro{llama-large_java_baseline}\phantom{ \tiny{(+0.00)}} & \UseMacro{qwen-small_java_baseline}\phantom{ \tiny{(+0.00)}} & \UseMacro{qwen-medium_java_baseline}\phantom{ \tiny{(+0.00)}} & \UseMacro{qwen-large_java_baseline}\phantom{ \tiny{(+0.00)}} & \UseMacro{dscoder-small_java_baseline}\phantom{ \tiny{(+0.00)}} & \UseMacro{dscoder-medium_java_baseline}\phantom{ \tiny{(+0.00)}} & \UseMacro{dscoder-large_java_baseline}\phantom{ \tiny{(+0.00)}} & \UseMacro{average_java_baseline}\phantom{ \tiny{(+0.00)}} \\
\UseMacro{camel-case-snake-case-no} & \textbf{\UseMacro{llama-small_java_camel-case-snake-case}} {\tiny (\cellcolor{green!20}\UseMacro{llama-small_java_camel-case-snake-case_delta})} & \UseMacro{llama-medium_java_camel-case-snake-case} {\tiny (\cellcolor{green!20}\UseMacro{llama-medium_java_camel-case-snake-case_delta})} & \UseMacro{llama-large_java_camel-case-snake-case} {\tiny (\cellcolor{green!20}\UseMacro{llama-large_java_camel-case-snake-case_delta})} & \UseMacro{qwen-small_java_camel-case-snake-case} {\tiny (\cellcolor{green!20}\UseMacro{qwen-small_java_camel-case-snake-case_delta})} & \UseMacro{qwen-medium_java_camel-case-snake-case} {\tiny (\cellcolor{green!20}\UseMacro{qwen-medium_java_camel-case-snake-case_delta})} & \UseMacro{qwen-large_java_camel-case-snake-case} {\tiny (\cellcolor{red!20}\UseMacro{qwen-large_java_camel-case-snake-case_delta})} & \UseMacro{dscoder-small_java_camel-case-snake-case} {\tiny (\cellcolor{red!20}\UseMacro{dscoder-small_java_camel-case-snake-case_delta})} & \UseMacro{dscoder-medium_java_camel-case-snake-case} {\tiny (\cellcolor{red!20}\UseMacro{dscoder-medium_java_camel-case-snake-case_delta})} & \UseMacro{dscoder-large_java_camel-case-snake-case} {\tiny (\cellcolor{red!20}\UseMacro{dscoder-large_java_camel-case-snake-case_delta})} & \UseMacro{average_java_camel-case-snake-case} {\tiny (\cellcolor{green!20}\UseMacro{average_java_camel-case-snake-case_delta})} \\
\UseMacro{camel-case-pascal-case-no} & \UseMacro{llama-small_java_camel-case-pascal-case} {\tiny (\cellcolor{green!20}\UseMacro{llama-small_java_camel-case-pascal-case_delta})} & \UseMacro{llama-medium_java_camel-case-pascal-case} {\tiny (\cellcolor{green!20}\UseMacro{llama-medium_java_camel-case-pascal-case_delta})} & \UseMacro{llama-large_java_camel-case-pascal-case} {\tiny (\cellcolor{red!20}\UseMacro{llama-large_java_camel-case-pascal-case_delta})} & \UseMacro{qwen-small_java_camel-case-pascal-case} {\tiny (\cellcolor{green!20}\UseMacro{qwen-small_java_camel-case-pascal-case_delta})} & \UseMacro{qwen-medium_java_camel-case-pascal-case} {\tiny (\cellcolor{green!20}\UseMacro{qwen-medium_java_camel-case-pascal-case_delta})} & \UseMacro{qwen-large_java_camel-case-pascal-case} {\tiny (\UseMacro{qwen-large_java_camel-case-pascal-case_delta})} & \textbf{\UseMacro{dscoder-small_java_camel-case-pascal-case}} {\tiny (\cellcolor{green!20}\UseMacro{dscoder-small_java_camel-case-pascal-case_delta})} & \UseMacro{dscoder-medium_java_camel-case-pascal-case} {\tiny (\cellcolor{green!20}\UseMacro{dscoder-medium_java_camel-case-pascal-case_delta})} & \UseMacro{dscoder-large_java_camel-case-pascal-case} {\tiny (\UseMacro{dscoder-large_java_camel-case-pascal-case_delta})} & \textbf{\UseMacro{average_java_camel-case-pascal-case}} {\tiny (\cellcolor{green!20}\UseMacro{average_java_camel-case-pascal-case_delta})} \\
\UseMacro{camel-case-screaming-snake-case-no} & \UseMacro{llama-small_java_camel-case-screaming-snake-case} {\tiny (\cellcolor{green!20}\UseMacro{llama-small_java_camel-case-screaming-snake-case_delta})} & \UseMacro{llama-medium_java_camel-case-screaming-snake-case} {\tiny (\cellcolor{green!20}\UseMacro{llama-medium_java_camel-case-screaming-snake-case_delta})} & \UseMacro{llama-large_java_camel-case-screaming-snake-case} {\tiny (\cellcolor{red!20}\UseMacro{llama-large_java_camel-case-screaming-snake-case_delta})} & \UseMacro{qwen-small_java_camel-case-screaming-snake-case} {\tiny (\cellcolor{green!20}\UseMacro{qwen-small_java_camel-case-screaming-snake-case_delta})} & \UseMacro{qwen-medium_java_camel-case-screaming-snake-case} {\tiny (\cellcolor{green!20}\UseMacro{qwen-medium_java_camel-case-screaming-snake-case_delta})} & \UseMacro{qwen-large_java_camel-case-screaming-snake-case} {\tiny (\cellcolor{red!20}\UseMacro{qwen-large_java_camel-case-screaming-snake-case_delta})} & \UseMacro{dscoder-small_java_camel-case-screaming-snake-case} {\tiny (\cellcolor{red!20}\UseMacro{dscoder-small_java_camel-case-screaming-snake-case_delta})} & \UseMacro{dscoder-medium_java_camel-case-screaming-snake-case} {\tiny (\cellcolor{red!20}\UseMacro{dscoder-medium_java_camel-case-screaming-snake-case_delta})} & \UseMacro{dscoder-large_java_camel-case-screaming-snake-case} {\tiny (\cellcolor{red!20}\UseMacro{dscoder-large_java_camel-case-screaming-snake-case_delta})} & \UseMacro{average_java_camel-case-screaming-snake-case} {\tiny (\cellcolor{red!20}\UseMacro{average_java_camel-case-screaming-snake-case_delta})} \\
\UseMacro{rparentheses-period-no} & \UseMacro{llama-small_java_rparentheses-period} {\tiny (\cellcolor{red!20}\UseMacro{llama-small_java_rparentheses-period_delta})} & \UseMacro{llama-medium_java_rparentheses-period} {\tiny (\cellcolor{red!20}\UseMacro{llama-medium_java_rparentheses-period_delta})} & \UseMacro{llama-large_java_rparentheses-period} {\tiny (\cellcolor{red!20}\UseMacro{llama-large_java_rparentheses-period_delta})} & \UseMacro{qwen-small_java_rparentheses-period} {\tiny (\cellcolor{green!20}\UseMacro{qwen-small_java_rparentheses-period_delta})} & \UseMacro{qwen-medium_java_rparentheses-period} {\tiny (\cellcolor{red!20}\UseMacro{qwen-medium_java_rparentheses-period_delta})} & \UseMacro{qwen-large_java_rparentheses-period} {\tiny (\UseMacro{qwen-large_java_rparentheses-period_delta})} & \UseMacro{dscoder-small_java_rparentheses-period} {\tiny (\cellcolor{red!20}\UseMacro{dscoder-small_java_rparentheses-period_delta})} & \textbf{\UseMacro{dscoder-medium_java_rparentheses-period}} {\tiny (\cellcolor{green!20}\UseMacro{dscoder-medium_java_rparentheses-period_delta})} & \UseMacro{dscoder-large_java_rparentheses-period} {\tiny (\cellcolor{green!20}\UseMacro{dscoder-large_java_rparentheses-period_delta})} & \UseMacro{average_java_rparentheses-period} {\tiny (\cellcolor{red!20}\UseMacro{average_java_rparentheses-period_delta})} \\
\UseMacro{op-lparentheses-no} & \UseMacro{llama-small_java_op-lparentheses} {\tiny (\cellcolor{red!20}\UseMacro{llama-small_java_op-lparentheses_delta})} & \UseMacro{llama-medium_java_op-lparentheses} {\tiny (\cellcolor{red!20}\UseMacro{llama-medium_java_op-lparentheses_delta})} & \UseMacro{llama-large_java_op-lparentheses} {\tiny (\cellcolor{green!20}\UseMacro{llama-large_java_op-lparentheses_delta})} & \UseMacro{qwen-small_java_op-lparentheses} {\tiny (\cellcolor{red!20}\UseMacro{qwen-small_java_op-lparentheses_delta})} & \UseMacro{qwen-medium_java_op-lparentheses} {\tiny (\cellcolor{green!20}\UseMacro{qwen-medium_java_op-lparentheses_delta})} & \UseMacro{qwen-large_java_op-lparentheses} {\tiny (\cellcolor{red!20}\UseMacro{qwen-large_java_op-lparentheses_delta})} & \UseMacro{dscoder-small_java_op-lparentheses} {\tiny (\cellcolor{red!20}\UseMacro{dscoder-small_java_op-lparentheses_delta})} & \UseMacro{dscoder-medium_java_op-lparentheses} {\tiny (\cellcolor{green!20}\UseMacro{dscoder-medium_java_op-lparentheses_delta})} & \UseMacro{dscoder-large_java_op-lparentheses} {\tiny (\cellcolor{green!20}\UseMacro{dscoder-large_java_op-lparentheses_delta})} & \UseMacro{average_java_op-lparentheses} {\tiny (\cellcolor{red!20}\UseMacro{average_java_op-lparentheses_delta})} \\
\UseMacro{double-plus-rparentheses-no} & \UseMacro{llama-small_java_double-plus-rparentheses} {\tiny (\cellcolor{red!20}\UseMacro{llama-small_java_double-plus-rparentheses_delta})} & \UseMacro{llama-medium_java_double-plus-rparentheses} {\tiny (\cellcolor{green!20}\UseMacro{llama-medium_java_double-plus-rparentheses_delta})} & \UseMacro{llama-large_java_double-plus-rparentheses} {\tiny (\UseMacro{llama-large_java_double-plus-rparentheses_delta})} & \UseMacro{qwen-small_java_double-plus-rparentheses} {\tiny (\cellcolor{green!20}\UseMacro{qwen-small_java_double-plus-rparentheses_delta})} & \UseMacro{qwen-medium_java_double-plus-rparentheses} {\tiny (\cellcolor{red!20}\UseMacro{qwen-medium_java_double-plus-rparentheses_delta})} & \UseMacro{qwen-large_java_double-plus-rparentheses} {\tiny (\cellcolor{green!20}\UseMacro{qwen-large_java_double-plus-rparentheses_delta})} & \UseMacro{dscoder-small_java_double-plus-rparentheses} {\tiny (\cellcolor{green!20}\UseMacro{dscoder-small_java_double-plus-rparentheses_delta})} & \UseMacro{dscoder-medium_java_double-plus-rparentheses} {\tiny (\cellcolor{red!20}\UseMacro{dscoder-medium_java_double-plus-rparentheses_delta})} & \UseMacro{dscoder-large_java_double-plus-rparentheses} {\tiny (\cellcolor{green!20}\UseMacro{dscoder-large_java_double-plus-rparentheses_delta})} & \UseMacro{average_java_double-plus-rparentheses} {\tiny (\cellcolor{green!20}\UseMacro{average_java_double-plus-rparentheses_delta})} \\
\UseMacro{period-asterisk-no} & \UseMacro{llama-small_java_period-asterisk} {\tiny (\cellcolor{green!20}\UseMacro{llama-small_java_period-asterisk_delta})} & \UseMacro{llama-medium_java_period-asterisk} {\tiny (\cellcolor{red!20}\UseMacro{llama-medium_java_period-asterisk_delta})} & \textbf{\UseMacro{llama-large_java_period-asterisk}} {\tiny (\cellcolor{green!20}\UseMacro{llama-large_java_period-asterisk_delta})} & \UseMacro{qwen-small_java_period-asterisk} {\tiny (\cellcolor{red!20}\UseMacro{qwen-small_java_period-asterisk_delta})} & \textbf{\UseMacro{qwen-medium_java_period-asterisk}} {\tiny (\cellcolor{green!20}\UseMacro{qwen-medium_java_period-asterisk_delta})} & \UseMacro{qwen-large_java_period-asterisk} {\tiny (\cellcolor{red!20}\UseMacro{qwen-large_java_period-asterisk_delta})} & \UseMacro{dscoder-small_java_period-asterisk} {\tiny (\cellcolor{red!20}\UseMacro{dscoder-small_java_period-asterisk_delta})} & \UseMacro{dscoder-medium_java_period-asterisk} {\tiny (\cellcolor{red!20}\UseMacro{dscoder-medium_java_period-asterisk_delta})} & \UseMacro{dscoder-large_java_period-asterisk} {\tiny (\cellcolor{green!20}\UseMacro{dscoder-large_java_period-asterisk_delta})} & \UseMacro{average_java_period-asterisk} {\tiny (\cellcolor{red!20}\UseMacro{average_java_period-asterisk_delta})} \\
\UseMacro{rparentheses-semicolon-no} & \UseMacro{llama-small_java_rparentheses-semicolon} {\tiny (\cellcolor{green!20}\UseMacro{llama-small_java_rparentheses-semicolon_delta})} & \textbf{\UseMacro{llama-medium_java_rparentheses-semicolon}} {\tiny (\cellcolor{green!20}\UseMacro{llama-medium_java_rparentheses-semicolon_delta})} & \UseMacro{llama-large_java_rparentheses-semicolon} {\tiny (\cellcolor{red!20}\UseMacro{llama-large_java_rparentheses-semicolon_delta})} & \UseMacro{qwen-small_java_rparentheses-semicolon} {\tiny (\cellcolor{green!20}\UseMacro{qwen-small_java_rparentheses-semicolon_delta})} & \UseMacro{qwen-medium_java_rparentheses-semicolon} {\tiny (\cellcolor{red!20}\UseMacro{qwen-medium_java_rparentheses-semicolon_delta})} & \UseMacro{qwen-large_java_rparentheses-semicolon} {\tiny (\cellcolor{green!20}\UseMacro{qwen-large_java_rparentheses-semicolon_delta})} & \UseMacro{dscoder-small_java_rparentheses-semicolon} {\tiny (\cellcolor{red!20}\UseMacro{dscoder-small_java_rparentheses-semicolon_delta})} & \UseMacro{dscoder-medium_java_rparentheses-semicolon} {\tiny (\cellcolor{red!20}\UseMacro{dscoder-medium_java_rparentheses-semicolon_delta})} & \UseMacro{dscoder-large_java_rparentheses-semicolon} {\tiny (\cellcolor{red!20}\UseMacro{dscoder-large_java_rparentheses-semicolon_delta})} & \UseMacro{average_java_rparentheses-semicolon} {\tiny (\cellcolor{red!20}\UseMacro{average_java_rparentheses-semicolon_delta})} \\
\UseMacro{op-semicolon-no} & \UseMacro{llama-small_java_op-semicolon} {\tiny (\cellcolor{red!20}\UseMacro{llama-small_java_op-semicolon_delta})} & \UseMacro{llama-medium_java_op-semicolon} {\tiny (\cellcolor{red!20}\UseMacro{llama-medium_java_op-semicolon_delta})} & \UseMacro{llama-large_java_op-semicolon} {\tiny (\cellcolor{red!20}\UseMacro{llama-large_java_op-semicolon_delta})} & \UseMacro{qwen-small_java_op-semicolon} {\tiny (\cellcolor{green!20}\UseMacro{qwen-small_java_op-semicolon_delta})} & \UseMacro{qwen-medium_java_op-semicolon} {\tiny (\cellcolor{red!20}\UseMacro{qwen-medium_java_op-semicolon_delta})} & \underline{\UseMacro{qwen-large_java_op-semicolon}} {\tiny (\cellcolor{red!20}\UseMacro{qwen-large_java_op-semicolon_delta})} & \UseMacro{dscoder-small_java_op-semicolon} {\tiny (\cellcolor{green!20}\UseMacro{dscoder-small_java_op-semicolon_delta})} & \UseMacro{dscoder-medium_java_op-semicolon} {\tiny (\cellcolor{red!20}\UseMacro{dscoder-medium_java_op-semicolon_delta})} & \UseMacro{dscoder-large_java_op-semicolon} {\tiny (\cellcolor{green!20}\UseMacro{dscoder-large_java_op-semicolon_delta})} & \UseMacro{average_java_op-semicolon} {\tiny (\cellcolor{red!20}\UseMacro{average_java_op-semicolon_delta})} \\
\UseMacro{rparentheses-rparentheses-no} & \UseMacro{llama-small_java_rparentheses-rparentheses} {\tiny (\cellcolor{green!20}\UseMacro{llama-small_java_rparentheses-rparentheses_delta})} & \UseMacro{llama-medium_java_rparentheses-rparentheses} {\tiny (\cellcolor{red!20}\UseMacro{llama-medium_java_rparentheses-rparentheses_delta})} & \UseMacro{llama-large_java_rparentheses-rparentheses} {\tiny (\cellcolor{red!20}\UseMacro{llama-large_java_rparentheses-rparentheses_delta})} & \UseMacro{qwen-small_java_rparentheses-rparentheses} {\tiny (\cellcolor{red!20}\UseMacro{qwen-small_java_rparentheses-rparentheses_delta})} & \UseMacro{qwen-medium_java_rparentheses-rparentheses} {\tiny (\UseMacro{qwen-medium_java_rparentheses-rparentheses_delta})} & \UseMacro{qwen-large_java_rparentheses-rparentheses} {\tiny (\cellcolor{red!20}\UseMacro{qwen-large_java_rparentheses-rparentheses_delta})} & \UseMacro{dscoder-small_java_rparentheses-rparentheses} {\tiny (\cellcolor{red!20}\UseMacro{dscoder-small_java_rparentheses-rparentheses_delta})} & \UseMacro{dscoder-medium_java_rparentheses-rparentheses} {\tiny (\cellcolor{green!20}\UseMacro{dscoder-medium_java_rparentheses-rparentheses_delta})} & \UseMacro{dscoder-large_java_rparentheses-rparentheses} {\tiny (\cellcolor{red!20}\UseMacro{dscoder-large_java_rparentheses-rparentheses_delta})} & \UseMacro{average_java_rparentheses-rparentheses} {\tiny (\cellcolor{red!20}\UseMacro{average_java_rparentheses-rparentheses_delta})} \\
\UseMacro{lparentheses-rparentheses-no} & \UseMacro{llama-small_java_lparentheses-rparentheses} {\tiny (\cellcolor{red!20}\UseMacro{llama-small_java_lparentheses-rparentheses_delta})} & \UseMacro{llama-medium_java_lparentheses-rparentheses} {\tiny (\cellcolor{red!20}\UseMacro{llama-medium_java_lparentheses-rparentheses_delta})} & \underline{\UseMacro{llama-large_java_lparentheses-rparentheses}} {\tiny (\cellcolor{red!20}\UseMacro{llama-large_java_lparentheses-rparentheses_delta})} & \underline{\UseMacro{qwen-small_java_lparentheses-rparentheses}} {\tiny (\cellcolor{red!20}\UseMacro{qwen-small_java_lparentheses-rparentheses_delta})} & \UseMacro{qwen-medium_java_lparentheses-rparentheses} {\tiny (\cellcolor{red!20}\UseMacro{qwen-medium_java_lparentheses-rparentheses_delta})} & \UseMacro{qwen-large_java_lparentheses-rparentheses} {\tiny (\cellcolor{green!20}\UseMacro{qwen-large_java_lparentheses-rparentheses_delta})} & \UseMacro{dscoder-small_java_lparentheses-rparentheses} {\tiny (\cellcolor{red!20}\UseMacro{dscoder-small_java_lparentheses-rparentheses_delta})} & \UseMacro{dscoder-medium_java_lparentheses-rparentheses} {\tiny (\cellcolor{red!20}\UseMacro{dscoder-medium_java_lparentheses-rparentheses_delta})} & \UseMacro{dscoder-large_java_lparentheses-rparentheses} {\tiny (\cellcolor{green!20}\UseMacro{dscoder-large_java_lparentheses-rparentheses_delta})} & \UseMacro{average_java_lparentheses-rparentheses} {\tiny (\cellcolor{red!20}\UseMacro{average_java_lparentheses-rparentheses_delta})} \\
\UseMacro{period-name-no} & \UseMacro{llama-small_java_period-name} {\tiny (\cellcolor{red!20}\UseMacro{llama-small_java_period-name_delta})} & \UseMacro{llama-medium_java_period-name} {\tiny (\cellcolor{red!20}\UseMacro{llama-medium_java_period-name_delta})} & \UseMacro{llama-large_java_period-name} {\tiny (\UseMacro{llama-large_java_period-name_delta})} & \UseMacro{qwen-small_java_period-name} {\tiny (\cellcolor{red!20}\UseMacro{qwen-small_java_period-name_delta})} & \UseMacro{qwen-medium_java_period-name} {\tiny (\cellcolor{red!20}\UseMacro{qwen-medium_java_period-name_delta})} & \UseMacro{qwen-large_java_period-name} {\tiny (\cellcolor{red!20}\UseMacro{qwen-large_java_period-name_delta})} & \UseMacro{dscoder-small_java_period-name} {\tiny (\cellcolor{red!20}\UseMacro{dscoder-small_java_period-name_delta})} & \UseMacro{dscoder-medium_java_period-name} {\tiny (\cellcolor{red!20}\UseMacro{dscoder-medium_java_period-name_delta})} & \textbf{\UseMacro{dscoder-large_java_period-name}} {\tiny (\cellcolor{green!20}\UseMacro{dscoder-large_java_period-name_delta})} & \UseMacro{average_java_period-name} {\tiny (\cellcolor{red!20}\UseMacro{average_java_period-name_delta})} \\
\UseMacro{lparentheses-name-no} & \UseMacro{llama-small_java_lparentheses-name} {\tiny (\cellcolor{red!20}\UseMacro{llama-small_java_lparentheses-name_delta})} & \UseMacro{llama-medium_java_lparentheses-name} {\tiny (\cellcolor{red!20}\UseMacro{llama-medium_java_lparentheses-name_delta})} & \UseMacro{llama-large_java_lparentheses-name} {\tiny (\cellcolor{red!20}\UseMacro{llama-large_java_lparentheses-name_delta})} & \UseMacro{qwen-small_java_lparentheses-name} {\tiny (\cellcolor{green!20}\UseMacro{qwen-small_java_lparentheses-name_delta})} & \UseMacro{qwen-medium_java_lparentheses-name} {\tiny (\UseMacro{qwen-medium_java_lparentheses-name_delta})} & \textbf{\UseMacro{qwen-large_java_lparentheses-name}} {\tiny (\cellcolor{green!20}\UseMacro{qwen-large_java_lparentheses-name_delta})} & \UseMacro{dscoder-small_java_lparentheses-name} {\tiny (\cellcolor{red!20}\UseMacro{dscoder-small_java_lparentheses-name_delta})} & \UseMacro{dscoder-medium_java_lparentheses-name} {\tiny (\cellcolor{red!20}\UseMacro{dscoder-medium_java_lparentheses-name_delta})} & \UseMacro{dscoder-large_java_lparentheses-name} {\tiny (\cellcolor{green!20}\UseMacro{dscoder-large_java_lparentheses-name_delta})} & \UseMacro{average_java_lparentheses-name} {\tiny (\cellcolor{red!20}\UseMacro{average_java_lparentheses-name_delta})} \\
\UseMacro{op-name-no} & \UseMacro{llama-small_java_op-name} {\tiny (\cellcolor{red!20}\UseMacro{llama-small_java_op-name_delta})} & \UseMacro{llama-medium_java_op-name} {\tiny (\cellcolor{red!20}\UseMacro{llama-medium_java_op-name_delta})} & \UseMacro{llama-large_java_op-name} {\tiny (\cellcolor{red!20}\UseMacro{llama-large_java_op-name_delta})} & \textbf{\UseMacro{qwen-small_java_op-name}} {\tiny (\cellcolor{green!20}\UseMacro{qwen-small_java_op-name_delta})} & \underline{\UseMacro{qwen-medium_java_op-name}} {\tiny (\cellcolor{red!20}\UseMacro{qwen-medium_java_op-name_delta})} & \UseMacro{qwen-large_java_op-name} {\tiny (\cellcolor{red!20}\UseMacro{qwen-large_java_op-name_delta})} & \UseMacro{dscoder-small_java_op-name} {\tiny (\cellcolor{red!20}\UseMacro{dscoder-small_java_op-name_delta})} & \UseMacro{dscoder-medium_java_op-name} {\tiny (\cellcolor{red!20}\UseMacro{dscoder-medium_java_op-name_delta})} & \UseMacro{dscoder-large_java_op-name} {\tiny (\cellcolor{green!20}\UseMacro{dscoder-large_java_op-name_delta})} & \UseMacro{average_java_op-name} {\tiny (\cellcolor{red!20}\UseMacro{average_java_op-name_delta})} \\
\UseMacro{op-all-no} & \underline{\UseMacro{llama-small_java_op-all}} {\tiny (\cellcolor{red!20}\UseMacro{llama-small_java_op-all_delta})} & \underline{\UseMacro{llama-medium_java_op-all}} {\tiny (\cellcolor{red!20}\UseMacro{llama-medium_java_op-all_delta})} & \UseMacro{llama-large_java_op-all} {\tiny (\cellcolor{red!20}\UseMacro{llama-large_java_op-all_delta})} & \UseMacro{qwen-small_java_op-all} {\tiny (\cellcolor{green!20}\UseMacro{qwen-small_java_op-all_delta})} & \UseMacro{qwen-medium_java_op-all} {\tiny (\cellcolor{red!20}\UseMacro{qwen-medium_java_op-all_delta})} & \UseMacro{qwen-large_java_op-all} {\tiny (\cellcolor{red!20}\UseMacro{qwen-large_java_op-all_delta})} & \underline{\UseMacro{dscoder-small_java_op-all}} {\tiny (\cellcolor{red!20}\UseMacro{dscoder-small_java_op-all_delta})} & \underline{\UseMacro{dscoder-medium_java_op-all}} {\tiny (\cellcolor{red!20}\UseMacro{dscoder-medium_java_op-all_delta})} & \underline{\UseMacro{dscoder-large_java_op-all}} {\tiny (\cellcolor{red!20}\UseMacro{dscoder-large_java_op-all_delta})} & \underline{\UseMacro{average_java_op-all}} {\tiny (\cellcolor{red!20}\UseMacro{average_java_op-all_delta})} \\
\midrule
\multicolumn{11}{c}{\UseMacro{TH-python-results}} \\
\midrule
\rowcolor[HTML]{C0C0C0} \UseMacro{baseline} & \UseMacro{llama-small_python_baseline}\phantom{ \tiny{(+0.00)}} & \UseMacro{llama-medium_python_baseline}\phantom{ \tiny{(+0.00)}} & \UseMacro{llama-large_python_baseline}\phantom{ \tiny{(+0.00)}} & \UseMacro{qwen-small_python_baseline}\phantom{ \tiny{(+0.00)}} & \UseMacro{qwen-medium_python_baseline}\phantom{ \tiny{(+0.00)}} & \UseMacro{qwen-large_python_baseline}\phantom{ \tiny{(+0.00)}} & \UseMacro{dscoder-small_python_baseline}\phantom{ \tiny{(+0.00)}} & \UseMacro{dscoder-medium_python_baseline}\phantom{ \tiny{(+0.00)}} & \UseMacro{dscoder-large_python_baseline}\phantom{ \tiny{(+0.00)}} & \UseMacro{average_python_baseline}\phantom{ \tiny{(+0.00)}} \\
\UseMacro{snake-case-camel-case-no} & \UseMacro{llama-small_python_snake-case-camel-case} {\tiny (\cellcolor{green!20}\UseMacro{llama-small_python_snake-case-camel-case_delta})} & \textbf{\UseMacro{llama-medium_python_snake-case-camel-case}} {\tiny (\cellcolor{green!20}\UseMacro{llama-medium_python_snake-case-camel-case_delta})} & \UseMacro{llama-large_python_snake-case-camel-case} {\tiny (\cellcolor{red!20}\UseMacro{llama-large_python_snake-case-camel-case_delta})} & \UseMacro{qwen-small_python_snake-case-camel-case} {\tiny (\cellcolor{red!20}\UseMacro{qwen-small_python_snake-case-camel-case_delta})} & \textbf{\UseMacro{qwen-medium_python_snake-case-camel-case}} {\tiny (\cellcolor{green!20}\UseMacro{qwen-medium_python_snake-case-camel-case_delta})} & \textbf{\UseMacro{qwen-large_python_snake-case-camel-case}} {\tiny (\cellcolor{green!20}\UseMacro{qwen-large_python_snake-case-camel-case_delta})} & \UseMacro{dscoder-small_python_snake-case-camel-case} {\tiny (\cellcolor{red!20}\UseMacro{dscoder-small_python_snake-case-camel-case_delta})} & \UseMacro{dscoder-medium_python_snake-case-camel-case} {\tiny (\cellcolor{red!20}\UseMacro{dscoder-medium_python_snake-case-camel-case_delta})} & \textbf{\UseMacro{dscoder-large_python_snake-case-camel-case}} {\tiny (\cellcolor{green!20}\UseMacro{dscoder-large_python_snake-case-camel-case_delta})} & \textbf{\UseMacro{average_python_snake-case-camel-case}} {\tiny (\cellcolor{green!20}\UseMacro{average_python_snake-case-camel-case_delta})} \\
\UseMacro{snake-case-pascal-case-no} & \UseMacro{llama-small_python_snake-case-pascal-case} {\tiny (\cellcolor{red!20}\UseMacro{llama-small_python_snake-case-pascal-case_delta})} & \UseMacro{llama-medium_python_snake-case-pascal-case} {\tiny (\cellcolor{green!20}\UseMacro{llama-medium_python_snake-case-pascal-case_delta})} & \UseMacro{llama-large_python_snake-case-pascal-case} {\tiny (\cellcolor{red!20}\UseMacro{llama-large_python_snake-case-pascal-case_delta})} & \UseMacro{qwen-small_python_snake-case-pascal-case} {\tiny (\cellcolor{red!20}\UseMacro{qwen-small_python_snake-case-pascal-case_delta})} & \UseMacro{qwen-medium_python_snake-case-pascal-case} {\tiny (\cellcolor{green!20}\UseMacro{qwen-medium_python_snake-case-pascal-case_delta})} & \UseMacro{qwen-large_python_snake-case-pascal-case} {\tiny (\cellcolor{green!20}\UseMacro{qwen-large_python_snake-case-pascal-case_delta})} & \UseMacro{dscoder-small_python_snake-case-pascal-case} {\tiny (\cellcolor{red!20}\UseMacro{dscoder-small_python_snake-case-pascal-case_delta})} & \UseMacro{dscoder-medium_python_snake-case-pascal-case} {\tiny (\cellcolor{red!20}\UseMacro{dscoder-medium_python_snake-case-pascal-case_delta})} & \UseMacro{dscoder-large_python_snake-case-pascal-case} {\tiny (\cellcolor{green!20}\UseMacro{dscoder-large_python_snake-case-pascal-case_delta})} & \UseMacro{average_python_snake-case-pascal-case} {\tiny (\cellcolor{red!20}\UseMacro{average_python_snake-case-pascal-case_delta})} \\
\UseMacro{snake-case-screaming-snake-case-no} & \UseMacro{llama-small_python_snake-case-screaming-snake-case} {\tiny (\cellcolor{red!20}\UseMacro{llama-small_python_snake-case-screaming-snake-case_delta})} & \UseMacro{llama-medium_python_snake-case-screaming-snake-case} {\tiny (\cellcolor{green!20}\UseMacro{llama-medium_python_snake-case-screaming-snake-case_delta})} & \underline{\UseMacro{llama-large_python_snake-case-screaming-snake-case}} {\tiny (\cellcolor{red!20}\UseMacro{llama-large_python_snake-case-screaming-snake-case_delta})} & \UseMacro{qwen-small_python_snake-case-screaming-snake-case} {\tiny (\cellcolor{red!20}\UseMacro{qwen-small_python_snake-case-screaming-snake-case_delta})} & \UseMacro{qwen-medium_python_snake-case-screaming-snake-case} {\tiny (\UseMacro{qwen-medium_python_snake-case-screaming-snake-case_delta})} & \UseMacro{qwen-large_python_snake-case-screaming-snake-case} {\tiny (\cellcolor{green!20}\UseMacro{qwen-large_python_snake-case-screaming-snake-case_delta})} & \underline{\UseMacro{dscoder-small_python_snake-case-screaming-snake-case}} {\tiny (\cellcolor{red!20}\UseMacro{dscoder-small_python_snake-case-screaming-snake-case_delta})} & \UseMacro{dscoder-medium_python_snake-case-screaming-snake-case} {\tiny (\cellcolor{red!20}\UseMacro{dscoder-medium_python_snake-case-screaming-snake-case_delta})} & \UseMacro{dscoder-large_python_snake-case-screaming-snake-case} {\tiny (\cellcolor{red!20}\UseMacro{dscoder-large_python_snake-case-screaming-snake-case_delta})} & \UseMacro{average_python_snake-case-screaming-snake-case} {\tiny (\cellcolor{red!20}\UseMacro{average_python_snake-case-screaming-snake-case_delta})} \\
\UseMacro{op-dash-no} & \UseMacro{llama-small_python_op-dash} {\tiny (\cellcolor{green!20}\UseMacro{llama-small_python_op-dash_delta})} & \UseMacro{llama-medium_python_op-dash} {\tiny (\cellcolor{green!20}\UseMacro{llama-medium_python_op-dash_delta})} & \UseMacro{llama-large_python_op-dash} {\tiny (\cellcolor{red!20}\UseMacro{llama-large_python_op-dash_delta})} & \UseMacro{qwen-small_python_op-dash} {\tiny (\cellcolor{red!20}\UseMacro{qwen-small_python_op-dash_delta})} & \UseMacro{qwen-medium_python_op-dash} {\tiny (\UseMacro{qwen-medium_python_op-dash_delta})} & \UseMacro{qwen-large_python_op-dash} {\tiny (\cellcolor{green!20}\UseMacro{qwen-large_python_op-dash_delta})} & \UseMacro{dscoder-small_python_op-dash} {\tiny (\cellcolor{red!20}\UseMacro{dscoder-small_python_op-dash_delta})} & \UseMacro{dscoder-medium_python_op-dash} {\tiny (\cellcolor{green!20}\UseMacro{dscoder-medium_python_op-dash_delta})} & \UseMacro{dscoder-large_python_op-dash} {\tiny (\cellcolor{red!20}\UseMacro{dscoder-large_python_op-dash_delta})} & \UseMacro{average_python_op-dash} {\tiny (\cellcolor{green!20}\UseMacro{average_python_op-dash_delta})} \\
\UseMacro{op-lsquarebracket-no} & \UseMacro{llama-small_python_op-lsquarebracket} {\tiny (\cellcolor{green!20}\UseMacro{llama-small_python_op-lsquarebracket_delta})} & \UseMacro{llama-medium_python_op-lsquarebracket} {\tiny (\cellcolor{green!20}\UseMacro{llama-medium_python_op-lsquarebracket_delta})} & \UseMacro{llama-large_python_op-lsquarebracket} {\tiny (\cellcolor{red!20}\UseMacro{llama-large_python_op-lsquarebracket_delta})} & \UseMacro{qwen-small_python_op-lsquarebracket} {\tiny (\cellcolor{red!20}\UseMacro{qwen-small_python_op-lsquarebracket_delta})} & \UseMacro{qwen-medium_python_op-lsquarebracket} {\tiny (\cellcolor{green!20}\UseMacro{qwen-medium_python_op-lsquarebracket_delta})} & \UseMacro{qwen-large_python_op-lsquarebracket} {\tiny (\cellcolor{red!20}\UseMacro{qwen-large_python_op-lsquarebracket_delta})} & \UseMacro{dscoder-small_python_op-lsquarebracket} {\tiny (\cellcolor{red!20}\UseMacro{dscoder-small_python_op-lsquarebracket_delta})} & \UseMacro{dscoder-medium_python_op-lsquarebracket} {\tiny (\cellcolor{green!20}\UseMacro{dscoder-medium_python_op-lsquarebracket_delta})} & \UseMacro{dscoder-large_python_op-lsquarebracket} {\tiny (\cellcolor{red!20}\UseMacro{dscoder-large_python_op-lsquarebracket_delta})} & \UseMacro{average_python_op-lsquarebracket} {\tiny (\cellcolor{red!20}\UseMacro{average_python_op-lsquarebracket_delta})} \\
\UseMacro{rsquarebracket-rparentheses-no} & \UseMacro{llama-small_python_rsquarebracket-rparentheses} {\tiny (\cellcolor{green!20}\UseMacro{llama-small_python_rsquarebracket-rparentheses_delta})} & \UseMacro{llama-medium_python_rsquarebracket-rparentheses} {\tiny (\cellcolor{green!20}\UseMacro{llama-medium_python_rsquarebracket-rparentheses_delta})} & \textbf{\UseMacro{llama-large_python_rsquarebracket-rparentheses}} {\tiny (\cellcolor{green!20}\UseMacro{llama-large_python_rsquarebracket-rparentheses_delta})} & \UseMacro{qwen-small_python_rsquarebracket-rparentheses} {\tiny (\cellcolor{red!20}\UseMacro{qwen-small_python_rsquarebracket-rparentheses_delta})} & \UseMacro{qwen-medium_python_rsquarebracket-rparentheses} {\tiny (\UseMacro{qwen-medium_python_rsquarebracket-rparentheses_delta})} & \underline{\UseMacro{qwen-large_python_rsquarebracket-rparentheses}} {\tiny (\cellcolor{red!20}\UseMacro{qwen-large_python_rsquarebracket-rparentheses_delta})} & \UseMacro{dscoder-small_python_rsquarebracket-rparentheses} {\tiny (\UseMacro{dscoder-small_python_rsquarebracket-rparentheses_delta})} & \UseMacro{dscoder-medium_python_rsquarebracket-rparentheses} {\tiny (\UseMacro{dscoder-medium_python_rsquarebracket-rparentheses_delta})} & \UseMacro{dscoder-large_python_rsquarebracket-rparentheses} {\tiny (\cellcolor{red!20}\UseMacro{dscoder-large_python_rsquarebracket-rparentheses_delta})} & \UseMacro{average_python_rsquarebracket-rparentheses} {\tiny (\cellcolor{red!20}\UseMacro{average_python_rsquarebracket-rparentheses_delta})} \\
\UseMacro{op-rsquarebracket-no} & \UseMacro{llama-small_python_op-rsquarebracket} {\tiny (\cellcolor{red!20}\UseMacro{llama-small_python_op-rsquarebracket_delta})} & \UseMacro{llama-medium_python_op-rsquarebracket} {\tiny (\cellcolor{green!20}\UseMacro{llama-medium_python_op-rsquarebracket_delta})} & \UseMacro{llama-large_python_op-rsquarebracket} {\tiny (\cellcolor{red!20}\UseMacro{llama-large_python_op-rsquarebracket_delta})} & \textbf{\UseMacro{qwen-small_python_op-rsquarebracket}} {\tiny (\cellcolor{green!20}\UseMacro{qwen-small_python_op-rsquarebracket_delta})} & \UseMacro{qwen-medium_python_op-rsquarebracket} {\tiny (\cellcolor{red!20}\UseMacro{qwen-medium_python_op-rsquarebracket_delta})} & \UseMacro{qwen-large_python_op-rsquarebracket} {\tiny (\cellcolor{green!20}\UseMacro{qwen-large_python_op-rsquarebracket_delta})} & \UseMacro{dscoder-small_python_op-rsquarebracket} {\tiny (\cellcolor{red!20}\UseMacro{dscoder-small_python_op-rsquarebracket_delta})} & \textbf{\UseMacro{dscoder-medium_python_op-rsquarebracket}} {\tiny (\cellcolor{green!20}\UseMacro{dscoder-medium_python_op-rsquarebracket_delta})} & \UseMacro{dscoder-large_python_op-rsquarebracket} {\tiny (\cellcolor{red!20}\UseMacro{dscoder-large_python_op-rsquarebracket_delta})} & \UseMacro{average_python_op-rsquarebracket} {\tiny (\cellcolor{red!20}\UseMacro{average_python_op-rsquarebracket_delta})} \\
\UseMacro{lsquarebracket-name-no} & \UseMacro{llama-small_python_lsquarebracket-name} {\tiny (\cellcolor{green!20}\UseMacro{llama-small_python_lsquarebracket-name_delta})} & \UseMacro{llama-medium_python_lsquarebracket-name} {\tiny (\cellcolor{red!20}\UseMacro{llama-medium_python_lsquarebracket-name_delta})} & \UseMacro{llama-large_python_lsquarebracket-name} {\tiny (\cellcolor{red!20}\UseMacro{llama-large_python_lsquarebracket-name_delta})} & \UseMacro{qwen-small_python_lsquarebracket-name} {\tiny (\UseMacro{qwen-small_python_lsquarebracket-name_delta})} & \UseMacro{qwen-medium_python_lsquarebracket-name} {\tiny (\cellcolor{red!20}\UseMacro{qwen-medium_python_lsquarebracket-name_delta})} & \UseMacro{qwen-large_python_lsquarebracket-name} {\tiny (\cellcolor{green!20}\UseMacro{qwen-large_python_lsquarebracket-name_delta})} & \UseMacro{dscoder-small_python_lsquarebracket-name} {\tiny (\cellcolor{red!20}\UseMacro{dscoder-small_python_lsquarebracket-name_delta})} & \UseMacro{dscoder-medium_python_lsquarebracket-name} {\tiny (\cellcolor{green!20}\UseMacro{dscoder-medium_python_lsquarebracket-name_delta})} & \UseMacro{dscoder-large_python_lsquarebracket-name} {\tiny (\cellcolor{red!20}\UseMacro{dscoder-large_python_lsquarebracket-name_delta})} & \UseMacro{average_python_lsquarebracket-name} {\tiny (\cellcolor{red!20}\UseMacro{average_python_lsquarebracket-name_delta})} \\
\UseMacro{rparentheses-colon-no} & \UseMacro{llama-small_python_rparentheses-colon} {\tiny (\cellcolor{red!20}\UseMacro{llama-small_python_rparentheses-colon_delta})} & \UseMacro{llama-medium_python_rparentheses-colon} {\tiny (\cellcolor{green!20}\UseMacro{llama-medium_python_rparentheses-colon_delta})} & \UseMacro{llama-large_python_rparentheses-colon} {\tiny (\cellcolor{green!20}\UseMacro{llama-large_python_rparentheses-colon_delta})} & \UseMacro{qwen-small_python_rparentheses-colon} {\tiny (\UseMacro{qwen-small_python_rparentheses-colon_delta})} & \UseMacro{qwen-medium_python_rparentheses-colon} {\tiny (\cellcolor{red!20}\UseMacro{qwen-medium_python_rparentheses-colon_delta})} & \UseMacro{qwen-large_python_rparentheses-colon} {\tiny (\cellcolor{green!20}\UseMacro{qwen-large_python_rparentheses-colon_delta})} & \UseMacro{dscoder-small_python_rparentheses-colon} {\tiny (\cellcolor{red!20}\UseMacro{dscoder-small_python_rparentheses-colon_delta})} & \UseMacro{dscoder-medium_python_rparentheses-colon} {\tiny (\cellcolor{green!20}\UseMacro{dscoder-medium_python_rparentheses-colon_delta})} & \UseMacro{dscoder-large_python_rparentheses-colon} {\tiny (\cellcolor{red!20}\UseMacro{dscoder-large_python_rparentheses-colon_delta})} & \UseMacro{average_python_rparentheses-colon} {\tiny (\cellcolor{red!20}\UseMacro{average_python_rparentheses-colon_delta})} \\
\UseMacro{rparentheses-rparentheses-no} & \UseMacro{llama-small_python_rparentheses-rparentheses} {\tiny (\cellcolor{red!20}\UseMacro{llama-small_python_rparentheses-rparentheses_delta})} & \UseMacro{llama-medium_python_rparentheses-rparentheses} {\tiny (\cellcolor{green!20}\UseMacro{llama-medium_python_rparentheses-rparentheses_delta})} & \UseMacro{llama-large_python_rparentheses-rparentheses} {\tiny (\cellcolor{green!20}\UseMacro{llama-large_python_rparentheses-rparentheses_delta})} & \UseMacro{qwen-small_python_rparentheses-rparentheses} {\tiny (\cellcolor{red!20}\UseMacro{qwen-small_python_rparentheses-rparentheses_delta})} & \UseMacro{qwen-medium_python_rparentheses-rparentheses} {\tiny (\cellcolor{green!20}\UseMacro{qwen-medium_python_rparentheses-rparentheses_delta})} & \UseMacro{qwen-large_python_rparentheses-rparentheses} {\tiny (\cellcolor{green!20}\UseMacro{qwen-large_python_rparentheses-rparentheses_delta})} & \UseMacro{dscoder-small_python_rparentheses-rparentheses} {\tiny (\cellcolor{red!20}\UseMacro{dscoder-small_python_rparentheses-rparentheses_delta})} & \UseMacro{dscoder-medium_python_rparentheses-rparentheses} {\tiny (\cellcolor{green!20}\UseMacro{dscoder-medium_python_rparentheses-rparentheses_delta})} & \UseMacro{dscoder-large_python_rparentheses-rparentheses} {\tiny (\cellcolor{red!20}\UseMacro{dscoder-large_python_rparentheses-rparentheses_delta})} & \UseMacro{average_python_rparentheses-rparentheses} {\tiny (\cellcolor{red!20}\UseMacro{average_python_rparentheses-rparentheses_delta})} \\
\UseMacro{lparentheses-rparentheses-no} & \UseMacro{llama-small_python_lparentheses-rparentheses} {\tiny (\cellcolor{red!20}\UseMacro{llama-small_python_lparentheses-rparentheses_delta})} & \underline{\UseMacro{llama-medium_python_lparentheses-rparentheses}} {\tiny (\cellcolor{red!20}\UseMacro{llama-medium_python_lparentheses-rparentheses_delta})} & \UseMacro{llama-large_python_lparentheses-rparentheses} {\tiny (\cellcolor{red!20}\UseMacro{llama-large_python_lparentheses-rparentheses_delta})} & \UseMacro{qwen-small_python_lparentheses-rparentheses} {\tiny (\cellcolor{red!20}\UseMacro{qwen-small_python_lparentheses-rparentheses_delta})} & \UseMacro{qwen-medium_python_lparentheses-rparentheses} {\tiny (\cellcolor{red!20}\UseMacro{qwen-medium_python_lparentheses-rparentheses_delta})} & \UseMacro{qwen-large_python_lparentheses-rparentheses} {\tiny (\cellcolor{red!20}\UseMacro{qwen-large_python_lparentheses-rparentheses_delta})} & \textbf{\UseMacro{dscoder-small_python_lparentheses-rparentheses}} {\tiny (\cellcolor{green!20}\UseMacro{dscoder-small_python_lparentheses-rparentheses_delta})} & \UseMacro{dscoder-medium_python_lparentheses-rparentheses} {\tiny (\cellcolor{red!20}\UseMacro{dscoder-medium_python_lparentheses-rparentheses_delta})} & \UseMacro{dscoder-large_python_lparentheses-rparentheses} {\tiny (\cellcolor{red!20}\UseMacro{dscoder-large_python_lparentheses-rparentheses_delta})} & \UseMacro{average_python_lparentheses-rparentheses} {\tiny (\cellcolor{red!20}\UseMacro{average_python_lparentheses-rparentheses_delta})} \\
\UseMacro{period-name-no} & \UseMacro{llama-small_python_period-name} {\tiny (\UseMacro{llama-small_python_period-name_delta})} & \UseMacro{llama-medium_python_period-name} {\tiny (\cellcolor{green!20}\UseMacro{llama-medium_python_period-name_delta})} & \UseMacro{llama-large_python_period-name} {\tiny (\cellcolor{red!20}\UseMacro{llama-large_python_period-name_delta})} & \UseMacro{qwen-small_python_period-name} {\tiny (\cellcolor{red!20}\UseMacro{qwen-small_python_period-name_delta})} & \UseMacro{qwen-medium_python_period-name} {\tiny (\cellcolor{red!20}\UseMacro{qwen-medium_python_period-name_delta})} & \UseMacro{qwen-large_python_period-name} {\tiny (\cellcolor{green!20}\UseMacro{qwen-large_python_period-name_delta})} & \UseMacro{dscoder-small_python_period-name} {\tiny (\cellcolor{red!20}\UseMacro{dscoder-small_python_period-name_delta})} & \UseMacro{dscoder-medium_python_period-name} {\tiny (\cellcolor{red!20}\UseMacro{dscoder-medium_python_period-name_delta})} & \UseMacro{dscoder-large_python_period-name} {\tiny (\cellcolor{red!20}\UseMacro{dscoder-large_python_period-name_delta})} & \UseMacro{average_python_period-name} {\tiny (\cellcolor{red!20}\UseMacro{average_python_period-name_delta})} \\
\UseMacro{lparentheses-name-no} & \UseMacro{llama-small_python_lparentheses-name} {\tiny (\cellcolor{green!20}\UseMacro{llama-small_python_lparentheses-name_delta})} & \UseMacro{llama-medium_python_lparentheses-name} {\tiny (\UseMacro{llama-medium_python_lparentheses-name_delta})} & \UseMacro{llama-large_python_lparentheses-name} {\tiny (\UseMacro{llama-large_python_lparentheses-name_delta})} & \UseMacro{qwen-small_python_lparentheses-name} {\tiny (\cellcolor{red!20}\UseMacro{qwen-small_python_lparentheses-name_delta})} & \UseMacro{qwen-medium_python_lparentheses-name} {\tiny (\cellcolor{red!20}\UseMacro{qwen-medium_python_lparentheses-name_delta})} & \UseMacro{qwen-large_python_lparentheses-name} {\tiny (\cellcolor{green!20}\UseMacro{qwen-large_python_lparentheses-name_delta})} & \UseMacro{dscoder-small_python_lparentheses-name} {\tiny (\cellcolor{red!20}\UseMacro{dscoder-small_python_lparentheses-name_delta})} & \UseMacro{dscoder-medium_python_lparentheses-name} {\tiny (\cellcolor{red!20}\UseMacro{dscoder-medium_python_lparentheses-name_delta})} & \UseMacro{dscoder-large_python_lparentheses-name} {\tiny (\cellcolor{red!20}\UseMacro{dscoder-large_python_lparentheses-name_delta})} & \UseMacro{average_python_lparentheses-name} {\tiny (\cellcolor{red!20}\UseMacro{average_python_lparentheses-name_delta})} \\
\UseMacro{op-name-no} & \textbf{\UseMacro{llama-small_python_op-name}} {\tiny (\cellcolor{green!20}\UseMacro{llama-small_python_op-name_delta})} & \UseMacro{llama-medium_python_op-name} {\tiny (\cellcolor{green!20}\UseMacro{llama-medium_python_op-name_delta})} & \UseMacro{llama-large_python_op-name} {\tiny (\cellcolor{red!20}\UseMacro{llama-large_python_op-name_delta})} & \UseMacro{qwen-small_python_op-name} {\tiny (\cellcolor{red!20}\UseMacro{qwen-small_python_op-name_delta})} & \underline{\UseMacro{qwen-medium_python_op-name}} {\tiny (\cellcolor{red!20}\UseMacro{qwen-medium_python_op-name_delta})} & \UseMacro{qwen-large_python_op-name} {\tiny (\cellcolor{green!20}\UseMacro{qwen-large_python_op-name_delta})} & \UseMacro{dscoder-small_python_op-name} {\tiny (\cellcolor{red!20}\UseMacro{dscoder-small_python_op-name_delta})} & \underline{\UseMacro{dscoder-medium_python_op-name}} {\tiny (\cellcolor{red!20}\UseMacro{dscoder-medium_python_op-name_delta})} & \underline{\UseMacro{dscoder-large_python_op-name}} {\tiny (\cellcolor{red!20}\UseMacro{dscoder-large_python_op-name_delta})} & \UseMacro{average_python_op-name} {\tiny (\cellcolor{red!20}\UseMacro{average_python_op-name_delta})} \\
\UseMacro{op-all-no} & \underline{\UseMacro{llama-small_python_op-all}} {\tiny (\cellcolor{red!20}\UseMacro{llama-small_python_op-all_delta})} & \UseMacro{llama-medium_python_op-all} {\tiny (\UseMacro{llama-medium_python_op-all_delta})} & \UseMacro{llama-large_python_op-all} {\tiny (\cellcolor{red!20}\UseMacro{llama-large_python_op-all_delta})} & \underline{\UseMacro{qwen-small_python_op-all}} {\tiny (\cellcolor{red!20}\UseMacro{qwen-small_python_op-all_delta})} & \UseMacro{qwen-medium_python_op-all} {\tiny (\cellcolor{red!20}\UseMacro{qwen-medium_python_op-all_delta})} & \UseMacro{qwen-large_python_op-all} {\tiny (\cellcolor{red!20}\UseMacro{qwen-large_python_op-all_delta})} & \UseMacro{dscoder-small_python_op-all} {\tiny (\cellcolor{red!20}\UseMacro{dscoder-small_python_op-all_delta})} & \UseMacro{dscoder-medium_python_op-all} {\tiny (\cellcolor{green!20}\UseMacro{dscoder-medium_python_op-all_delta})} & \UseMacro{dscoder-large_python_op-all} {\tiny (\cellcolor{red!20}\UseMacro{dscoder-large_python_op-all_delta})} & \underline{\UseMacro{average_python_op-all}} {\tiny (\cellcolor{red!20}\UseMacro{average_python_op-all_delta})} \\
\bottomrule
\end{tabular}
}

}
{\scriptsize\makecell[l]{
Background color: baseline in grey, variants better than baseline in green, and variants worse than baseline in red.\\
The best variant is highlighted in \textbf{bold} and the worst variant is \uline{underlined}.
}}
\end{center}
\vspace{-10pt}
\end{table*}

\subsection{Metrics}
\label{sec:methodology:metrics}

Recall that each experiment on a given \{benchmark, model, \rewriterule{}\} triplet compares the \emph{baseline} outputs (given the original inputs) and the \emph{variant} outputs (given the inputs after applying \rewriterule).
The benchmark provides a set of tests to evaluate whether each output is correct or incorrect.
We define \acc as the percentage of correct outputs, and 
\textbf{\accdelta} as the variant's \acc minus the baseline's \acc.

The \accdelta metric, although intuitive, has two limitations:
(1)~\acc improvements and degradations on individual samples cancel out;
(2)~some samples may not be affected by a \rewriterule if the left-hand side substring does not appear in the input; the outputs of those samples will never change.
To address these, we introduce an unbiased metric called \textbf{\sensitivity}, defined as the percentage of the samples whose output correctness flips (from correct to incorrect or vice versa) out of the samples whose input is changed by the \rewriterule.
A lower \sensitivity indicates that the model is more robust against the token changes introduced by a \rewriterule;
when averaged across all \rewriterules, it reflects how sensitive the model is to the LLM-PL tokenization misalignment.

\section{Evaluation}
\label{sec:results}


\subsection{Results}
\label{sec:results:all-results}


\Cref{tab:accuracy} shows the \acc and \accdelta of each model on each \rewriterule.
We can observe that most \rewriterules cause measurable changes in model accuracy, ranging from -2.90 to +0.32 absolute percentage points if averaging across all models.
The largest \accdelta of \UseMacro{llama-medium_java_op-all_delta}\% happens on \UseMacro{llama-medium-short} for Java benchmarks, whose \acc drops from \UseMacro{llama-medium_java_baseline}\% to \UseMacro{llama-medium_java_op-all}\% when applying \rewriterule \UseMacro{op-all-no} (adding space after each operator).
Considering advances in LLM performance are sometimes claimed with around 1 percentage point margin, these accuracy deltas caused by simple \rewriterules are non-negligible.


\begin{figure}[t!]
\begin{center}
\begin{minipage}{.95\linewidth}
\includegraphics[width=\linewidth]{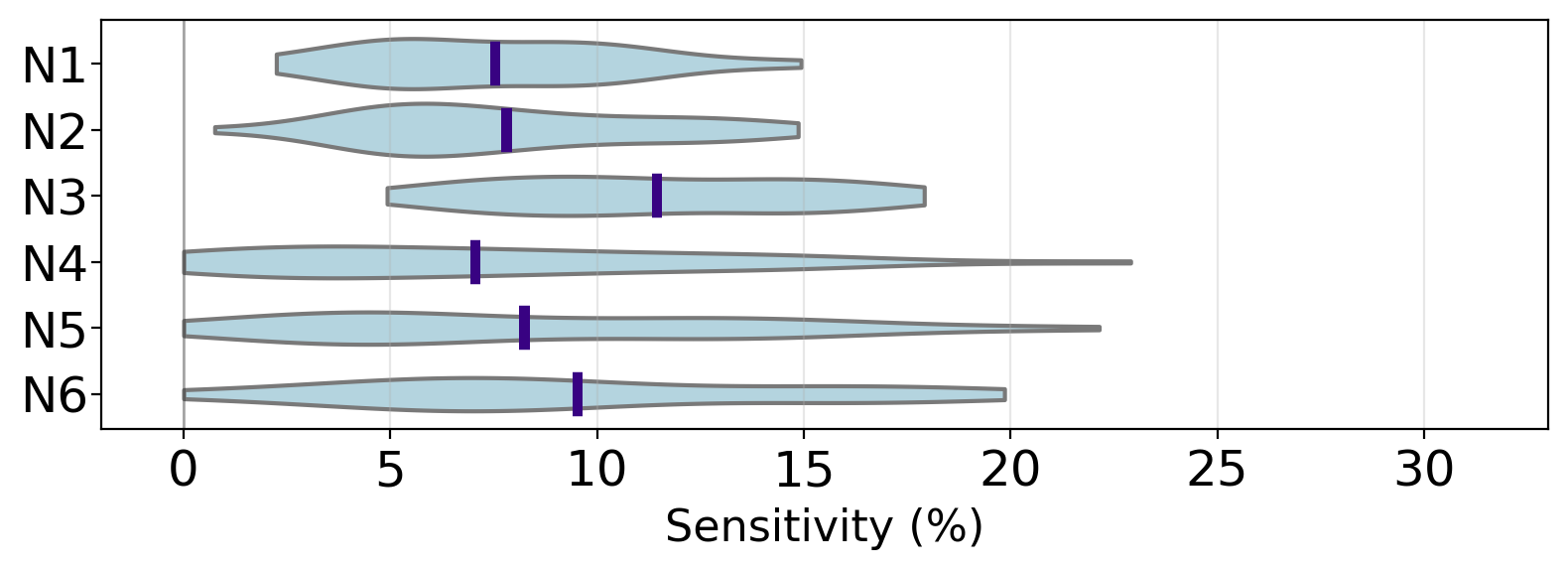}
\vspace{-20pt}
\subcaption{grouped by naming \rewriterule}
\label{fig:sensitivity:naming}
\end{minipage}\\\vspace{3pt}
\begin{minipage}{.95\linewidth}
\includegraphics[width=\linewidth]{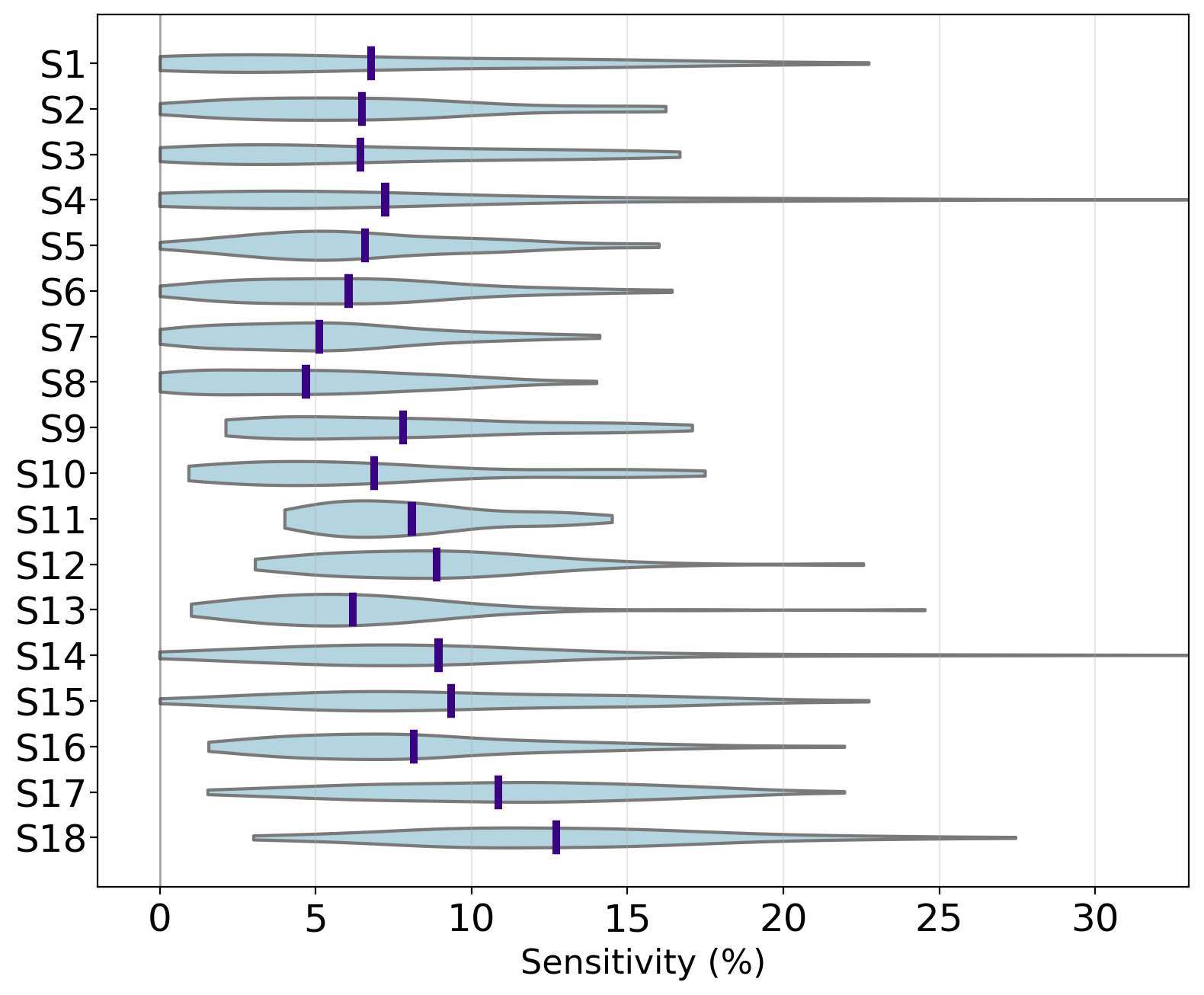}
\vspace{-20pt}
\subcaption{grouped by spacing \rewriterule}
\label{fig:sensitivity:spacing}
\end{minipage}\\\vspace{3pt}
\begin{minipage}{.95\linewidth}
\includegraphics[width=\linewidth]{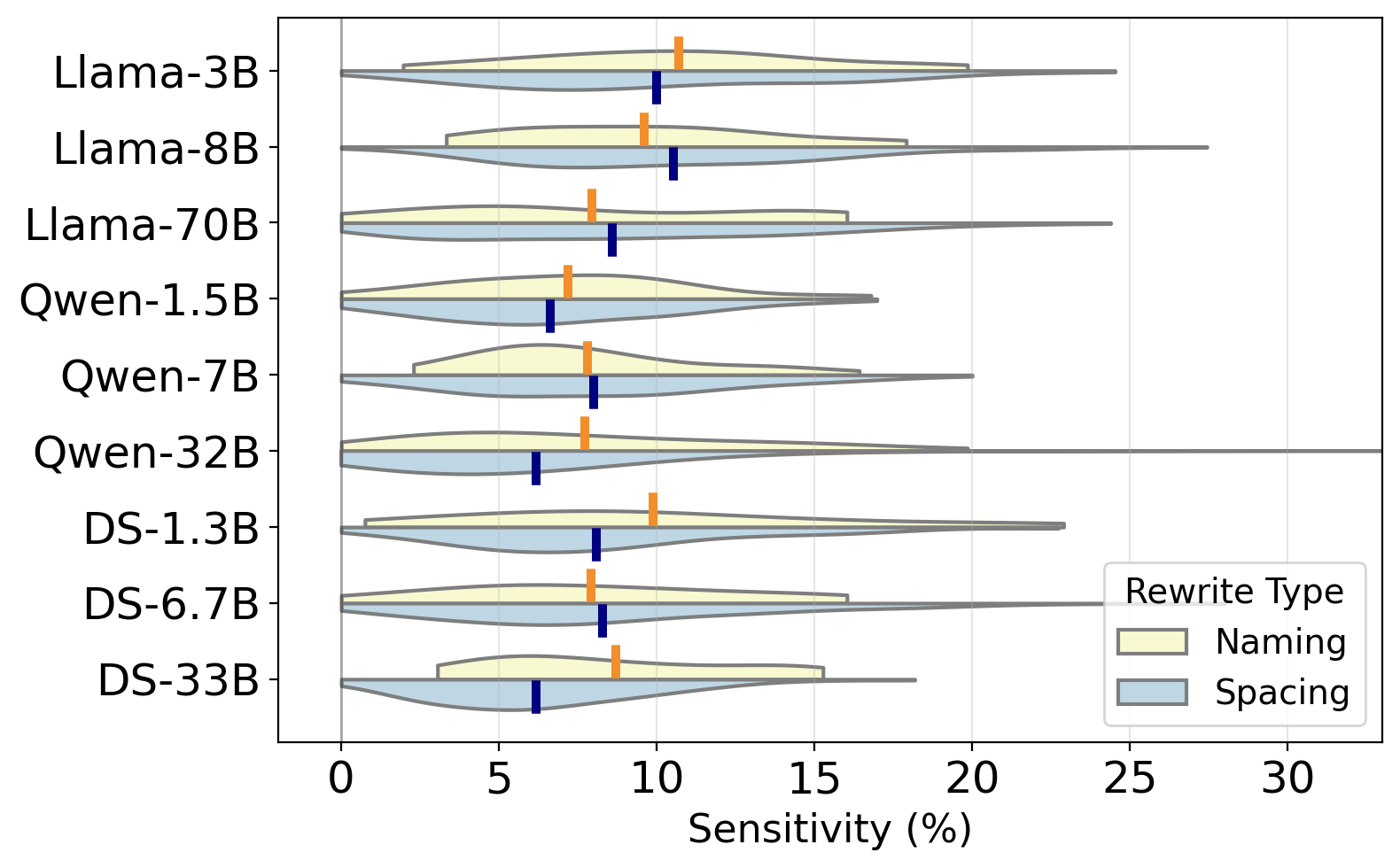}
\vspace{-20pt}
\subcaption{grouped by model}
\label{fig:sensitivity:model}
\end{minipage}
\vspace{-5pt}
\caption{Violin plots of \sensitivity distributions.}
\label{fig:sensitivity}
\end{center}
\vspace{-10pt}
\end{figure}

The impact of misaligned tokenization is more apparent in the \sensitivity metric, as shown in the distribution plots in \Cref{fig:sensitivity}.
The average \sensitivity is 9.26\% for naming \rewrites and 8.29\% for spacing \rewrites.
Among the naming \rewrites (\Cref{fig:sensitivity:naming}), LLMs are relatively less sensitive to transductions between \aCamelCase and \aSnakeCase (\UseMacro{camel-case-snake-case-no} and \UseMacro{snake-case-camel-case-no}),
likely because \aCamelCase and \aScreamingSnakeCase are less frequent.
This finding implies that the casing styles of identifiers, while technically convey no semantic meaning in PLs, are an important factor in LLMs' understanding of code.
In \Cref{fig:sensitivity:spacing}, we can see that LLMs' average \sensitivity is over 10\% for the two ``wildcard'' spacing \rewriterules (\UseMacro{op-name-no} and \UseMacro{op-all-no}).
Other spacing \rewriterules result in varying levels of \sensitivity, among which the most impactful ones are
\UseMacro{period-name-no} (adding space between period and identifier),
\UseMacro{lparentheses-rparentheses-no} (adding space between a pair of parentheses),
and \UseMacro{op-semicolon-no} (adding space between operator and semicolon).
In terms of the average \sensitivity of models (\Cref{fig:sensitivity:model}),
we observe that \llama models are more sensitive than the other two series,
but all models persist a non-negligible \sensitivity of at least \UseMacro{qwen-large_all-spacing_sensitivity}\% (\UseMacro{qwen-large-short} on spacing \rewriterules).



\subsection{Impact of Model Size}
\label{sec:results:model-size}

\begin{table}[t]
\begin{center}
\caption{Impact of model size on \sensitivity.}
\label{tab:model-size-sensitivity}
\vspace{-5pt}
\resizebox{.9\columnwidth}{!}{%
\begin{tabular}{@{}ll|rrr@{}}
\toprule
\textbf{Rewrite Rule} & \textbf{Model Series} & \textbf{S} & \textbf{M} & \textbf{L} \\
\midrule
\multirow{3}{*}{\textbf{Naming}}  & Llama-3 & \UseMacro{llama-small_all-naming_sensitivity} & \UseMacro{llama-medium_all-naming_sensitivity} & \UseMacro{llama-large_all-naming_sensitivity} \\
  & Qwen2.5-Coder & \UseMacro{qwen-small_all-naming_sensitivity} & \UseMacro{qwen-medium_all-naming_sensitivity} & \UseMacro{qwen-large_all-naming_sensitivity} \\
  & DeepSeek-Coder & \UseMacro{dscoder-small_all-naming_sensitivity} & \UseMacro{dscoder-medium_all-naming_sensitivity} & \UseMacro{dscoder-large_all-naming_sensitivity} \\
\midrule
\multirow{3}{*}{\textbf{Spacing}}  & Llama-3 & \UseMacro{llama-small_all-spacing_sensitivity} & \UseMacro{llama-medium_all-spacing_sensitivity} & \UseMacro{llama-large_all-spacing_sensitivity} \\
  & Qwen2.5-Coder & \UseMacro{qwen-small_all-spacing_sensitivity} & \UseMacro{qwen-medium_all-spacing_sensitivity} & \UseMacro{qwen-large_all-spacing_sensitivity} \\
  & DeepSeek-Coder & \UseMacro{dscoder-small_all-spacing_sensitivity} & \UseMacro{dscoder-medium_all-spacing_sensitivity} & \UseMacro{dscoder-large_all-spacing_sensitivity} \\
\bottomrule 
\end{tabular} 

}
\end{center}
\vspace{-10pt}
\end{table}

\begin{table}[t]
\begin{center}
\caption{Impact of identifier fragment changes on \sensitivity. ``Unchanged'' samples do not have any identifier fragment change, and ``Changed'' samples have at least one identifier fragment change.}
\label{tab:fragment-sensitivity}
\vspace{-5pt}
\resizebox{.9\columnwidth}{!}{%

\begin{tabular}{@{}ll|rr@{}}
\toprule
\textbf{Rewrite Rule} & \textbf{Model} & \textbf{Unchanged} & \textbf{Changed} \\
\midrule
\multirow{3}{*}{\textbf{Naming}} & \UseMacro{llama-large-short} & \UseMacro{llama-large_fragment-unchanged-naming_sensitivity} & \UseMacro{llama-large_fragment-changed-naming_sensitivity} \\
 & \UseMacro{qwen-large-short} & \UseMacro{qwen-large_fragment-unchanged-naming_sensitivity} & \UseMacro{qwen-large_fragment-changed-naming_sensitivity} \\
 & \UseMacro{deepseek-coder-33b-short} & \UseMacro{dscoder-large_fragment-unchanged-naming_sensitivity} & \UseMacro{dscoder-large_fragment-changed-naming_sensitivity} \\
\midrule
\multirow{3}{*}{\textbf{Spacing}} & \UseMacro{llama-large-short} & \UseMacro{llama-large_fragment-unchanged-spacing_sensitivity} & \UseMacro{llama-large_fragment-changed-spacing_sensitivity} \\
 & \UseMacro{qwen-large-short} & \UseMacro{qwen-large_fragment-unchanged-spacing_sensitivity} & \UseMacro{qwen-large_fragment-changed-spacing_sensitivity} \\
 & \UseMacro{deepseek-coder-33b-short} & \UseMacro{dscoder-large_fragment-unchanged-spacing_sensitivity} & \UseMacro{dscoder-large_fragment-changed-spacing_sensitivity} \\
\bottomrule
\end{tabular}

}
\end{center}
\vspace{-10pt}
\end{table}

We investigate whether larger models are less sensitive to tokenization changes,
with the general assumption of larger models being more robust.
\Cref{tab:model-size-sensitivity} shows the the average \sensitivity of models at different sizes, where the small, medium, and large models in each series are compared on a row.
While the small and medium models are at around the same level of \sensitivity, 
the large models are usually less sensitive (i.e., more robust) than their smaller counterparts, with only one exception of \UseMacro{qwen-large-short} on naming \rewriterules.

We also perform statistically significant tests via Wilcoxon signed-rank test~\cite{conover1999practical}.
The results show that the differences are not significant for naming rules, but significant for spacing rules (except between the small and medium models for \qwen and \dscoder series).

\subsection{Impact of Identifier Fragment Changes}
\label{sec:results:fragment-changes}


We noticed that identifiers are frequently tokenized into different \subwords before and after applying \rewriterules.
For example, \llama tokenizes `\texttt{\textvisiblespace sortedLst}' into three tokens [\texttt{`\textvisiblespace sorted', `L', `st'}], and applying \UseMacro{camel-case-snake-case-no} changes it into two tokens [\texttt{`\textvisiblespace sorted', `\_lst'}].
We define this case as \emph{identifier fragment change}: the list of fragments (tokens but ignoring spaces and underscores) changes before and after applying \rewriterules.
Using this concept, we can categorize the samples into two groups, one without any identifier fragment change (i.e., ``Unchanged''), and the other with at least one identifier fragment change (i.e., ``Changed'').

\Cref{tab:fragment-sensitivity} shows the average \sensitivity of models on the two groups of samples; note that we focus on the large model in each series in this analysis.
The identifier fragment changed group shows consistently higher \sensitivity than the unchanged group,
with the largest difference on \UseMacro{dscoder-large-short} for naming \rewriterules (\UseMacro{dscoder-large_fragment-changed-naming_sensitivity}\% vs. \UseMacro{dscoder-large_fragment-unchanged-naming_sensitivity}\%).
This finding suggests that how identifiers are tokenized into \subwords play an important role in LLMs' understanding of code.
Arguably, identifiers are frequently \emph{not} tokenized into semantically meaningful \subwords (such as the `\texttt{\textvisiblespace sortedLst}' example),
which may fundamentally limit the model's code comprehension and generation capabilities.




\begin{table}[t!]
\centering
\begin{small}
\caption{Word frequency of rewrite rules' left-hand side (LHS) and right-hand side (RHS) on GitHub. Ratio is the percentage of RHS to LHS word frequency.}
\label{tab:github-frequency}
\vspace{-5pt}
\resizebox{.8\columnwidth}{!}{%

\begin{tabular}{@{}lrrr@{}}
\toprule
\textbf{Rewrite Rule} & \textbf{LHS} & \textbf{RHS} & \textbf{Ratio [\%]} \\
\midrule
\multicolumn{4}{c}{\textbf{Java}} \\
\midrule
\UseMacro{rparentheses-period-no}: \UseMacro{\UseMacro{rparentheses-period-no}-short} & \UseMacro{java_rparentheses-period_github-frequency_w} & \UseMacro{java_rparentheses-period_spaced_github-frequency_w} & \UseMacro{java_rparentheses-period_github-frequency_ratio_w} \\
\UseMacro{double-plus-rparentheses-no}: \UseMacro{\UseMacro{double-plus-rparentheses-no}-short} & \UseMacro{java_double-plus-rparentheses_github-frequency_w} & \UseMacro{java_double-plus-rparentheses_spaced_github-frequency_w} & \UseMacro{java_double-plus-rparentheses_github-frequency_ratio_w} \\
\UseMacro{period-asterisk-no}: \UseMacro{\UseMacro{period-asterisk-no}-short} & \UseMacro{java_period-asterisk_github-frequency_w} & \UseMacro{java_period-asterisk_spaced_github-frequency_w} & \UseMacro{java_period-asterisk_github-frequency_ratio_w} \\
\UseMacro{rparentheses-semicolon-no}: \UseMacro{\UseMacro{rparentheses-semicolon-no}-short} & \UseMacro{java_rparentheses-semicolon_github-frequency_w} & \UseMacro{java_rparentheses-semicolon_spaced_github-frequency_w} & \UseMacro{java_rparentheses-semicolon_github-frequency_ratio_w} \\
\UseMacro{rparentheses-rparentheses-no}: \UseMacro{\UseMacro{rparentheses-rparentheses-no}-short} & \UseMacro{java_rparentheses-rparentheses_github-frequency_w} & \UseMacro{java_rparentheses-rparentheses_spaced_github-frequency_w} & \UseMacro{java_rparentheses-rparentheses_github-frequency_ratio_w} \\
\UseMacro{lparentheses-rparentheses-no}: \UseMacro{\UseMacro{lparentheses-rparentheses-no}-short} & \UseMacro{java_lparentheses-rparentheses_github-frequency_w} & \UseMacro{java_lparentheses-rparentheses_spaced_github-frequency_w} & \UseMacro{java_lparentheses-rparentheses_github-frequency_ratio_w} \\
\UseMacro{period-name-no}: \UseMacro{\UseMacro{period-name-no}-short} & \UseMacro{java_period-name_github-frequency_w} & \UseMacro{java_period-name_spaced_github-frequency_w} & \UseMacro{java_period-name_github-frequency_ratio_w} \\
\UseMacro{lparentheses-name-no}: \UseMacro{\UseMacro{lparentheses-name-no}-short} & \UseMacro{java_lparentheses-name_github-frequency_w} & \UseMacro{java_lparentheses-name_spaced_github-frequency_w} & \UseMacro{java_lparentheses-name_github-frequency_ratio_w} \\
\midrule
\multicolumn{4}{c}{\textbf{Python}} \\
\midrule
\UseMacro{rsquarebracket-rparentheses-no}: \UseMacro{\UseMacro{rsquarebracket-rparentheses-no}-short} & \UseMacro{python_rsquarebracket-rparentheses_github-frequency_w} & \UseMacro{python_rsquarebracket-rparentheses_spaced_github-frequency_w} & \UseMacro{python_rsquarebracket-rparentheses_github-frequency_ratio_w} \\
\UseMacro{lsquarebracket-name-no}: \UseMacro{\UseMacro{lsquarebracket-name-no}-short} & \UseMacro{python_lsquarebracket-name_github-frequency_w} & \UseMacro{python_lsquarebracket-name_spaced_github-frequency_w} & \UseMacro{python_lsquarebracket-name_github-frequency_ratio_w} \\
\UseMacro{rparentheses-colon-no}: \UseMacro{\UseMacro{rparentheses-colon-no}-short} & \UseMacro{python_rparentheses-colon_github-frequency_w} & \UseMacro{python_rparentheses-colon_spaced_github-frequency_w} & \UseMacro{python_rparentheses-colon_github-frequency_ratio_w} \\
\UseMacro{rparentheses-rparentheses-no}: \UseMacro{\UseMacro{rparentheses-rparentheses-no}-short} & \UseMacro{python_rparentheses-rparentheses_github-frequency_w} & \UseMacro{python_rparentheses-rparentheses_spaced_github-frequency_w} & \UseMacro{python_rparentheses-rparentheses_github-frequency_ratio_w} \\
\UseMacro{lparentheses-rparentheses-no}: \UseMacro{\UseMacro{lparentheses-rparentheses-no}-short} & \UseMacro{python_lparentheses-rparentheses_github-frequency_w} & \UseMacro{python_lparentheses-rparentheses_spaced_github-frequency_w} & \UseMacro{python_lparentheses-rparentheses_github-frequency_ratio_w} \\
\UseMacro{period-name-no}: \UseMacro{\UseMacro{period-name-no}-short} & \UseMacro{python_period-name_github-frequency_w} & \UseMacro{python_period-name_spaced_github-frequency_w} & \UseMacro{python_period-name_github-frequency_ratio_w} \\
\UseMacro{lparentheses-name-no}: \UseMacro{\UseMacro{lparentheses-name-no}-short} & \UseMacro{python_lparentheses-name_github-frequency_w} & \UseMacro{python_lparentheses-name_spaced_github-frequency_w} & \UseMacro{python_lparentheses-name_github-frequency_ratio_w} \\
\bottomrule
\end{tabular}

}
\end{small}
\vspace{-10pt}
\end{table}

\section{Root Cause Analyses}
\label{sec:interpretation}

In addition to quantifying its impact, we also study \emph{why} LLMs are sensitive to tokenization changes,
along two aspects:
(1)~word frequency in the \pretraining corpus (\Cref{sec:interpretation:word-frequency});
(2)~LLM's hidden states before and after the \rewriterule (\Cref{sec:interpretation:hidden-state}).

\subsection{Word Frequency Analysis}
\label{sec:interpretation:word-frequency}


Our hypothesis is that there is a correlation between \sensitivity and the word frequencies of the \rewriterule's left-hand side and right-hand side.
If the ratio of right-hand side to left-hand side word frequency is small (meaning right-hand side is rare in the corpus), LLMs will likely perform worse after applying the \rewriterule.
We measure the word frequencies on GitHub, a primary source of code data in LLMs' \pretraining corpora.\footnote{
We use GitHub's search feature to measure word frequencies; 
due to the limitation in regular expressions and characters that can be used in the search string, we can only conduct this analysis on a subset of the spacing \rewriterules.}

\Cref{tab:github-frequency} shows the word frequencies of the \rewriterules, and the ratio (in percentages) of the right-hand side to the left-hand side word frequency.
The ratio is always less than 100\%, 
which explains why LLMs exhibit non-negligible \sensitivity to all \rewriterules.
Some \rewriterules with low ratio, e.g., \UseMacro{lparentheses-rparentheses-no}, also exhibit high \sensitivity in \Cref{fig:sensitivity:spacing}.

\subsection{Hidden State Analysis}
\label{sec:interpretation:hidden-state}



\begin{figure}[t!]
  \centering
  \begin{minipage}{\linewidth}
  \centering
  \includegraphics[width=.8\linewidth,trim={0 0 0 0},clip]{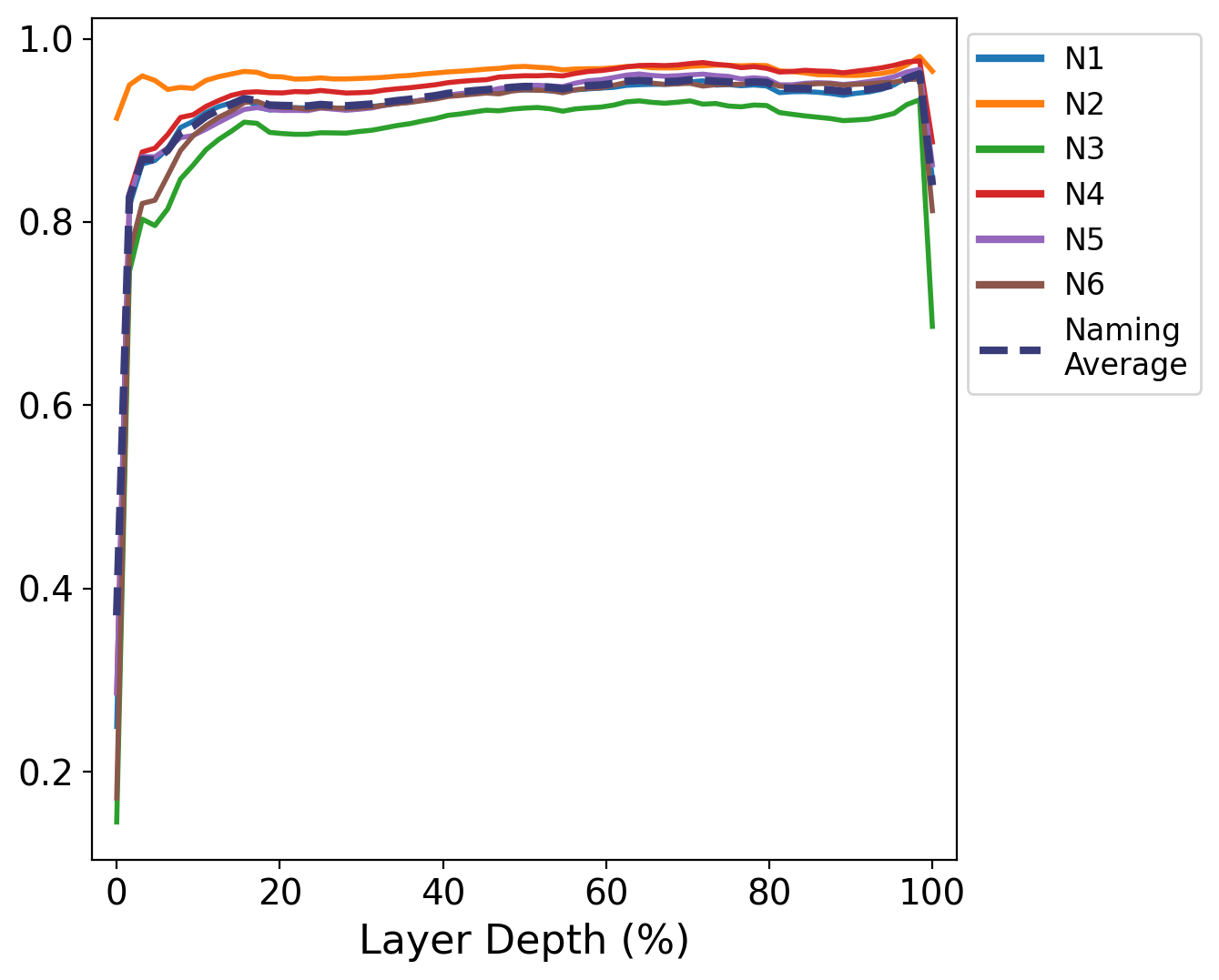}
  \vspace{-5pt}
  \subcaption{naming \rewriterules}
  \label{fig:hidden-state-similarity:naming}
  \end{minipage}
  \begin{minipage}{\linewidth}
  \centering
  \includegraphics[width=.8\linewidth,trim={0 0 0 0},clip]{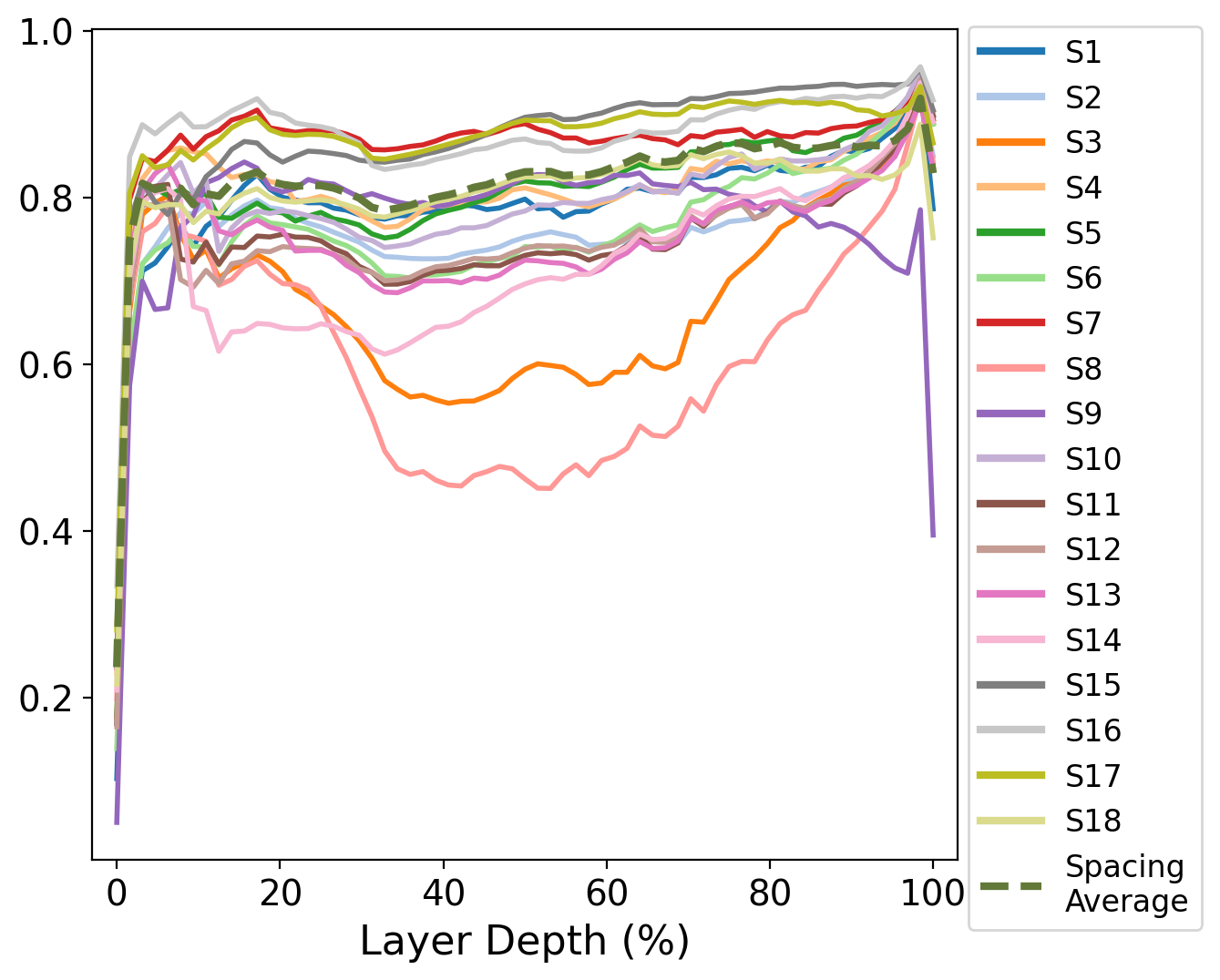}
  \vspace{-5pt}
  \subcaption{spacing \rewriterules}
  \label{fig:hidden-state-similarity:spacing}
  \end{minipage}
\vspace{-5pt}
\caption{The similarity of each layer's hidden states before and after applying \rewriterules.}
\label{fig:hidden-state-similarity}
\vspace{-10pt}
\end{figure}

LLMs' hidden states represent their internal comprehension and reasoning processes, which may help explain their sensitive to tokenization changes.
We compare the hidden states before and after applying the \rewriterules.
For each tokens sequence changed, we extract the hidden states of the \emph{last} token in the sequence, which summarizes the information of the entire sequence.
We focus this analysis on the best-performing LLM, \UseMacro{qwen-large-short}.

We first measure the cosine similarity between the hidden states before and after applying the \rewriterules.
\Cref{fig:hidden-state-similarity} shows correlation between the layer from which the hidden states are extracted and the similarity.
For both naming and spacing \rewriterules, the similarity starts from almost 0 in the first (input) layer, increases (and stabilizes in most cases) in middle layers, and drops again at the last (output) layer.
This observation is consistent with the information bottleneck theory~\citep{saxe2019information}, which states that the middle layers capture the compressed semantic information.
Interestingly, in \Cref{fig:hidden-state-similarity:spacing}, we observe that for some spacing \rewriterules (\UseMacro{lparentheses-rparentheses-no} and \UseMacro{rparentheses-period-no}), the similarity in middle layers is also low, implying that the model sees the before and after versions as semantically different.
These \rewriterules match the ones that LLMs are most sensitive to in \Cref{fig:sensitivity:spacing}.


\begin{figure*}[t!]
\begin{center}
\begin{minipage}{.32\linewidth}
\includegraphics[width=\linewidth]{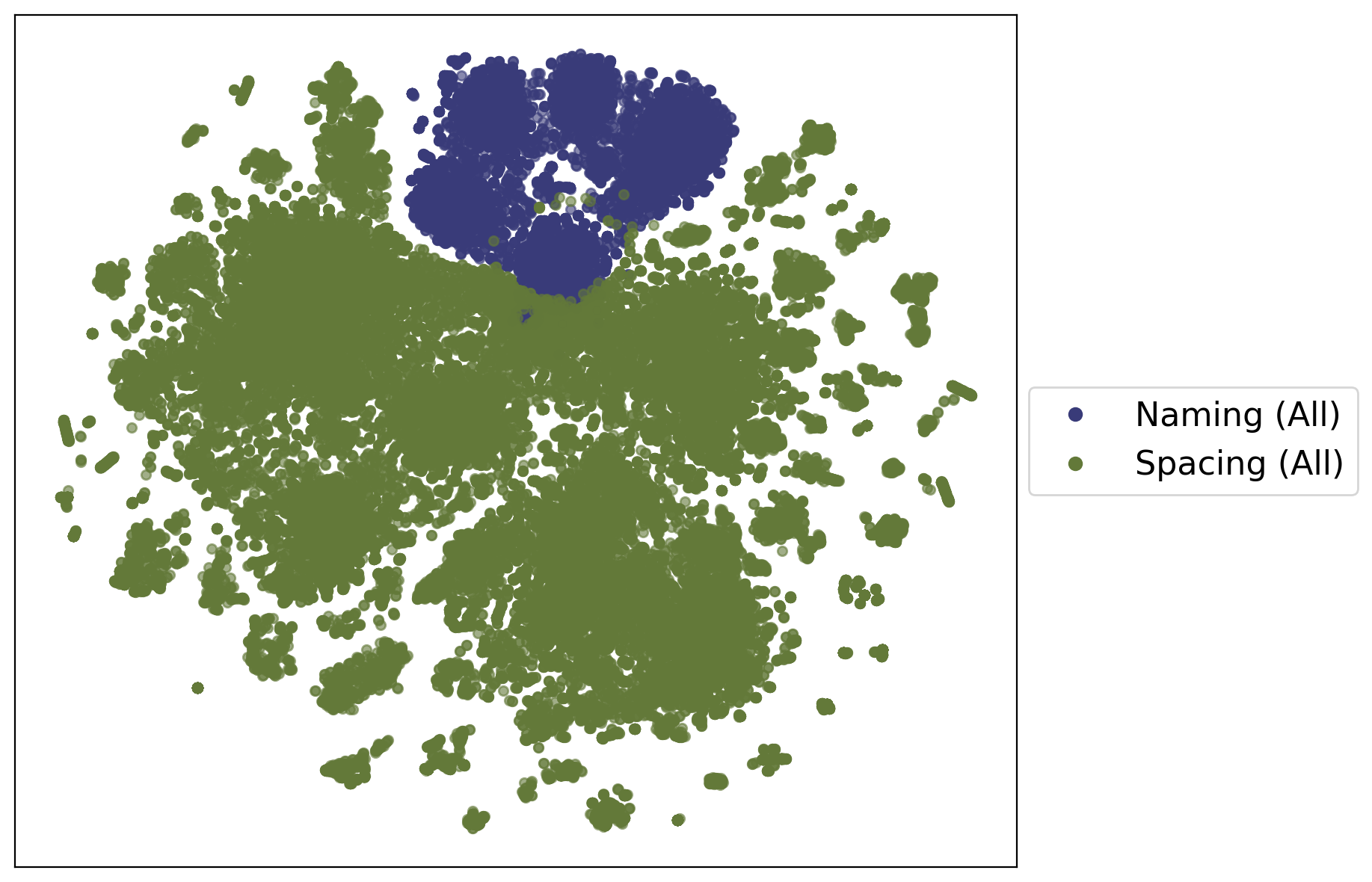}
\subcaption{naming vs. spacing}
\label{fig:hidden-state-visualization:overall}
\end{minipage}
\begin{minipage}{.32\linewidth}
\includegraphics[width=\linewidth]{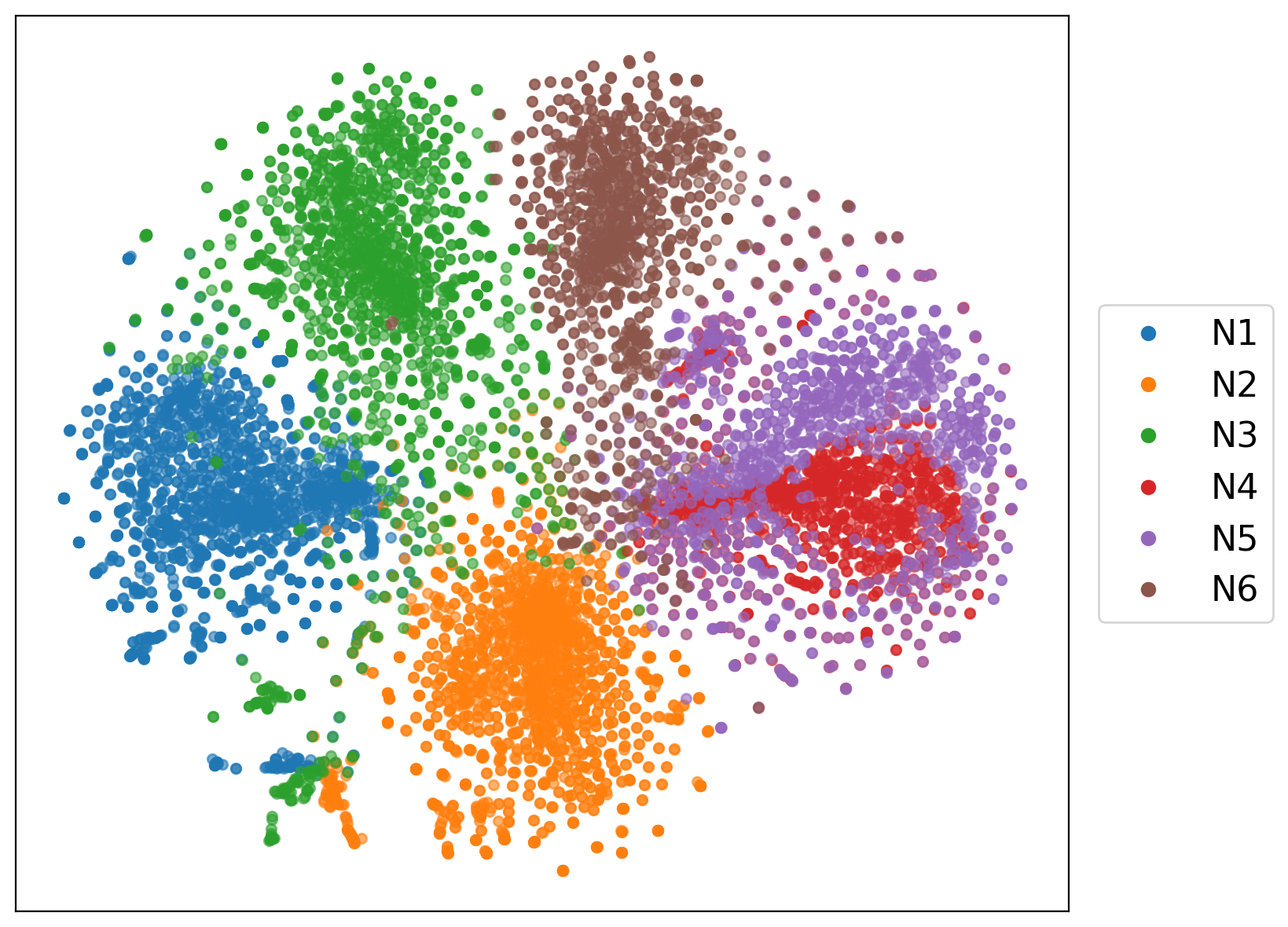}
\subcaption{naming \rewriterules}
\label{fig:hidden-state-visualization:naming}
\end{minipage}
\begin{minipage}{.32\linewidth}
\includegraphics[width=\linewidth]{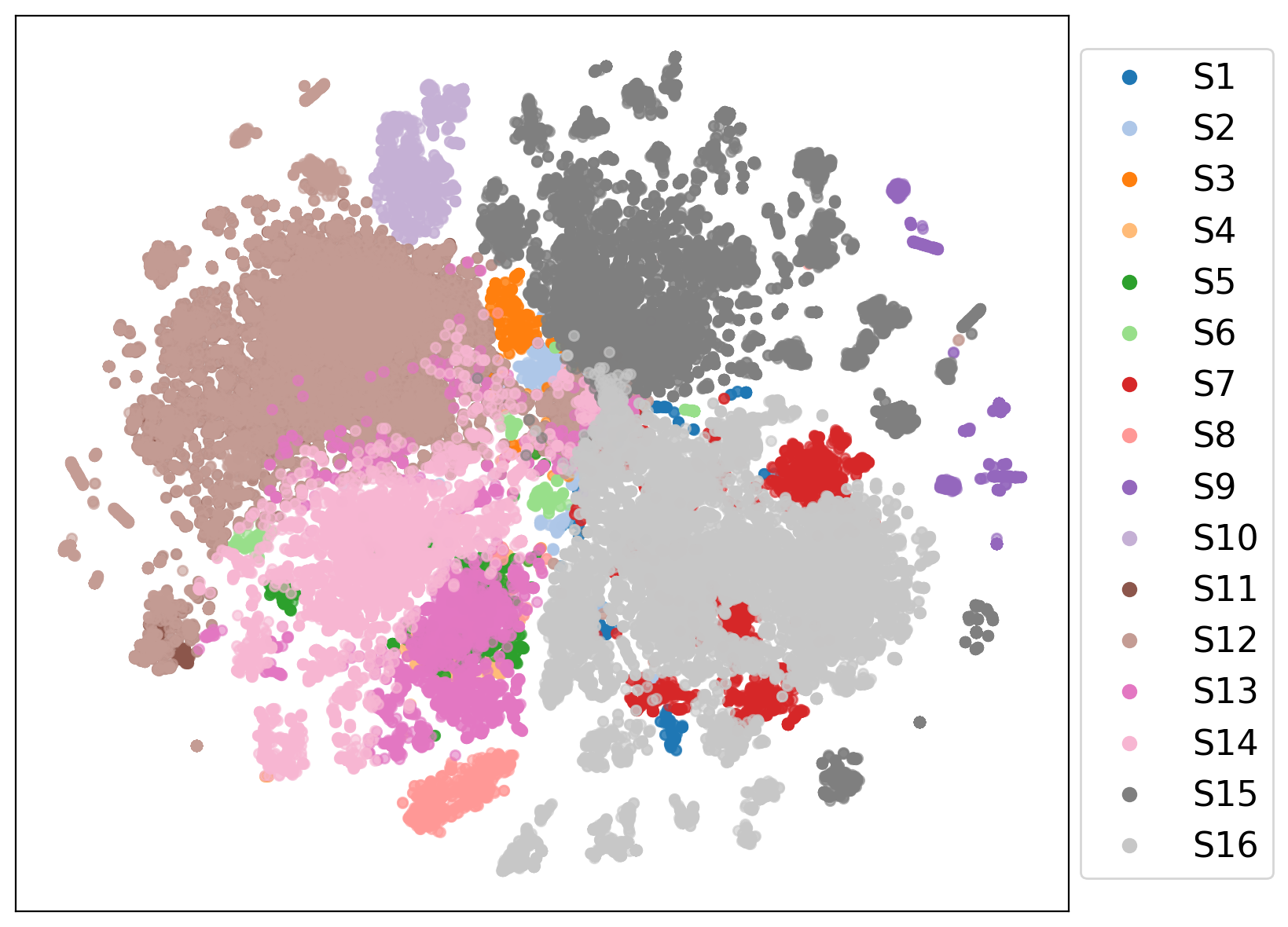}
\subcaption{spacing \rewriterules}
\label{fig:hidden-state-visualization:spacing}
\end{minipage}
\vspace{-5pt}
\caption{Visualizations of the hidden state diffs using t-SNE~\citep{maaten2008visualizing}.}
\label{fig:hidden-state-visualization}
\end{center}
\vspace{-10pt}
\end{figure*}

Then, we compute the \emph{hidden state diffs} as the hidden states after applying \rewriterules minus those before applying,
on the medium layer of the model which should best capture semantic information.
\Cref{fig:hidden-state-visualization} shows the visualizations of the hidden state diffs using t-SNE~\citep{maaten2008visualizing}.
We observe that the diffs of naming and spacing \rewriterules are clearly distinguishable (\Cref{fig:hidden-state-visualization:overall}),
so are the diffs of naming (\Cref{fig:hidden-state-visualization:naming}) and spacing \rewriterules (\Cref{fig:hidden-state-visualization:spacing}, note that \UseMacro{op-name-no} and \UseMacro{op-all-no} are excluded since they are supersets of other \rewriterules).
This confirms that the hidden states, especially from the middle layers, are good representations of semantic information and may be utilized to mitigate the tokenization changes.

\section{Related Work}
\label{sec:related}


\paragraph{Tokenization}
Most modern LLMs use \subword tokenizers such as BPE~\citep{SennrichETAL16BPE}, which create vocabularies based on how often character sequences occur together. 
The resulting token types do not always correspond to meaningful words or code elements, and can vary depending on how the tokenizer was trained. 
For example, \citet{liu2025superbpespacetravellanguage} shows that allowing token merges across whitespace boundaries produces more meaningful units, compared to tokenizers that always split at spaces. \citet{chirkova2023codebpeinvestigatingsubtokenizationoptions} introduces a tokenizer designed to better align with PL syntax, achieving lower token counts while preserving model performance. 
These studies show that tokenization can influence how well a model understands and generates code, and our work builds on this line of inquiry by quantifying the effects of semantic-preserving tokenization changes.

\paragraph{Robustness to Representation Variations}
Another important question is how robust LLMs are to variations in tokenization and representation at inference time. 
\citet{zheng2025brokentokenslanguagemodel} show that instruction-tuned models can often retain high performance even when inputs are tokenized in unconventional or character-level formats, suggesting that such models may learn generalizable internal representations.
However, their study also shows a measurable performance drop compared to standard tokenizations, and other work highlights further limitations. 
\citet{wang2025tokenizationmattersdegradinglarge} find that adversarial changes to token boundaries can significantly degrade model predictions, especially in models that have not undergone instruction tuning. 
In structured domains like chemistry, \citet{D5DD00176E} demonstrate that LLMs produce inconsistent outputs across semantically equivalent molecular representations.
These findings suggest that LLMs remain sensitive to surface-level variations.
Our work contributes to this line by focusing specifically on PLs.

\paragraph{Syntax-Aware Code Modeling}
To address the mismatch between \subword tokenization and PL grammar, several approaches incorporate grammar constraints into the LLM decoding process. 
Synchromesh~\citep{poesia2022synchromeshreliablecodegeneration} and PICARD~\citep{scholak-etal-2021-picard} enforce syntactic validity at generation time by using runtime parsing to filter out invalid token continuations. 
SynCode~\citep{ugare2024syncodellmgenerationgrammar} improves the efficiency of such methods by constructing a DFA-based mask that precomputes token legality while explicitly handling partial tokens. 
Boundless BPE~\citep{schmidt2025boundlessbytepairencoding} removes fixed pretokenizers and enables dynamic boundary selection, allowing the model to learn tokens that correspond to syntactic or semantic units. 
Together, these efforts aim to align LLM outputs more closely with formal code structure, a disconnect that our work quantifies by measuring how semantics-preserving tokenization variations affect model behavior.
\section{Conclusions}
\label{sec:conclusion}

This work studies the tokenization misalignment between \subword-based LLMs and PL grammar. 
While \subword tokenizers like BPE are widely used in code LLMs, they segment inputs based on frequency statistics, not grammar, leading to token boundaries that may not align with syntactic units in code. 
Through a suite of semantic-preserving rewrite rules, our framework \Tool shows that even minor formatting changes, such as whitespace edits or identifier renamings, can cause substantial shifts in model outputs.
These effects hold across nine coding LLMs and three tasks (fixing, summarization, and translation).
These findings motivate future research for grammar-aware or domain-adaptive tokenizers that more faithfully reflect PL structure.

\clearpage
\section*{\label{sec:limitations}Limitations}
While our study shows limitations of current tokenizer designs in code LLMs, our analysis focuses on a targeted set of semantic-preserving rewrites based on common formatting and naming conventions; these do not encompass all potential sources of tokenization drift. Second, although we evaluate nine widely used code LLMs, our findings may not generalize to models with fundamentally different architectures (e.g., state space models~\citep{gu2022efficiently}) or tokenization strategies (e.g., character-level or grammar-driven tokenizers~\citep{Kim_Jernite_Sontag_Rush_2016}). Third, our work centers on measurement and diagnosis, and we do not explore mitigation strategies. Future work could investigate tokenizer retraining, ensemble decoding over multiple tokenizations, or architectural modifications to improve the alignment between token boundaries and programming language syntax.

\section*{Acknowledgments}
We thank Yu Liu for valuable comments and feedback.
This work was supported in part by Compute Ontario (computeontario.ca) and the Digital Research Alliance of Canada (alliancecan.ca). It was also partially supported by a Natural Sciences and Engineering Research Council of Canada (NSERC) Discovery Grant (RGPIN-2024-04909) and a start-up grant from the University of Waterloo.
Yuntian Deng is additionally supported by an NSERC Discovery Grant (RGPIN-2024-05178) and a start-up grant from the University of Waterloo.

\bibliography{bib}

\clearpage
\appendix

\section{Use of LLMs}
We used an LLM-based writing assistant to polish grammar. All ideas, analyses, experiments, and scientific claims are our own, and we take full responsibility for the content of this work.

\section{Additional Background: Tokenizer Differences Between LLMs}
\label{sec:appendix-background}

\begin{figure*}[t]
\centering
\includegraphics[width=\linewidth]{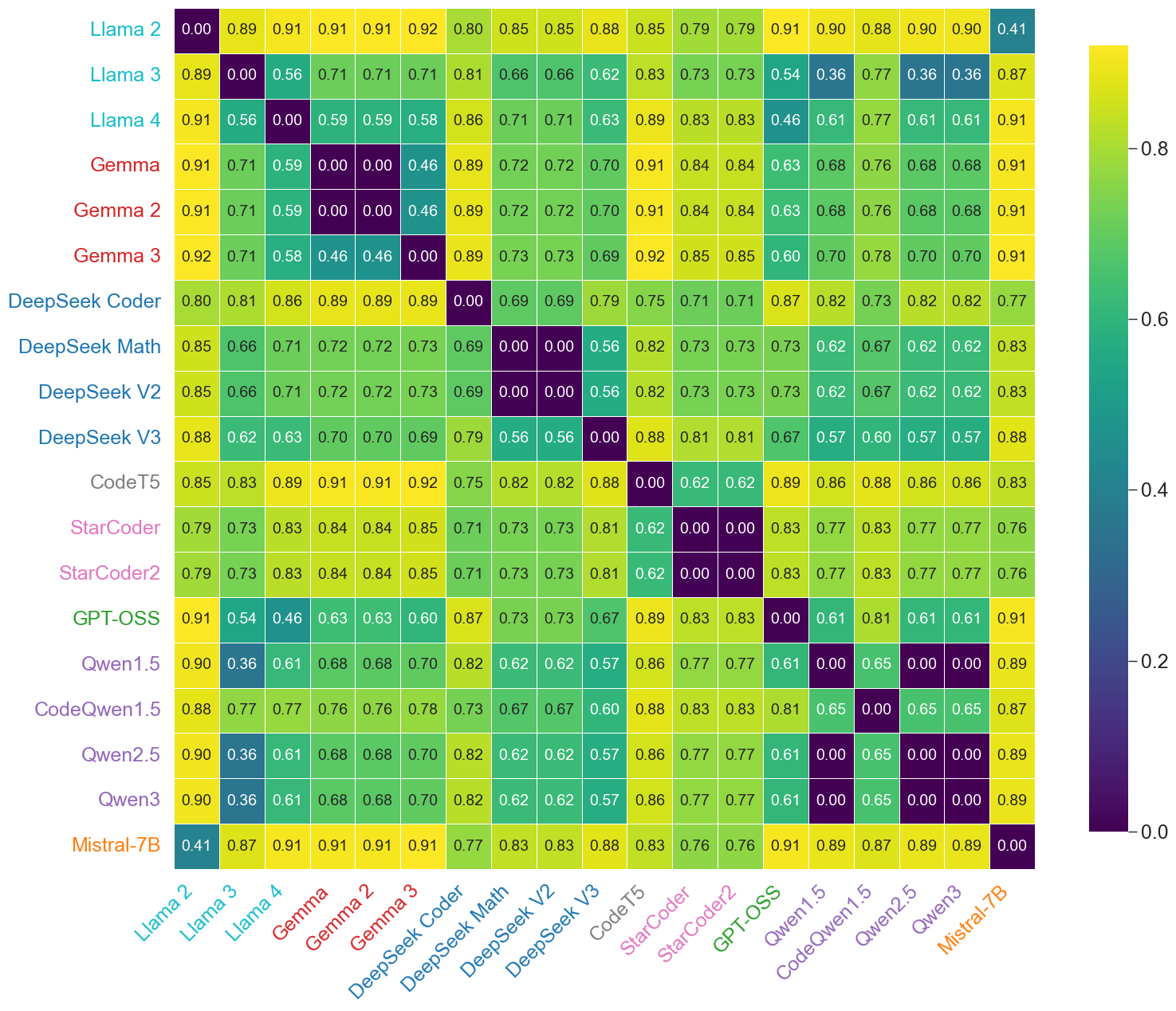}
\caption{Heatmap of vocabulary distances between tokenizers (Full ver.).}
\label{fig:vocb-heatmap-all}
\end{figure*}

\Cref{fig:vocb-heatmap-all} shows the heatmap of vocabulary distances between tokenizers, which includes 19 popular open-source (coding) LLMs from 8 model families. 
Notably, most LLMs adopt a pre-tokenization strategy that splits text into linguistically and layout-meaningful chunks before a byte-level BPE. 
While details vary by family, common choices include isolating short digit runs (often 1--3; Qwen and some DeepSeek variants prefer per-digit), 
treating contiguous letters with combining marks as words, splitting punctuation and symbol runs (sometimes with an optional leading space), 
and separating newline blocks and longer space runs. Non-Latin scripts such as Han/Hiragana/Katakana (and in some cases Hangul) are taken as contiguous spans. 
Family differences that matter for our study include LLaMA-3 explicitly detaching English clitics, CodeQwen-1.5 disabling pre-tokenization (leaving underscores and long ASCII spans intact), 
DeepSeek-Coder using code-oriented splits (letters, punctuation, newlines, CJK, digits), and DeepSeek-V3/LLaMA-4/GPT-OSS converging on a similar unified scheme. 
In practice, more aggressive pre-segmentation tends to make models tolerant to superficial spacing around symbols but sensitive to numeric chunk boundaries, 
whereas byte-only or lightly pre-segmented designs make underscore and identifier edits more likely to introduce new token boundaries.

\section{Additional Experimental Methodology}
\label{sec:appendix-methodology}

\subsection{Benchmarks Normalization}
\label{sec:appendix-methodology:benchmarks-normalization}

To ensure that our semantic-preserving naming/spacing rewrite rules (\Cref{sec:methodology:rewrite}) do not spuriously break compilation or tests, we perform a light-weight normalization pass before evaluation. 

For the bug fixing and code summarization tasks from HumanEvalPack~\citep{MuennighoffETAL23OctoPack}, we first canonicalize Java identifier style to \aCamelCase from \aSnakeCase\footnote{HumanEvalPack~\citep{MuennighoffETAL23OctoPack} translates the HumanEval~\cite{ChenETAL21Evaluating} benchmark from Python to other PLs (including Java), but all the identifiers were remained in \aSnakeCase regardless of the target PL.}, 
then propagate any renamings consistently to tests, entry points, and declarations to preserve their functionalities.

For the code translation tasks,
we start from the Avatar and CodeNet benchmarks prepared by~\citet{pan2024lost}, following their task definitions and tests.
We fixed some samples with harness-compatibility issues that would otherwise cause false negatives and prune a small number of unsalvageable or pathological samples (e.g., extremely long inputs or cases that time out), without changing the underlying problem semantics. 
And finally, we dropped 6 \texttt{python2java} tasks and 4 \texttt{java2python} tasks in Avatar that we could not fix.

The most common adjustments fall into a few categories: (i) IO/formatting normalization. For example, we replace non-portable characters such as U+FFFD or segmentation markers like U+2581 
with ASCII equivalents; ensure consistent tokenization by splitting on spaces instead of empty strings; remove trailing spaces/newlines; standardize numeric output with Java DecimalFormat or 
Python f-strings to fixed precision; (ii) test correctness fixes where expected outputs were inconsistent with the reference implementation or ordering; and (iii) minimal code-context edits that 
preserve semantics but align with tests (e.g., renaming helper methods where tokenizer-specific splits would otherwise occur, adding @Override annotations, or make Scanner/FastScanner usage consistent). 
All edits are specified once, applied uniformly to baseline and variant inputs, and never conditioned on model outputs.

\subsection{Rewrite Algorithms}
\label{sec:appendix-methodology:rewrite-algorithms}

To mutatively rewrite a code context on naming, we first parse it to obtain a code token index and two identifier sets: 
(i) immutable identifiers derived from configured immutable types (e.g., Java: \texttt{importDeclaration}, \texttt{methodCall}; Python: \texttt{import\_as\_name}, \texttt{trailer}); 
(ii) declaration identifiers that are safe to rename (excluding Java methods annotated with \Code{@Override}). We restrict candidates by casing using regexes, specifically, snake case matches \verb![a-z0-9]+(?:_[A-Za-z0-9]+)+! and camel case identifier matches the regex \verb![a-z]+(?:[A-Z]+[A-Za-z0-9]+[A-Za-z0-9]*)+!. For each eligible identifier, 
we segment its lexeme by a well designed regex, convert from the source to the target case, 
and record the absolute character positions in the original string where underscores would be inserted or removed (edit events). For HumanEval tasks, we additionally propagate the same renamings to tests, 
entry points, and declarations to keep the harness consistent, and these are treated as optional ancillary patches and do not alter the core algorithm. 
The immutable/declaration settings aim to maximize safe coverage while preserving compilation and test pass behavior.

Spacing rewrite follows the same structure but, instead of changing identifier lexemes, we insert exactly one space between adjacent tokens when their kinds match a configured 
token‑type bigram (former, latter) from \Cref{tab:rewrite-rules}. For each match, we insert whitespace at the boundary between the two tokens, 
record an insertion event at that position, and update offsets.

Conceptually, although rewrites are defined over PL tokens, the notion of a fragment uses \llmtokens. For each rewrite site, we consider the minimal contiguous list of \llmtokens that 
covers the affected PL tokens (identifiers for naming and the two code tokens of each combination for spacing) as the fragment. Our fragment-change classification is based on an analysis of all fragments' transformation in a code context. Specifically, a merge occurs when at least one old \llmtoken boundary inside those spans disappears, and 
a split occurs when at least one new boundary appears after rewriting. To detect and analyze all \llmtoken boundary transformations, we compute \llmtoken start positions before and after rewriting with the same \llmtokenizer. We ignore boundaries created exactly at an edit site between two code tokens or those created right next to the edit site within one code token. 
Meaning that for insertions, 
we disregard any boundary introduced by the inserted whitespace between two code tokens that were fully or partially combined into one \llmtoken, 
as well as those caused by standalone underscores immediately to its right (a behavior commonly observed in the DeepSeek-Coder or CodeQwen-1.5 tokenizer, where underscores are usually treated as a single token), 
as encoded by the various edit masks classified by edit types in Algorithm~\ref{alg:fragment-classify}. The loop in Algorithm~\ref{alg:fragment-classify} 
shifts the original boundary set by the cumulative $\delta$ pre edit to align coordinate systems, builds the masks for shift edits and edit‑adjacent positions for insert operation, and then compares adjusted old versus filtered new starts. 
Specifically, let $S_{old}$ and $S_{new}$ be the sets of \llmtoken starts before and after rewriting, and after masking specific edit sites, we compute $A{=}S_{old}\setminus S_{new}$ (old boundaries lost) 
and $B{=}S_{new}\setminus S_{old}$ (new boundaries gained). The label is \texttt{unchanged} if $A{=}\emptyset$ and $B{=}\emptyset$, \texttt{merged} if $A{\neq}\emptyset$ and $B{=}\emptyset$, 
\texttt{split} if $A{=}\emptyset$ and $B{\neq}\emptyset$, and \texttt{mixed} otherwise.

\renewcommand{\algorithmicrequire}{\textbf{ Input:}}
\renewcommand{\algorithmicensure}{\textbf{ Output:}}

\begin{algorithm*}[t]
\caption{Naming Rewrite}\label{alg:naming-rewrite-alg}
\begin{algorithmic}[1]
\Require $C$: code context, $P$: code parser, $I_{types}$: immutable identifier types, $\rho_{src}$: source case regex, $tgt$: target case, (optional) $ExtraPatches$: extra patches.
\Ensure $C'$, $E$, (optional) $ExtraPatches'$.
\LComment{$\text{TokIdx}$: list of $(x,\tau,[i,j))$ where $x$=code token, $\tau$=token kind, $[i,j)$=char span}
\State $(\text{TokIdx}, S_{im}, S_{dec}) \gets \textsc{Index}(C, P, I_{types})$ 
\State $E \gets [\,],\; R \gets \emptyset,\; C' \gets C,\; O \gets 0$ \Comment{$E$: list of edit underscore events $(pos,\delta)$; $O$: total offset}
\State \
\For{$(x,\tau,[i,j)) \in \text{TokIdx}$ in ascending $i$}
  \If{$\tau=\text{id} \land (x\notin S_{im} \lor x\in S_{dec}) \land \textsc{RegexCheck}(x,\rho_{src})$}
    \State $y \gets \textsc{CaseConv}(x,tgt)$ \Comment{rewrite the identifier to the target case}
    \State $\Delta_{list} \gets \textsc{DiffUnderlinePos}(x,y,i)$ \Comment{return a list of add/del underscore events $(pos,\delta)$}
    \State $E \gets \textsc{Append}\big(E,\; \Delta_{list})$
    \State $R[x] \gets y$
    \State $C' \gets \textsc{Concat}(C'[0{:}(i{+}O)] , \;y , \; C'[(i{+}O{+}|x|){:}|C'|])$ \Comment{string concatenation}
    \State $O \gets O + (|y| - |x|)$
  \EndIf
\EndFor
\State \
\State $ExtraPatches' \gets \textsc{ApplyRewrites}(ExtraPatches, R)$
\State \Return $(C',\, E,\, ExtraPatches')$
\end{algorithmic}
\end{algorithm*}

\begin{algorithm*}[t]
\caption{Spacing Rewrite}\label{alg:spacing-rewrite-alg}
\begin{algorithmic}[1]
\Require $C$: code context, $P$: code parser, $(K_f,K_\ell)$: token-type bigram.
\Ensure $C'$, $E$.
\LComment{$\text{TokIdx}$: list of $(x,\tau,[i,j))$ where $x$=code token, $\tau$=token kind, $[i,j)$=char span}
\State $\text{TokIdx} \gets \textsc{Index}(C, P)$
\State $E \gets [\,],\; C' \gets C,\; O \gets 0$ \Comment{$E$: list of insert events $(pos,+1)$; $O$: total offset}
\State \
\For{$k \gets 0$ \textbf{to} $|\text{TokIdx}|-2$}
    \State $(x_f,\tau_f,[i_f,j_f)) \gets \text{TokIdx}[k]$; \; $(x_\ell,\tau_\ell,[i_\ell,j_\ell)) \gets \text{TokIdx}[k{+}1]$
    \If{$\textsc{Match}(\tau_f,K_f) \;\land\; \textsc{Match}(\tau_\ell,K_\ell)$}
    \State $E \gets \textsc{Append}(E,\, (i_\ell, {+}1))$
    \State $C' \gets \textsc{Concat}\big(C'[0{:}(j_f{+}O)],\; $"\textvisiblespace"$,\; C'[(i_\ell{+}O){:}|C'|]\big)$ \Comment{insert one space}
    \State $O \gets O + 1$
    \EndIf
\EndFor
\State \
\State \Return $(C',\, E)$
\end{algorithmic}
\end{algorithm*}

\begin{algorithm*}[t]
\caption{Fragment-Change Classification (\textsc{Classify})}\label{alg:fragment-classify}
\begin{algorithmic}[1]
\Require $C$: original code context, $C'$: new code context, $E$: list of edit events $(pos,\delta)$ with $\delta\in\{\pm 1\}$, $EditType$: edit type, $T$: \llmtokenizer.
\Ensure $type \in \{\texttt{unchanged},\,\texttt{merged},\,\texttt{split},\,\texttt{mixed}\}$
\State $L^{old} \gets \textsc{PosLLMTokens}(C, T)$ \Comment{cumulative first character positions of \llmtokens in $C$}
\State $L^{new} \gets \textsc{PosLLMTokens}(C',T)$
\State $S_{old} \gets \textsc{Set}(L^{old})$, $S_{new} \gets \textsc{Set}(L^{new})$, $S_{ed} \gets \{\, pos \mid (pos,\delta)\in E\,\}$, $S_{ed}^{+} \gets \emptyset$
\State $O \gets 0$ \Comment{cumulative offset from prior edits}
\State \
\For{each $(pos,\delta)$ in $E$}
    \State $a \gets pos + O$ \Comment{adjusted position of this edit}
    \State $S_{old} \gets \{\, p{+}\delta \;\textbf{if}\; p > a \;\textbf{else}\; p \mid p\in S_{old}\,\}$
    \State $S_{ed} \gets \{\, e{+}\delta \;\textbf{if}\; e > a \;\textbf{else}\; e \mid e\in S_{ed}\,\}$
    \State $S_{ed}^{+} \gets S_{ed}^{+} \cup \{a +\max(\delta, 0)\}$
    \State $O \gets O + \delta$
\EndFor
\If{$EditType$ = underscore}
\State $S_{new} \gets S_{new} \setminus (S_{ed}^{+} \setminus S_{ed})$ \Comment{ignore starts next-to inserted standalone underscore edit boundaries}
\ElsIf{$EditType$ = whitespace}
\State $S_{new} \gets S_{new} \setminus (S_{ed} \setminus S_{old})$ \Comment{ignore new starts created at whitespace edit boundaries}
\EndIf
\State \
\State $A \gets S_{old} \setminus S_{new}$ \Comment{$A$: lost tokens after rewrite (some tokens merged)}
\State $B \gets S_{new} \setminus S_{old}$ \Comment{$B$: gained tokens after rewrite (some tokens split)}
\State \
\If{$A \neq \emptyset$ \textbf{and} $B = \emptyset$}
    \State \Return \texttt{merged}
\ElsIf{$A = \emptyset$ \textbf{and} $B \neq \emptyset$}
    \State \Return \texttt{split}
\ElsIf{$A \neq \emptyset$ \textbf{and} $B \neq \emptyset$}
    \State \Return \texttt{mixed}
\Else
    \State \Return \texttt{unchanged}
\EndIf
\end{algorithmic}
\end{algorithm*}

\subsection{Metrics Computation Algorithm}
\label{sec:appendix-methodology:metrics-computation-algorithm}

We evaluate on two input programming language subsets for \acc and \accdelta, $X_p$ (Python inputs) and $X_j$ (Java inputs), where their union $X\,{=}\,X_j\cup X_p$ with $|X|\,{=}\,1546$. 
For a fixed rewrite rule $w_i$ and model $m$, let $T_i$ be the deterministic transformation that applies $w_i$ to an input $x\in X$, and we define $T_0$ as no rule would be applied on the input.
And we let $W{=}\{i|i = 0,1, \cdots, 24\}$ denote the assignment set for all rules where $i{=}0$ is the baseline, $i{>}0$ means the variant of applying rule $w_i$. 
Running the model yields code $f_m(T_i(x))$, which the harness evaluates on the test set $\mathcal{T}(x)$. We define the test‑level pass fraction
$$
 r_{m,i}(x) \;\triangleq\; \frac{1}{|\mathcal{T}(x)|} \sum_{t\in\mathcal{T}(x)} [[f_m(T_i(x))]]_t,
$$
where $[[f_m(T_i(x))]]_t \in \{0, 1\}$ denotes the execution result of program $f_m(T_i(x))$ from test $t$ \cite{guan2025your}.
Follow that we define the task‑level correctness indicator $Y_{m, i}(x) \triangleq \mathbb{I}\{ r_{m,i}(x) = 1 \} \in \{0,1\}$. So the \acc of a rule assignment $i \in W$ on a set $S\in\{X_p, X_j\}$ is
$$
 \text{\Acc}_i(m;S) \;=\; \frac{1}{|S|} \sum_{x\in S} Y_{m,i}(x).
$$
We report \accdelta as $\Delta\text{\acc}_i(m;S) \triangleq \text{\Acc}_i(m;S) - \text{\Acc}_0(m;S)$, where $i \in W$ and $i\neq 0$. Not all inputs would be modified by a given rule, we therefore define the actually‑affected subset
$$
 X'_i \;\triangleq\; \{\,x\in X : T_i(x) \neq x\,\},
$$
whose summed sizes for all rules classified by model series are shown in \Cref{tab:effect-size}. Then our proposed \sensitivity measures how often correctness flips among affected inputs only by
$$
 \text{\Sensitivity}_i(m) \triangleq \frac{1}{|X'_i|} \sum_{x\in X'_i} \big| Y_{m,i}(x) - Y_{m,0}(x) \big|.
$$
Intuitively, \accdelta captures net gains/losses which may cancel when aggregating, whereas \sensitivity isolates the flip rate on inputs whose tokens were actually changed by $w_i$.

\subsection{Experimental Environment}
\label{sec:appendix-methodology:experimental-environment}

We conduct all experiments on an NVIDIA H100 GPU cluster, consuming approximately 1840 GPU-hours in total across runs. 
All model checkpoints are obtained from the Hugging Face Hub and loaded with the
Hugging Face Transformers library (v4.53.2) \cite{wolf-etal-2020-transformers}. Unless otherwise stated, models are executed
in fp32, the only exceptions are \UseMacro{llama-large}, \UseMacro{qwen-large}, and \UseMacro{dscoder-large},
which we run in fp16.
All evaluations use the bigcode-evaluation-harness framework \cite{bigcode-evaluation-harness} with its standard protocols. 
We use deterministic decoding without sampling and a batch size of 1 throughout. 
All tests are executed with Java 21.0.1 and Python 3.8.
The maximum generation length is set to 1,024 tokens for HumanEvalPack and Avatar tasks, and 2,048 tokens for CodeNet tasks. 
For t-SNE visualizations, we use scikit-learn v1.7.1 (\texttt{sklearn.manifold.TSNE}) with perplexity to 70 and use the Barnes–Hut method with 1000 iterations, PCA initialization, learning\_rate='auto', and n\_jobs=16 \cite{scikit-learn}.

\section{Additional Results and Analysis}
\label{sec:appendix-results}

\begin{figure*}[t]
\centering
\includegraphics[width=\textwidth]{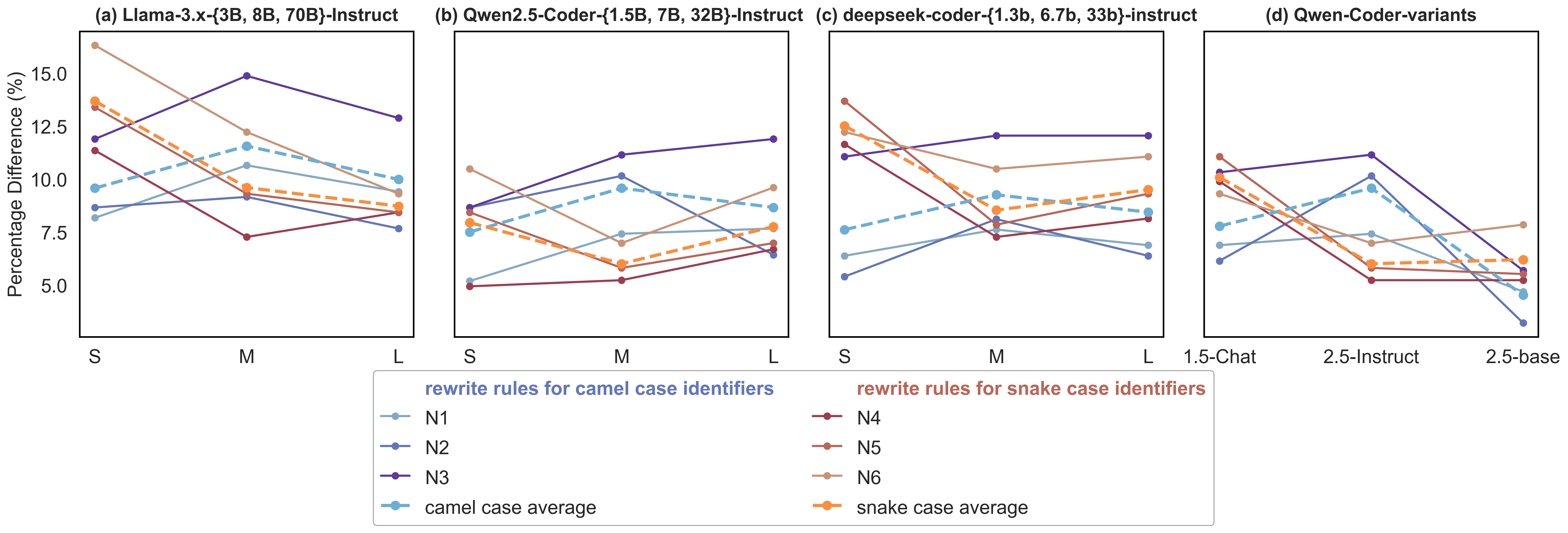}
\caption{Percentage difference for naming rewrite transformations.}
\label{fig:diff_pct_id}
\end{figure*}

\begin{figure*}[t]
\centering
\includegraphics[width=\textwidth]{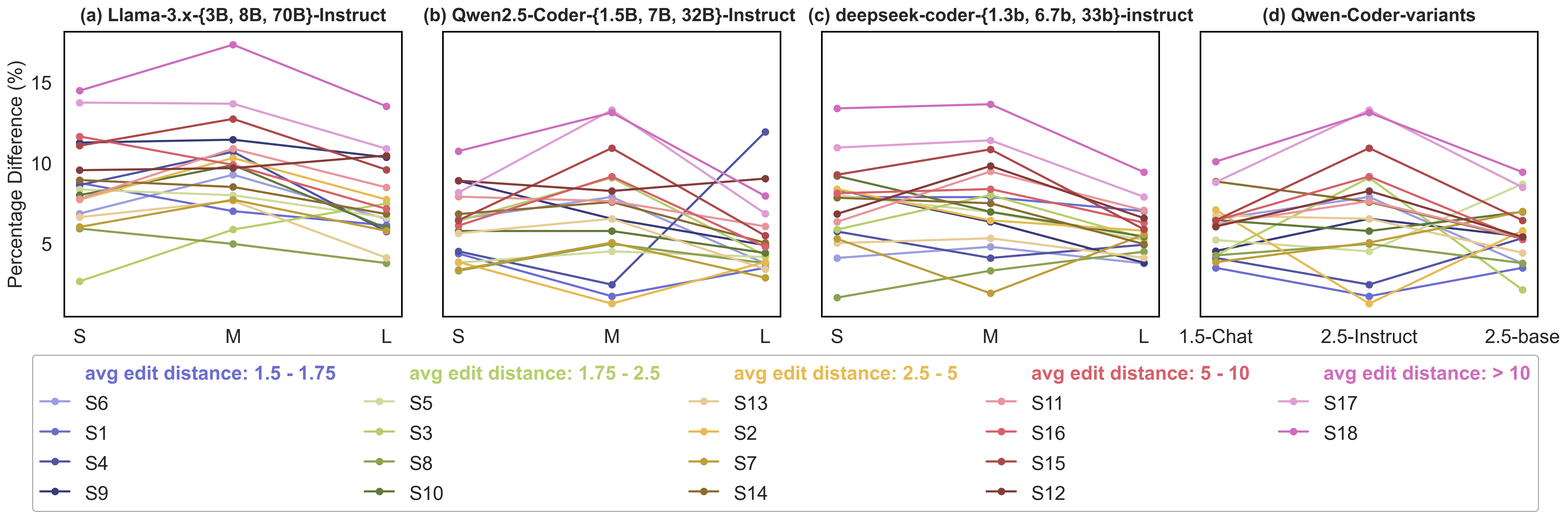}
\caption{Percentage difference for spacing rewrite transformations.}
\label{fig:diff_pct_op}
\end{figure*}

\begin{figure}[t]
\centering
\includegraphics[width=\linewidth]{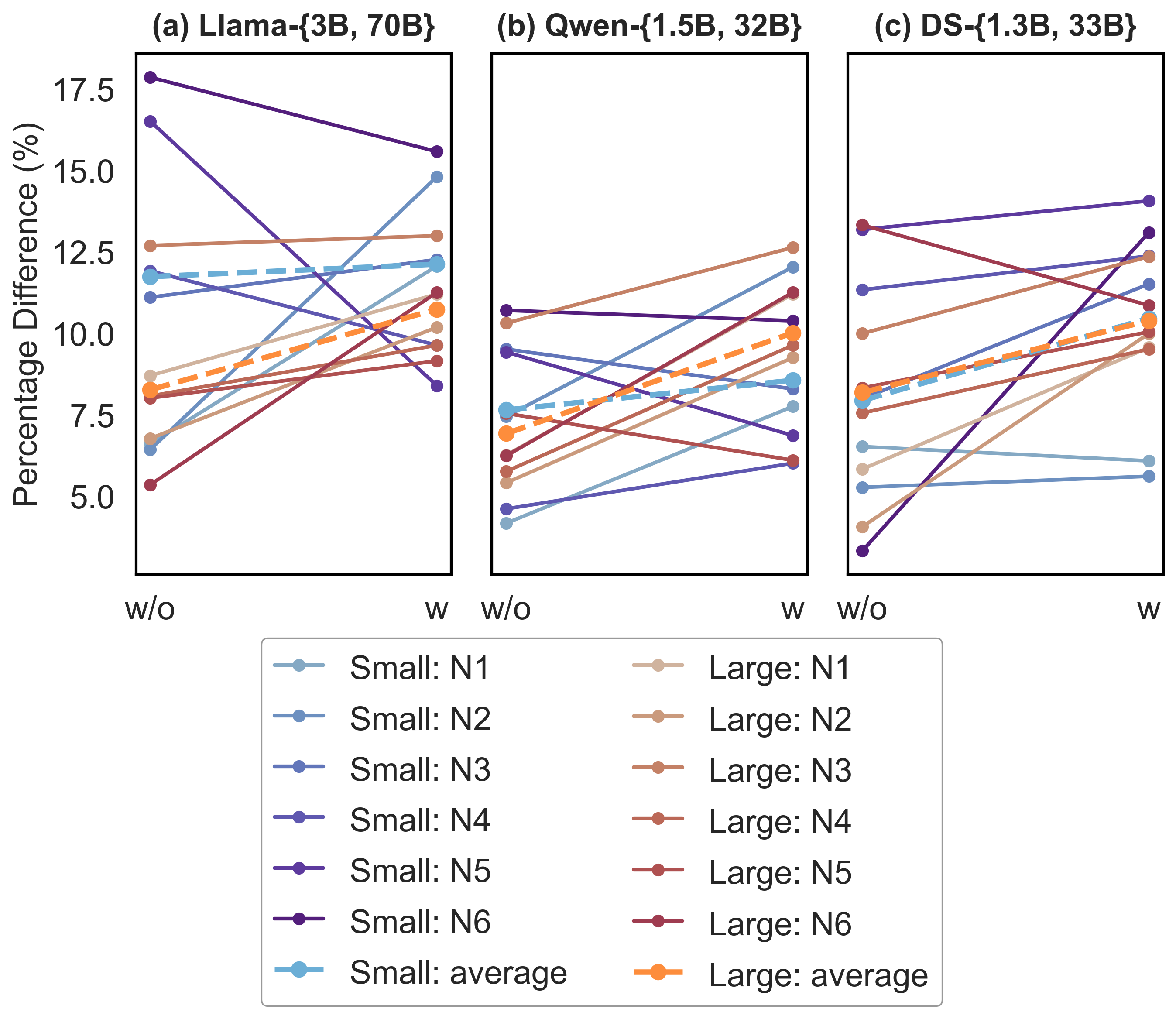}
\caption{Naming rewrite rules percentage difference (with or without fragment change).}
\label{fig:frag_diff_pct_id}
\end{figure}

\begin{figure}[t]
\centering
\includegraphics[width=\linewidth]{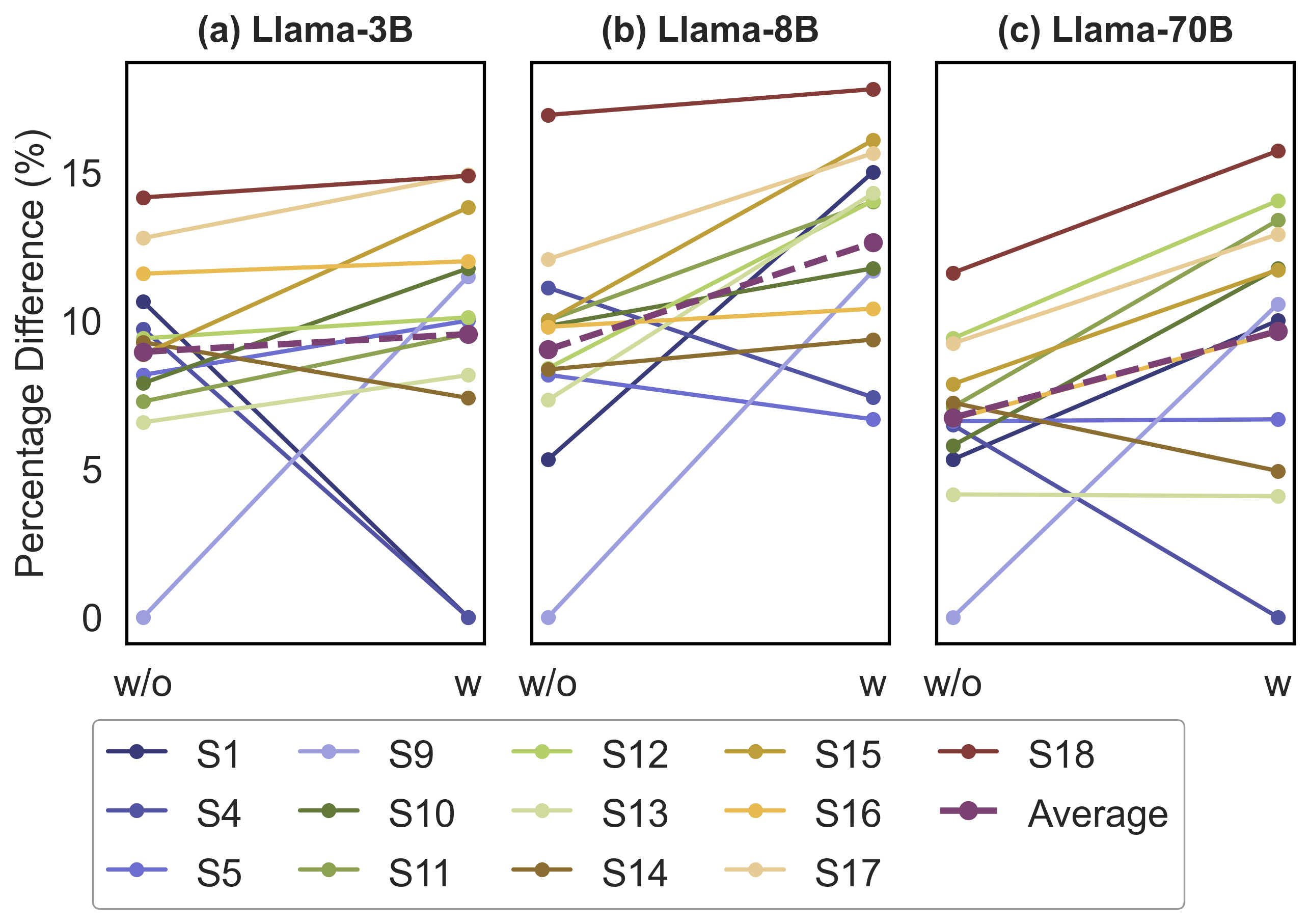}
\caption{(Llama series) Spacing rewrite rules percentage difference (with or without fragment change).}
\label{fig:frag_diff_pct_op_llama}
\end{figure}

\begin{figure}[t]
\centering
\includegraphics[width=\linewidth]{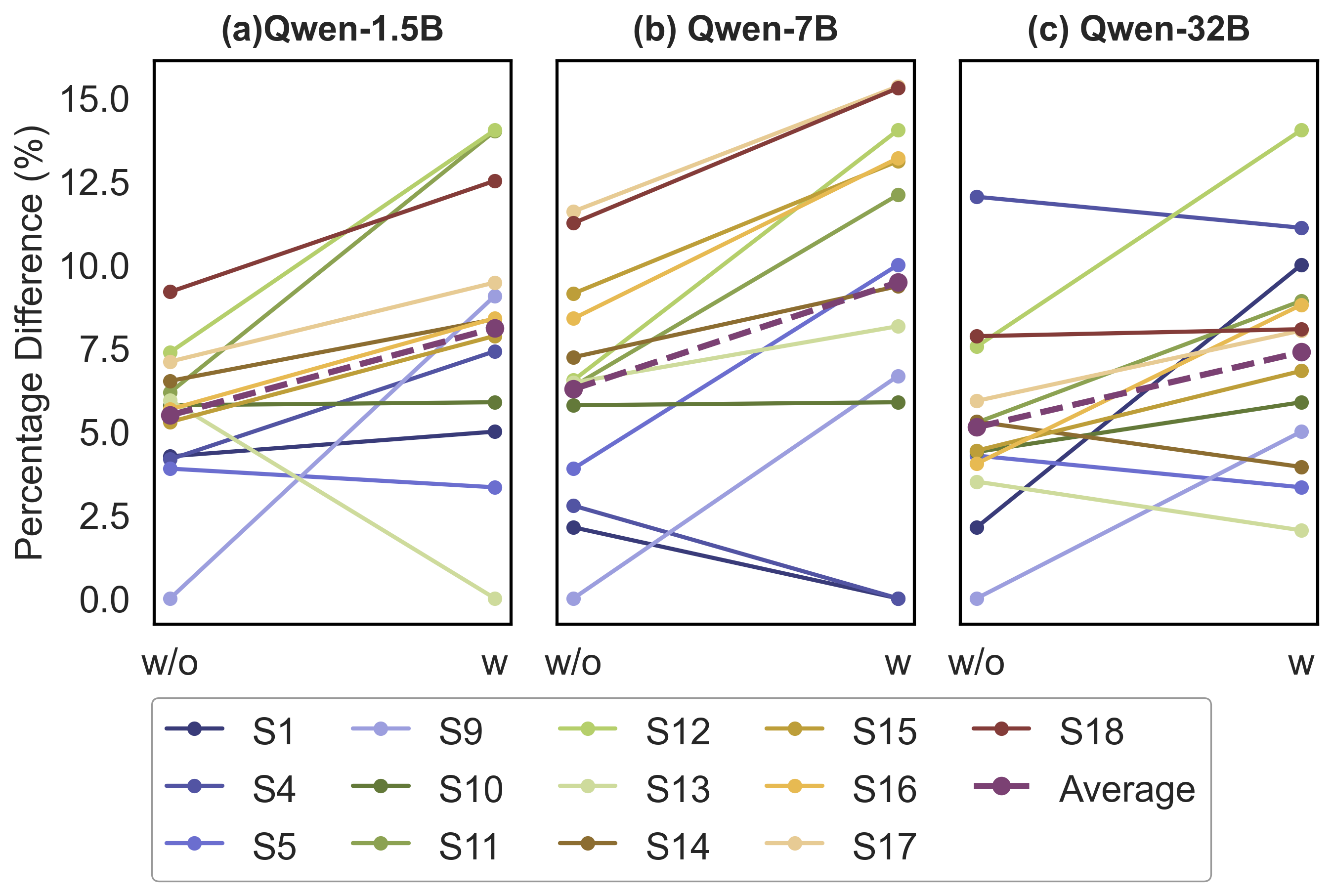}
\caption{(Qwen series) Spacing rewrite rules percentage difference (with or without fragment change).}
\label{fig:frag_diff_pct_op_qwen}
\end{figure}

\begin{figure}[t]
\centering
\includegraphics[width=\linewidth]{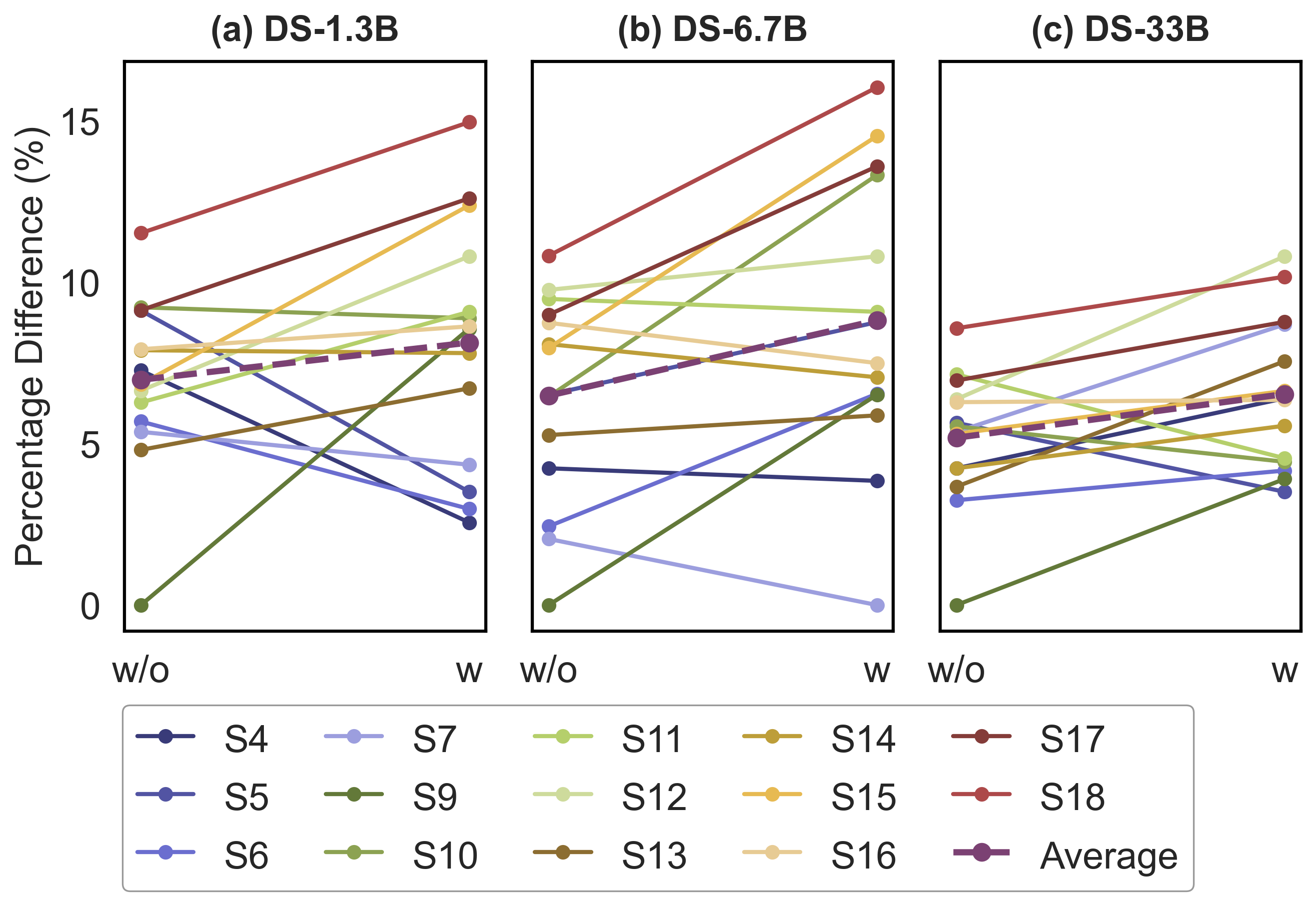}
\caption{(Deepseek series) Spacing rewrite rules percentage difference (with or without fragment change).}
\label{fig:frag_diff_pct_op_dscoder}
\end{figure}

\begin{figure}[t]
\centering
\includegraphics[width=\linewidth]{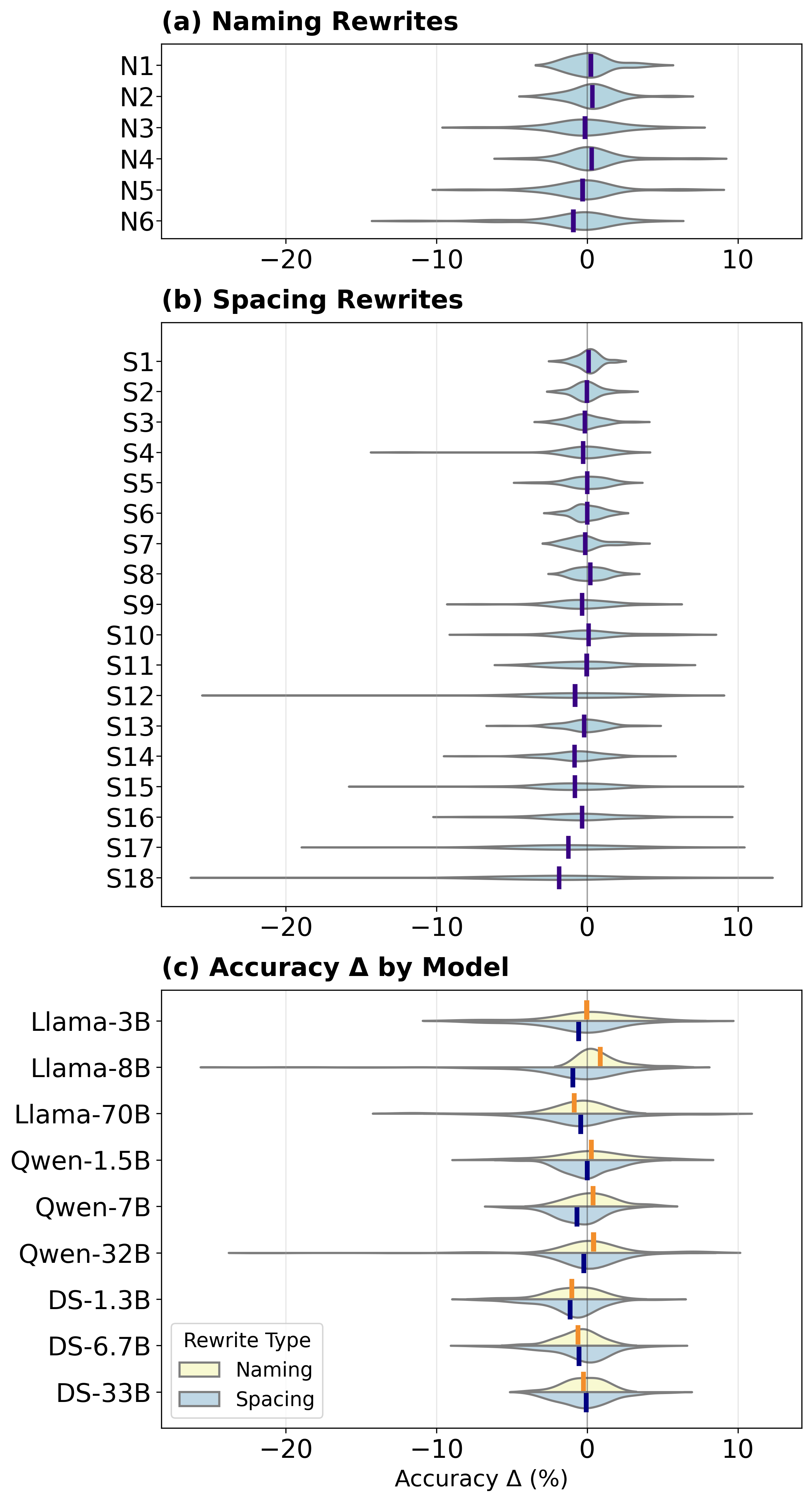}
\caption{Distribution of \accdelta per \rewriterule across models and benchmarks.}
\label{fig:acc_delta}
\end{figure}

In \Cref{fig:diff_pct_id,fig:diff_pct_op}, the line plots summarize \sensitivity for each \rewriterule. In \Cref{fig:frag_diff_pct_id}, comparing samples with and without identifier fragment change shows the overall trend of \sensitivity on different model size. The per‑series breakdowns in \Cref{fig:frag_diff_pct_op_llama,fig:frag_diff_pct_op_qwen,fig:frag_diff_pct_op_dscoder} echo this pattern across Llama‑3, Qwen2.5‑Coder, and DeepSeek‑Coder, while Llama tends to be more sensitive overall, all families exhibit a variation in \sensitivity between "changed" and "unchanged" groups. \\
\Cref{fig:acc_delta} shows the distribution of \accdelta per \rewriterule. Compared to \sensitivity, \accdelta may not be ideal for quantifying robustness because gains and losses cancel and many samples are unaffected. \\
\Cref{tab:effect-size} reports, for each \rewriterule, the number of benchmark samples actually modified, stratified by model series. \Cref{tab:fragment-sensitivity-all} provides the full breakdown of \sensitivity by fragment‑change category. Together, these tables clarify both the scope of input perturbations and the source of robustness differences observed in the main results.

\begin{table*}[t]
\centering
\caption{Samples that been modified by \rewriterule, broken down by model series.}
\label{tab:effect-size}

\begin{tabular}{llrrrrrr}
\toprule
\multirow{2}{*}{\textbf{Rewrite Rule}} & \multirow{2}{*}{\textbf{Model Series}} & \multirow{2}{*}{\textbf{Total}} & \multirow{2}{*}{\textbf{Unchanged}} & \multicolumn{4}{c}{\textbf{Changed}} \\ \cline{5-8}
 & & & & {\scriptsize\textbf{All}} & {\scriptsize\textbf{Merged}} & {\scriptsize\textbf{Split}} & {\scriptsize\textbf{Mixed}} \\
\midrule
\multirow{4}{*}{\textbf{Naming}} & Llama-3 & \UseMacro{Llama-3_naming_all} & \UseMacro{Llama-3_naming_unchanged} & \UseMacro{Llama-3_naming_changed} & \UseMacro{Llama-3_naming_merged} & \UseMacro{Llama-3_naming_split} & \UseMacro{Llama-3_naming_mixed} \\
 & Qwen2.5-Coder & \UseMacro{Qwen25-Coder_naming_all} & \UseMacro{Qwen25-Coder_naming_unchanged} & \UseMacro{Qwen25-Coder_naming_changed} & \UseMacro{Qwen25-Coder_naming_merged} & \UseMacro{Qwen25-Coder_naming_split} & \UseMacro{Qwen25-Coder_naming_mixed} \\
 & deepseek-coder & \UseMacro{deepseek-coder_naming_all} & \UseMacro{deepseek-coder_naming_unchanged} & \UseMacro{deepseek-coder_naming_changed} & \UseMacro{deepseek-coder_naming_merged} & \UseMacro{deepseek-coder_naming_split} & \UseMacro{deepseek-coder_naming_mixed} \\
 & CodeQwen1.5 & \UseMacro{CodeQwen15_naming_all} & \UseMacro{CodeQwen15_naming_unchanged} & \UseMacro{CodeQwen15_naming_changed} & \UseMacro{CodeQwen15_naming_merged} & \UseMacro{CodeQwen15_naming_split} & \UseMacro{CodeQwen15_naming_mixed} \\
\midrule
\multirow{4}{*}{\textbf{Spacing}} & Llama-3 & \UseMacro{Llama-3_spacing_all} & \UseMacro{Llama-3_spacing_unchanged} & \UseMacro{Llama-3_spacing_changed} & \UseMacro{Llama-3_spacing_merged} & \UseMacro{Llama-3_spacing_split} & \UseMacro{Llama-3_spacing_mixed} \\
 & Qwen2.5-Coder & \UseMacro{Qwen25-Coder_spacing_all} & \UseMacro{Qwen25-Coder_spacing_unchanged} & \UseMacro{Qwen25-Coder_spacing_changed} & \UseMacro{Qwen25-Coder_spacing_merged} & \UseMacro{Qwen25-Coder_spacing_split} & \UseMacro{Qwen25-Coder_spacing_mixed} \\
 & deepseek-coder & \UseMacro{deepseek-coder_spacing_all} & \UseMacro{deepseek-coder_spacing_unchanged} & \UseMacro{deepseek-coder_spacing_changed} & \UseMacro{deepseek-coder_spacing_merged} & \UseMacro{deepseek-coder_spacing_split} & \UseMacro{deepseek-coder_spacing_mixed} \\
 & CodeQwen1.5 & \UseMacro{CodeQwen15_spacing_all} & \UseMacro{CodeQwen15_spacing_unchanged} & \UseMacro{CodeQwen15_spacing_changed} & \UseMacro{CodeQwen15_spacing_merged} & \UseMacro{CodeQwen15_spacing_split} & \UseMacro{CodeQwen15_spacing_mixed} \\
\bottomrule
\end{tabular}

\end{table*}



\begin{table*}[!t]
\centering
\caption{Impact of different types of fragment change on \sensitivity (Full ver.).}
\label{tab:fragment-sensitivity-all}
\resizebox{\textwidth}{!}{%

\begin{tabular}{ll|r|rr|rrr}
\toprule
\multirow{2}{*}{\textbf{Rewrite Rule}} & \multirow{2}{*}{\textbf{Model}} & \multirow{2}{*}{\textbf{Total}} & \multirow{2}{*}{\textbf{Unchanged}} & \multirow{2}{*}{\textbf{Changed (all)}} & \multicolumn{3}{c}{\textbf{Changed (subcategories)}} \\
 & & & & & {\scriptsize\textbf{Merged}} & {\scriptsize\textbf{Split}} & {\scriptsize\textbf{Mixed}} \\
\midrule
\multirow{9}{*}{\textbf{Naming}} & \UseMacro{llama-small-init}  & \UseMacro{llama-small_all-naming_sensitivity} & \UseMacro{llama-small_fragment-unchanged-naming_sensitivity} & \UseMacro{llama-small_fragment-changed-naming_sensitivity} & \UseMacro{llama-small_merging-naming_sensitivity} & \UseMacro{llama-small_splitting-naming_sensitivity} & \UseMacro{llama-small_mixing-naming_sensitivity} \\
 & \UseMacro{llama-medium-init} & \UseMacro{llama-medium_all-naming_sensitivity} & \UseMacro{llama-medium_fragment-unchanged-naming_sensitivity} & \UseMacro{llama-medium_fragment-changed-naming_sensitivity} & \UseMacro{llama-medium_merging-naming_sensitivity} & \UseMacro{llama-medium_splitting-naming_sensitivity} & \UseMacro{llama-medium_mixing-naming_sensitivity} \\
 & \UseMacro{llama-large-init} & \UseMacro{llama-large_all-naming_sensitivity} & \UseMacro{llama-large_fragment-unchanged-naming_sensitivity} & \UseMacro{llama-large_fragment-changed-naming_sensitivity} & \UseMacro{llama-large_merging-naming_sensitivity} & \UseMacro{llama-large_splitting-naming_sensitivity} & \UseMacro{llama-large_mixing-naming_sensitivity} \\ \cline{2-8}
 & \UseMacro{qwen-small-init} & \UseMacro{qwen-small_all-naming_sensitivity} & \UseMacro{qwen-small_fragment-unchanged-naming_sensitivity} & \UseMacro{qwen-small_fragment-changed-naming_sensitivity} & \UseMacro{qwen-small_merging-naming_sensitivity} & \UseMacro{qwen-small_splitting-naming_sensitivity} & \UseMacro{qwen-small_mixing-naming_sensitivity} \\
 & \UseMacro{qwen-medium-init} & \UseMacro{qwen-medium_all-naming_sensitivity} & \UseMacro{qwen-medium_fragment-unchanged-naming_sensitivity} & \UseMacro{qwen-medium_fragment-changed-naming_sensitivity} & \UseMacro{qwen-medium_merging-naming_sensitivity} & \UseMacro{qwen-medium_splitting-naming_sensitivity} & \UseMacro{qwen-medium_mixing-naming_sensitivity} \\
 & \UseMacro{qwen-large-init} & \UseMacro{qwen-large_all-naming_sensitivity} & \UseMacro{qwen-large_fragment-unchanged-naming_sensitivity} & \UseMacro{qwen-large_fragment-changed-naming_sensitivity} & \UseMacro{qwen-large_merging-naming_sensitivity} & \UseMacro{qwen-large_splitting-naming_sensitivity} & \UseMacro{qwen-large_mixing-naming_sensitivity} \\ \cline{2-8}
 & \UseMacro{dscoder-small-init} & \UseMacro{dscoder-small_all-naming_sensitivity} & \UseMacro{dscoder-small_fragment-unchanged-naming_sensitivity} & \UseMacro{dscoder-small_fragment-changed-naming_sensitivity} & \UseMacro{dscoder-small_merging-naming_sensitivity} & \UseMacro{dscoder-small_splitting-naming_sensitivity} & \UseMacro{dscoder-small_mixing-naming_sensitivity} \\
 & \UseMacro{dscoder-medium-init} & \UseMacro{dscoder-medium_all-naming_sensitivity} & \UseMacro{dscoder-medium_fragment-unchanged-naming_sensitivity} & \UseMacro{dscoder-medium_fragment-changed-naming_sensitivity} & \UseMacro{dscoder-medium_merging-naming_sensitivity} & \UseMacro{dscoder-medium_splitting-naming_sensitivity} & \UseMacro{dscoder-medium_mixing-naming_sensitivity} \\
 & \UseMacro{dscoder-large-init} & \UseMacro{dscoder-large_all-naming_sensitivity} & \UseMacro{dscoder-large_fragment-unchanged-naming_sensitivity} & \UseMacro{dscoder-large_fragment-changed-naming_sensitivity} & \UseMacro{dscoder-large_merging-naming_sensitivity} & \UseMacro{dscoder-large_splitting-naming_sensitivity} & \UseMacro{dscoder-large_mixing-naming_sensitivity} \\
\midrule
\multirow{9}{*}{\textbf{Spacing}} & \UseMacro{llama-small-init} & \UseMacro{llama-small_all-spacing_sensitivity} & \UseMacro{llama-small_fragment-unchanged-spacing_sensitivity} & \UseMacro{llama-small_fragment-changed-spacing_sensitivity} & \UseMacro{llama-small_merging-spacing_sensitivity} & \UseMacro{llama-small_splitting-spacing_sensitivity} & \UseMacro{llama-small_mixing-spacing_sensitivity} \\
 & \UseMacro{llama-medium-init} & \UseMacro{llama-medium_all-spacing_sensitivity} & \UseMacro{llama-medium_fragment-unchanged-spacing_sensitivity} & \UseMacro{llama-medium_fragment-changed-spacing_sensitivity} & \UseMacro{llama-medium_merging-spacing_sensitivity} & \UseMacro{llama-medium_splitting-spacing_sensitivity} & \UseMacro{llama-medium_mixing-spacing_sensitivity} \\
 & \UseMacro{llama-large-init} & \UseMacro{llama-large_all-spacing_sensitivity} & \UseMacro{llama-large_fragment-unchanged-spacing_sensitivity} & \UseMacro{llama-large_fragment-changed-spacing_sensitivity} & \UseMacro{llama-large_merging-spacing_sensitivity} & \UseMacro{llama-large_splitting-spacing_sensitivity} & \UseMacro{llama-large_mixing-spacing_sensitivity} \\ \cline{2-8}
 & \UseMacro{qwen-small-init} & \UseMacro{qwen-small_all-spacing_sensitivity} & \UseMacro{qwen-small_fragment-unchanged-spacing_sensitivity} & \UseMacro{qwen-small_fragment-changed-spacing_sensitivity} & \UseMacro{qwen-small_merging-spacing_sensitivity} & \UseMacro{qwen-small_splitting-spacing_sensitivity} & \UseMacro{qwen-small_mixing-spacing_sensitivity} \\
 & \UseMacro{qwen-medium-init} & \UseMacro{qwen-medium_all-spacing_sensitivity} & \UseMacro{qwen-medium_fragment-unchanged-spacing_sensitivity} & \UseMacro{qwen-medium_fragment-changed-spacing_sensitivity} & \UseMacro{qwen-medium_merging-spacing_sensitivity} & \UseMacro{qwen-medium_splitting-spacing_sensitivity} & \UseMacro{qwen-medium_mixing-spacing_sensitivity} \\
 & \UseMacro{qwen-large-init} & \UseMacro{qwen-large_all-spacing_sensitivity} & \UseMacro{qwen-large_fragment-unchanged-spacing_sensitivity} & \UseMacro{qwen-large_fragment-changed-spacing_sensitivity} & \UseMacro{qwen-large_merging-spacing_sensitivity} & \UseMacro{qwen-large_splitting-spacing_sensitivity} & \UseMacro{qwen-large_mixing-spacing_sensitivity} \\ \cline{2-8}
 & \UseMacro{dscoder-small-init} & \UseMacro{dscoder-small_all-spacing_sensitivity} & \UseMacro{dscoder-small_fragment-unchanged-spacing_sensitivity} & \UseMacro{dscoder-small_fragment-changed-spacing_sensitivity} & \UseMacro{dscoder-small_merging-spacing_sensitivity} & \UseMacro{dscoder-small_splitting-spacing_sensitivity} & \UseMacro{dscoder-small_mixing-spacing_sensitivity} \\
 & \UseMacro{dscoder-medium-init} & \UseMacro{dscoder-medium_all-spacing_sensitivity} & \UseMacro{dscoder-medium_fragment-unchanged-spacing_sensitivity} & \UseMacro{dscoder-medium_fragment-changed-spacing_sensitivity} & \UseMacro{dscoder-medium_merging-spacing_sensitivity} & \UseMacro{dscoder-medium_splitting-spacing_sensitivity} & \UseMacro{dscoder-medium_mixing-spacing_sensitivity} \\
 & \UseMacro{dscoder-large-init} & \UseMacro{dscoder-large_all-spacing_sensitivity} & \UseMacro{dscoder-large_fragment-unchanged-spacing_sensitivity} & \UseMacro{dscoder-large_fragment-changed-spacing_sensitivity} & \UseMacro{dscoder-large_merging-spacing_sensitivity} & \UseMacro{dscoder-large_splitting-spacing_sensitivity} & \UseMacro{dscoder-large_mixing-spacing_sensitivity} \\
\bottomrule
\end{tabular}

}
\end{table*}

\end{document}